\documentclass{article}
\usepackage{microtype}
\usepackage{graphicx}
\usepackage{subcaption}
\usepackage{booktabs} 
\usepackage{hyperref}

\usepackage[accepted]{icml2026}

\usepackage{url}

\usepackage{minitoc}
\mtcsetrules{parttoc}{off}                

\usepackage[utf8]{inputenc} 
\usepackage[T1]{fontenc}    
\usepackage{hyperref}       
\usepackage{url}            
\usepackage{booktabs}       
\usepackage{amsfonts}       
\usepackage{nicefrac}       
\usepackage{microtype}      
\usepackage{xcolor}         

\usepackage{graphicx}
\usepackage{svg}
\usepackage{multirow}
\usepackage{amsmath,amssymb,amsfonts}
\usepackage{amsthm}
\usepackage{mathrsfs}
\usepackage[title]{appendix}
\usepackage{xcolor}
\usepackage{textcomp}
\usepackage{manyfoot}
\usepackage{booktabs}
\usepackage{listings}
\usepackage{tikz}
\usetikzlibrary{shapes, arrows.meta, positioning}

\usepackage{amsmath}
\usepackage{amssymb}
\usepackage{amsthm}
\usepackage{graphicx}
\usepackage{csquotes}
\usepackage{mathtools}
\usepackage{pst-node}
\usepackage{booktabs}
\usepackage{subcaption}
\usepackage[title]{appendix}
\usepackage{listings}
\usepackage{placeins}
\usepackage{bm}
\usepackage{graphicx}
\usepackage{makecell}
\usepackage{arydshln}
\usepackage{pifont}

\usepackage{wrapfig}

\theoremstyle{plain}

\newtheorem{proposition}{Proposition}

\newtheorem*{proposition*}{Proposition}
\newtheorem*{lemma*}{Lemma}

\theoremstyle{definition}
\newtheorem{definition}{Definition}

\theoremstyle{remark}
\newtheorem{remark}{Remark}

\usepackage{enumitem}

\newcommand{\ie}{\textit{i}.\textit{e}., }
\newcommand{\eg}{\textit{e}.\textit{g}., }

\DeclareMathOperator{\tr}{\operatorname{tr}}
\DeclareMathOperator*{\argmin}{\operatorname{arg\,min}}

\theoremstyle{definition}

\newcommand{\xmark}{\ding{55}} 
\newcommand{\cmark}{\ding{51}}

\newcommand{\method}{FAHNES}

\begin{document}

\twocolumn[
\icmltitle{Feature-Aware (Hyper)graph Generation via Next-Scale Prediction}

\icmlsetsymbol{equal}{*}

\begin{icmlauthorlist}
  \icmlauthor{Dorian Gailhard}{yyy}
  \icmlauthor{Enzo Tartaglione}{yyy}
  \icmlauthor{Lirida Naviner}{yyy}
  \icmlauthor{Jhony H. Giraldo}{yyy}
\end{icmlauthorlist}

\icmlaffiliation{yyy}{LTCI, Télécom Paris, Institut Polytechnique de Paris, France}

\icmlcorrespondingauthor{Dorian Gailhard}{dorian.gailhard@telecom-paris.fr}
\icmlcorrespondingauthor{Enzo Tartaglione}{enzo.tartaglione@telecom-paris.fr}
\icmlcorrespondingauthor{Lirida Naviner}{lirida.naviner@telecom-paris.fr}
\icmlcorrespondingauthor{Jhony H. Giraldo}{jhony.giraldo@telecom-paris.fr}

\icmlkeywords{Machine Learning, Graph Generation, Hypergraph}

\vskip 0.3in
]

\printAffiliationsAndNotice{}

\begin{abstract}

Graph generative models perform well on small-scale structured data but struggle to scale to large, complex structures.
Hierarchical approaches improve scalability but often ignore node and edge features, which are critical in real-world applications.
In this paper, we propose \method~(\textbf{f}eature-\textbf{a}ware (\textbf{h}yper)graph generation via \textbf{ne}xt-\textbf{s}cale prediction), a hierarchical framework that jointly generates topology and features for graphs and hypergraphs.
\method~progressively constructs the final sample through localized expansion and refinement, guided by a novel node budget controlling granularity and ensuring cross-scale consistency. 
Experiments on synthetic, 3D mesh, and graph point cloud datasets demonstrate competitive or state-of-the-art performance while uniquely scaling to large-scale graphs and hypergraphs with features.
Our code is open source\footnote{\url{https://github.com/DorianGailhard/FAHNES}}.
\end{abstract}

\section{Introduction}

Graphs and hypergraphs capture complex relationships---pairwise in graphs and multi-way in hypergraphs---making them essential tools for representing and synthesizing structured data.
These structures are central to applications ranging from molecular design and materials discovery to electronic circuits and 3D shape modeling \citep{pmlr-v97-kajino19a,molecularbiology, luo2024dehnneffectiveneuralmodel}.
Generating such discrete geometric structures is therefore an important challenge in machine learning \citep{zhu2022survey}.
In most real-world settings, topology alone is insufficient: node and edge (or hyperedge) features often carry important semantic or geometric information, such as node coordinates in 3D meshes.

Despite their success, existing methods for featured graph generation struggle to scale.
Most of these approaches use flat architectures that generate the entire structure at once, leading to quadratic computational and memory complexities \citep{vignac2023digress,eijkelboom2024variationalflowmatchinggraph,xu2024discretestatecontinuoustimediffusiongraph} and limiting their use to moderately sized graphs. 
To address scalability, hierarchical generation strategies have been proposed for both graphs and hypergraphs \citep{bergmeister2024efficient,gailhard2024hygenediffusionbasedhypergraphgeneration}.
These methods build the structure in stages, starting from a coarse representation and progressively upsampling and refining it, which allows them to handle much larger topologies.
However, they focus only on unfeatured structures.

\begin{figure}[t]
    \centering
    \includegraphics[width=\columnwidth]{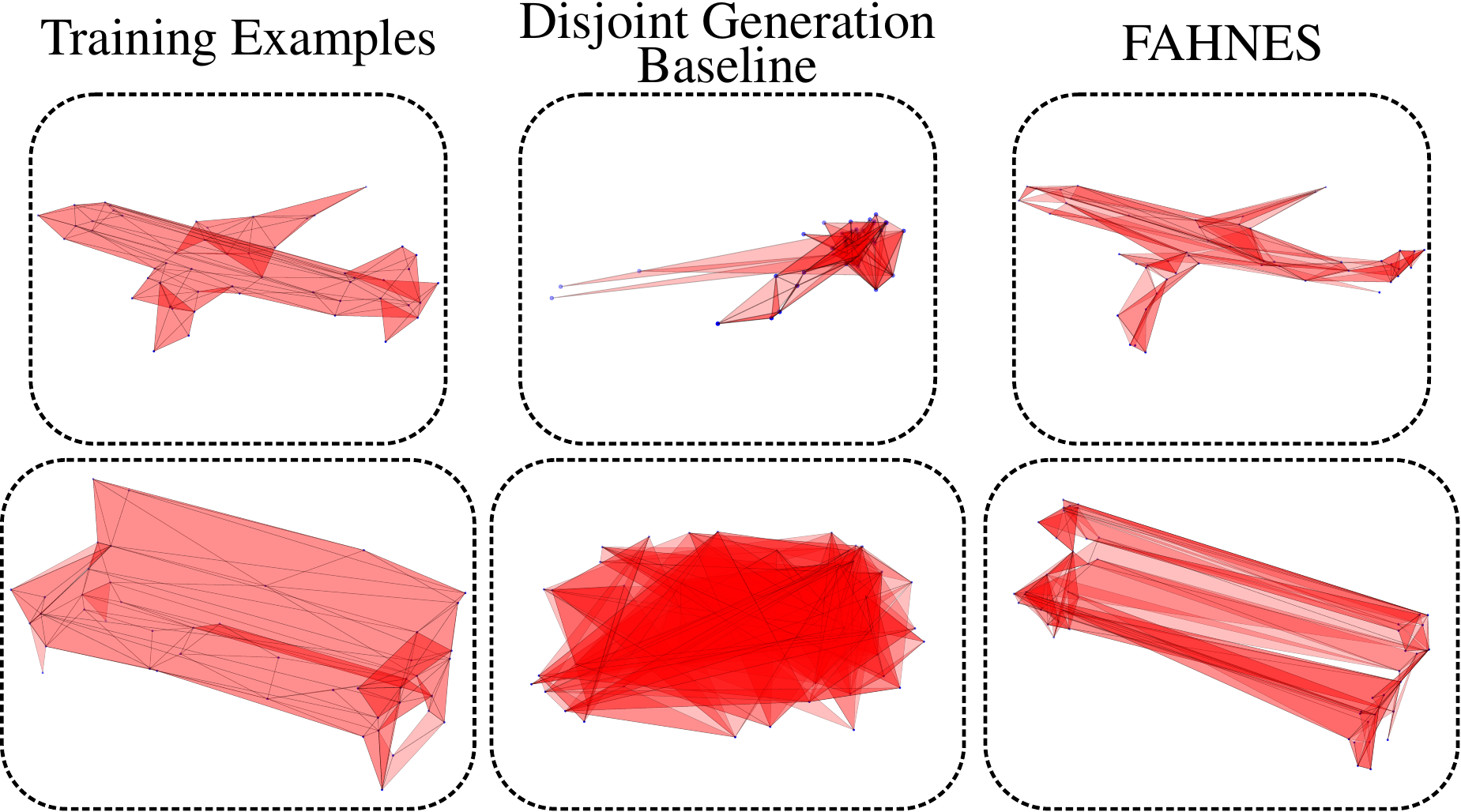}
    \caption{Examples of featured hypergraphs generated by a sequential disjoint generation baseline and our model (\method).}
    \label{fig:teaser}
\end{figure}

Extending them to generate features is challenging as different regions of a graph or hypergraph grow at different rates during the refinement process, making it difficult to maintain consistency across scales.
Furthermore, sequentially generating topology first and then features---rather than modeling them jointly---is ineffective in complex settings, as illustrated in Fig. \ref{fig:teaser}.

To overcome these limitations, we propose \method~(\textbf{f}eature-\textbf{a}ware (\textbf{h}yper)graph generation via \textbf{ne}xt-\textbf{s}cale prediction), a hierarchical generative framework that jointly models topology and features for both graphs and hypergraphs.\footnote{As hypergraphs generalize graphs, we present \method~for hypergraphs; the methodology is then easily adapted for graphs.}
\method~builds multi-scale representations \citep{TianVAR, bergmeister2024efficient} through node coarsening and reconstructs fine structures via localized expansion, while directly predicting features alongside topology at each stage.
A fundamental component of our model is a \textit{node budget}, which encodes local growth constraints and helps maintain consistency across regions that expand at different rates.
In addition, we propose a \textit{multi-scale graph optimal-transport (OT) coupling}, which generalizes the minibatch OT coupling \citep{tong2024improvinggeneralizingflowbasedgenerative,pooladian2023multisampleflowmatchingstraightening}
to our hierarchical graph setting, improving training stability and coherence during generation.
Together, these components enable scalable generation of large, featured graphs and hypergraphs across diverse domains, from 3D meshes to point clouds.
Our main contributions are:

\begin{itemize}[leftmargin=0.5cm]
    \setlength\itemsep{0pt}
    \item We introduce the first hierarchical generative method for graphs and hypergraphs capable of jointly generating topology and features with quasi-linear computational complexity (Section~\ref{sec:overview}).
    \item We propose a novel node budget that enhances global structural coherence in hierarchical graph and hypergraph generation (Sections~\ref{sec:coarsening} and \ref{sec:expansion}).
    \item To address the permutation-alignment problem in graphs, we propose a theoretically grounded multi-scale OT coupling for graph and hypergraph generation (Section~\ref{sec:ot_coupling}).
    \item We validate \method~on both synthetic and real-world datasets, demonstrating strong performance in jointly generating topology and features (Section~\ref{sec:experiments}).
\end{itemize}


\section{Related Work}

\textbf{Graph and hypergraph generation with deep learning}.
Graph generation has seen significant progress in recent years. Early approaches, such as GraphVAE \citep{simonovsky2018graphvaegenerationsmallgraphs}, used autoencoders to embed graphs into latent spaces for sampling. Subsequent models employed recurrent neural networks to sequentially generate adjacency matrices \citep{you2018graphrnngeneratingrealisticgraphs}. More recently, diffusion-based methods enabled permutation-invariant graph generation \citep{niu2020permutationinvariantgraphgeneration, vignac2023digress}, with extensions incorporating structural priors such as node degrees \citep{chen2023efficientdegreeguidedgraphgeneration}.
Many prior methods jointly model node features and topology but remain limited to small graphs due to scalability issues. In particular, diffusion-based models \citep{vignac2023digress, eijkelboom2024variationalflowmatchinggraph, xu2024discretestatecontinuoustimediffusiongraph} operate on complete graphs, corrupting topology and features and learning to denoise them. This limits scalability due to the quadratic number of possible edges.

To mitigate this, hierarchical methods have been proposed.
\citet{bergmeister2024efficient} introduced a scalable graph generation framework based on a coarsen-then-expand approach, merging nodes to form coarse representations and progressively reconstructing finer details.
This framework was extended to hypergraphs by \citet{gailhard2024hygenediffusionbasedhypergraphgeneration}, allowing the generation of higher-order relations involving more than two nodes.
However, both methods focus only on topology generation, neglecting node and edge or hyperedge features that are essential for many applications.

Hierarchical graph generation has also been explored in molecular modeling \citep{qiang2023coarse, kuznetsov2021molgrow, zhu2023molhfhierarchicalnormalizingflow}.
However, these methods are highly tailored to the molecular domain and are difficult to extend to large graphs.
In contrast, our approach jointly models features and connectivity at all scales and is not restricted to molecular generation.

\textbf{Applications of hypergraph generation}.
Generative models that jointly capture higher-order structure and associated features are relevant to many domains beyond conventional graphs.
For example, 3D shape modeling involves surfaces (\eg triangles, quads, or polygons) that go beyond pairwise connectivity. 
Existing methods often rely on fixed topology, quantization, or autoregressive transformer-based modeling \citep{nash2020polygenautoregressivegenerativemodel, siddiqui2023meshgptgeneratingtrianglemeshes}, introducing limitations in flexibility and scalability.
Hypergraphs offer a general representation by treating faces as hyperedges, naturally capturing both geometry and higher-order connectivity.

Unlike prior work that focuses solely on topology or relies on flat, non-scalable designs, we present the first unified, scalable approach for jointly modeling topology and features for graphs and hypergraphs.
\method~enhances hierarchical generative modeling with feature integration, node budget, and multi-scale graph OT coupling.

\begin{figure*}[t]
    \centering
    \includegraphics[width=0.97\textwidth]{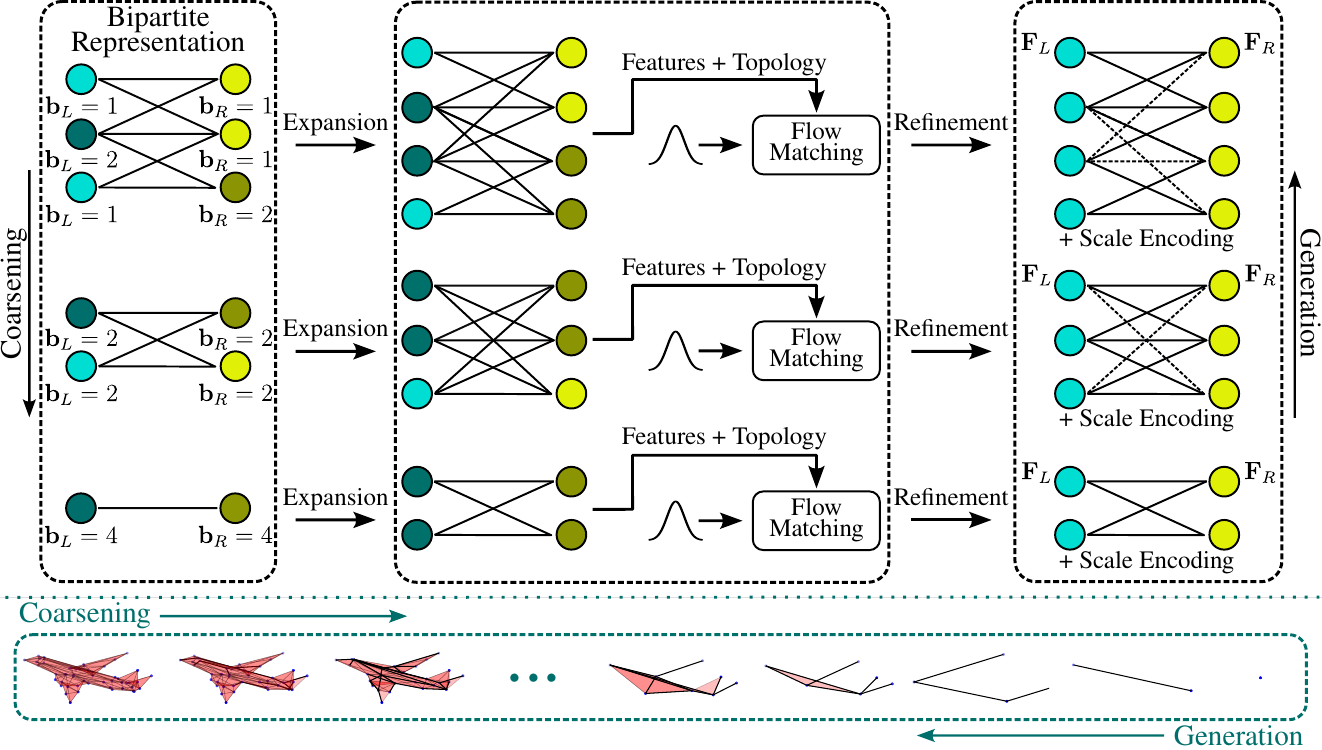}
    \caption{Our framework adopts a \textit{coarsening-expansion} strategy.
    \textit{i}) During training, input hypergraphs are progressively coarsened by merging nodes and hyperedges, yielding a multi-scale representation.
    Node features are averaged during merging, and node budgets are summed.
    \textit{ii}) The model is then trained to predict which nodes were merged at each scale.
    \textit{iii}) In the \textit{expansion} phase, merged nodes (shown in dark in the leftmost column) are expanded back (copies shown in dark), inheriting their parent’s features, node budget, and connectivity.
    In the \textit{refinement} phase, the model is trained to (\textit{a}) identify which edges should be removed (dotted lines), (\textit{b}) predict the node budgets of the children based on the parent's node budget, and (\textit{c}) refine the features of the newly expanded nodes.
    }
    \label{fig:pipeline}
\end{figure*}

\section{Feature-Aware (Hyper)graph Generation via Next‑Scale Prediction}
\label{sec:method}

\noindent \textbf{Notations}.
We use calligraphic letters (\eg $\mathcal{V}$) for sets, with cardinality $\vert \mathcal{V} \vert$.
Bold uppercase letters denote matrices (\eg $\mathbf{A}$), and bold lowercase letters denote vectors (\eg $\mathbf{X}$).
The transpose and element-wise multiplication are written $(\cdot)^{\top}$ and $\odot$, respectively.
$\lfloor \cdot \rceil$ denotes rounding to the nearest integer, and $\operatorname{Id}(x)=x$ is the identity function.

\noindent \textbf{Basic definitions}.
We define a graph \( G = (\mathcal{V}, \mathcal{E}) \) as a pair consisting of a set of vertices \( \mathcal{V} \) and a set of edges \( \mathcal{E} \subseteq \mathcal{V} \times \mathcal{V} \). 
Graphs may also carry node and edge features, represented by matrices \( \mathbf{F}_{\mathcal{V}} \in \mathbb{R}^{|\mathcal{V}| \times m} \) and \( \mathbf{F}_{\mathcal{E}} \in \mathbb{R}^{|\mathcal{E}| \times l} \), where \( m \) and \( l \) denote the dimensions of the features.
A bipartite graph $B=(\mathcal V_L,\mathcal V_R,\mathcal E)$ consists of two disjoint vertex sets $\mathcal V_L$ and $\mathcal V_R$, with edges only between the two sets, \ie \( \mathcal{E} \subseteq \mathcal{V}_L \times \mathcal{V}_R \). 
The full set of nodes is \( \mathcal{V} = \mathcal{V}_L \cup \mathcal{V}_R \). 
In this work, we consider node features for bipartite graphs, denoted by \( \mathbf{F}_L \) for the left-side nodes and \( \mathbf{F}_R \) for the right-side nodes.

A hypergraph $H$ extends the concept of a graph and is specified by a pair $(\mathcal{V}, \mathcal{E})$, where $\mathcal{V}$ represents the vertex set and $\mathcal{E}$ contains hyperedges, with each $e \in \mathcal{E}$ being a subset of $\mathcal{V}$.
Similar to graphs, hypergraphs may have node and hyperedge features, which are denoted $\mathbf{F}_{\mathcal{V}}$ and $\mathbf{F}_{\mathcal{E}}$.
We consider two fundamental graph-based representations of hypergraphs: clique and star expansions.
The clique expansion transforms a hypergraph $H$ into a graph $C = (\mathcal{V}, \mathcal{E}_c)$, where hyperedges are replaced by cliques, \ie $\mathcal{E}_c = \{(u, v) \mid \exists~e \in \mathcal{E} : u, v \in e\}$.
The star expansion converts a hypergraph $H$ into a bipartite structure $B = (\mathcal{V}_L, \mathcal{V}_R, \mathcal{E}_b)$, where $\mathcal{V}_L = \mathcal{V}$, $\mathcal{V}_R = \mathcal{E}$, and $\mathcal{E}_b = \{(v, e) \mid v \in \mathcal{V}_L, e \in \mathcal{V}_R, v \in e \text{ in } H\}$.
Left side nodes $\mathcal{V}_L$ correspond to the nodes of the hypergraph, while right side nodes $\mathcal{V}_R$ represent hyperedges.

Our objective is to develop a generative model that can sample from the underlying distribution of a dataset of featured hypergraphs (or graphs) $(H_1, \dots, H_N)$, \ie to learn the joint distribution over both topology and features.
We present \method~in the more general setting of hypergraphs; the method can be easily adapted to graphs, as discussed in Section~\ref{sec:generalization_graphs}.
Complete mathematical proofs of all propositions in this paper appear in Appendix~\ref{app:proofs}.

\subsection{Overview} \label{sec:overview}

\method~follows a hierarchical pipeline (Fig.~\ref{fig:pipeline}):
\textit{i}) during training, multi-scale representations of input (hyper)graphs are constructed; \textit{ii}) a model learns to reconstruct higher scales from lower ones.
We use a coarsening–expansion framework \citep{bergmeister2024efficient, gailhard2024hygenediffusionbasedhypergraphgeneration}.
Multi-scale representations are obtained via spectrum-preserving coarsening \citep{loukas2018graph} on the clique expansion of the hypergraph.
The process is then reversed through learned expansion and refinement, reconstructing topology and features at progressively finer scales.

To address uneven growth across regions, we introduce a \textit{node budget}: each cluster is assigned a node budget indicating its remaining expansions. 
Node budgets are recursively divided among child nodes, starting from a single super-node with the full budget and ending when all clusters have a budget of one. 
This provides more local control over the final node count and improves global consistency compared to prior methods \citep{bergmeister2024efficient,gailhard2024hygenediffusionbasedhypergraphgeneration}, which only append the desired size to node embeddings.
At each expansion step, the model also predicts the mean feature of future child nodes conditioned on the parent’s feature, extending the scale-wise autoregressive idea in \cite{ren2024flowarscalewiseautoregressiveimage} so that predictions at one scale guide those at the next.

\subsection{Coarsening with node budget}
\label{sec:coarsening}

We adopt the coarsening strategy in \cite{gailhard2024hygenediffusionbasedhypergraphgeneration}, representing hypergraphs at multiple resolutions by merging nodes into clusters.
Node pairs are merged via spectrum-preserving coarsening \citep{loukas2018graph} applied to the hypergraph's clique expansion.
These mergings are mirrored in the bipartite representation by merging left-side nodes and collapsing duplicate hyperedges.
We restrict node mergings to at most two per cluster, which bounds hyperedge mergings to at most three \citep{gailhard2024hygenediffusionbasedhypergraphgeneration}.
For each resulting cluster, we track two quantities.
First, the \textit{node budget} equals the number of nodes in the cluster: it is initialized to $1$ for each node and hyperedge, summed upon merging, and assigned to the resulting super-cluster.
This component constrains local growth by specifying how many fine-level nodes a coarse cluster should expand into.
Second, the super-cluster feature is computed as the weighted mean of its constituent features, representing the average feature of the contained nodes.

We denote node (left‑side) budgets and features by $\mathbf{b}_L \in \mathbb{N}^{\vert \mathcal{V}_L \vert}$ and $\mathbf{F}_L$, and hyperedge (right‑side) budgets and features by $\mathbf{b}_R \in \mathbb{N}^{\vert \mathcal{V}_R \vert}$ and $\mathbf{F}_R$.
\begin{definition}[Bipartite graph coarsening]
\label{def::bipartite_coarsening_featured}
Let $H$ be a hypergraph, $C = (\mathcal{V}_c, \mathcal{E}_c)$ its clique expansion, and $B = (\mathcal{V}_L, \mathcal{V}_R, \mathcal{E}_b, \mathbf{b}_L, \mathbf{b}_R, \mathbf{F}_L, \mathbf{F}_R)$ its \textit{featured} bipartite representation. 
Let $\mathcal{P}_L = \{\mathcal{V}^{(1)}, \ldots, \mathcal{V}^{(n)}\}$ be a partitioning\footnote{That is, $\mathcal{V}^{(p)} \subseteq \mathcal{V}_L$, $\bigcup_{i=1}^{{n}} \mathcal{V}^{(i)} = \mathcal{V}_L$, and ${\mathcal{V}^{(i)} \cap \mathcal{V}^{(j)} = \emptyset~\forall~1\leq i,j \leq n}$.} of the node set $\mathcal{V}_L$ such that each set $\mathcal{V}^{(p)}$ induces a connected subgraph in $C$.
We construct an intermediate coarsening $\tilde{B}(B, \mathcal{P}_L) = (\bar{\mathcal{V}}_L, \mathcal{V}_R, \bar{\mathcal{E}}_b, \bar{\mathbf{b}}_L, \mathbf{b}_R, \bar{\mathbf{F}}_L, \mathbf{F}_R)$ by merging each $\mathcal{V}^{(p)}$ into a single node $v^{(p)} \in \bar{\mathcal{V}}_L$, and by defining:
\begin{equation}
    \bar{\mathbf{b}}_L[p] = \sum_{v \in \mathcal{V}^{(p)}} \mathbf{b}_L[v], \, \bar{\mathbf{F}}_L[p] = \frac{1}{ \bar{\mathbf{b}}_L[p] } \sum_{v \in \mathcal{V}^{(p)}} \mathbf{b}_L[v] \mathbf{F}_L[v],
\end{equation}
for every merged node $\mathcal{V}^{(p)}$.
An edge $e_{\{p,q\}} \in \bar{\mathcal{E}}_b$ is added between $v^{(p)} \in \bar{\mathcal{V}}_L$ and $v^{(q)} \in \mathcal{V}_R$ if there exists an edge $e_{\{i, q\}} \in \mathcal{E}_b$ in the original bipartite representation between some $v^{(i)} \in \mathcal{V}^{(p)}$ and $v^{(q)}$.

To complete the coarsening process, we define an equivalence relation $v_1 \sim v_2 \iff \mathcal{N}(v_1) = \mathcal{N}(v_2)$ on $\mathcal{V}_R$, where $\mathcal{N}(v)$ denotes the set of neighbors of $v$, \ie we consider right side nodes having the same set of neighbors as equivalent, or in other words we consider hyperedges containing the same set of nodes as equivalent.
This induces a partitioning $\mathcal{P}_R = \{\mathcal{V}_R^{(1)}, \dots, \mathcal{V}_R^{(m)}\}$, allowing us to construct the fully coarsened bipartite representation $\bar{B}(\tilde{B}, \mathcal{P}_L) = (\bar{\mathcal{V}}_L, \bar{\mathcal{V}}_R, \bar{\mathcal{E}}_b, \bar{\mathbf{b}}_L, \bar{\mathbf{b}}_R, \bar{\mathbf{F}}_L, \bar{\mathbf{F}}_R)$ by merging each part $\mathcal{V}_R^{(p)}$ into a single node $v_R^{(p)} \in \bar{\mathcal{V}}_R$, similarly to the construction of $\bar{\mathcal{V}}_L$, and by defining: 
\begin{equation}
    \bar{\mathbf{b}}_R[p] = \sum_{v \in \mathcal{V}_R^{(p)}} \mathbf{b}_R[v], \, \bar{\mathbf{F}}_R[p] = \frac{1}{ \bar{\mathbf{b}}_R[p] } \sum_{v \in \mathcal{V}_R^{(p)}} \mathbf{b}_R[v] \mathbf{F}_R[v],
\end{equation}
for every merged hyperedge $\mathcal{V}_R^{(p)}$.
\end{definition}

\begin{remark}
\label{rem:coarsening_pipeline}
Informally, we cluster nodes in the clique expansion, merge the corresponding left-side nodes of the bipartite graph, and compute their node budgets and features as described above.
Right-side nodes (hyperedges) connected to the same left-side nodes are then merged.
In our implementation, we select nodes for clustering using spectrum-preserving coarsening \citep{loukas2018graph}.
In the final reduction---when only one node and one hyperedge remain---we replace their features with zero matrices to obtain a feature-agnostic initialization for generation.
Please note that cluster size vectors introduced in \citep{bergmeister2024efficient} correspond to the sizes of all sets $\mathcal{V}_L^{(p)}$ and $\mathcal{V}_R^{(p)}$, \ie the number of mergings for each cluster at this specific step, and not the number of nodes absorbed since the initialization of coarsening---which we call \textit{node budget}.
Further details about the coarsening methodology are provided in Appendix \ref{app::coarsening}.
Some visual examples of coarsening sequences are provided in Appendix \ref{app:coarsening_example}.
\end{remark}

We provide the following theoretical result regarding the feature merging process in \method.
\begin{proposition}
Let $\mathbf{X}$ be the node feature matrix and $\mathcal{P} = \{\mathcal{C}_k\}_{k=1}^K$ a partition of the vertices into clusters. The optimal downsampled features minimizing the reconstruction error are the cluster-wise barycenters:
\begin{equation}
\mathbf{X}^*_{\mathcal{C}_k} = \frac{1}{|\mathcal{C}_k|} \sum_{v \in \mathcal{C}_k} \mathbf{X}_v \quad \text{for $1 \leq k \leq K$}.
\end{equation}
\end{proposition}

\subsection{Expansion and Refinement with node budget}
\label{sec:expansion}

Once multi-scale representations are obtained via coarsening, a model is trained to reverse this process.
We define the inverse of coarsening as a sequence of two stages: expansion (upsampling by duplicating nodes and hyperedges) and refinement (adjusting connectivity, node budgets, and features).
This process operates exclusively on the bipartite representation.

Since our framework only conditions on the total number of nodes in the final hypergraph, we maintain a single node budget vector $\mathbf{b} \in \mathbb{N}^{\vert \mathcal{V}_L \vert}$ for the node (left-side) partition and discard the hyperedge (right-side) node budget. 
To undo mergings, clusters are recursively split into multiple child nodes.
Each child initially copies the parent's connections and inherits its feature and node budget, before a subsequent refinement step then: \textit{i}) redistributes the node budget among the children, \textit{ii}) updates their features, and \textit{iii}) removes edges that should not persist at the finer scale.
\begin{definition}[Bipartite graph expansion] \label{def::expansion}
    Given a bipartite graph $B = (\mathcal{V}_L, \mathcal{V}_R, \mathcal{E}, \mathbf{b}, \mathbf{F}_L, \mathbf{F}_R)$ and expansion size vectors $\mathbf{v}_L \in \mathbb{N}^{|\mathcal{V}_L|}$, $\mathbf{v}_R \in \mathbb{N}^{|\mathcal{V}_R|}$, denoting the number of duplications for nodes (left side) and hyperedges (right side), respectively.
    Let $\tilde{B}(B, \mathbf{v}_L, \mathbf{v}_R) = (\tilde{\mathcal{V}}_L, \tilde{\mathcal{V}}_R, \tilde{\mathcal{E}}, \mathbf{b}^{\text{expanded}}, \mathbf{F}_L^{\text{expanded}}, \mathbf{F}_R^{\text{expanded}})$ denote the expansion of $B$, whose node sets, node budgets, and features are given by:
    \begin{equation}
        \begin{split}
            & \tilde{\mathcal{V}}_L = \bigcup_{p=1}^{|\mathcal V_L|}\mathcal V_L^{(p)}, \mathcal{V}_L^{(p)} = \{\,v_L^{(p,i)}\,\}_{i=1}^{\mathbf v_L[p]} \text{~~for~~} 1 \leq p \leq |\mathcal{V}_L|, \\
            & \tilde{\mathcal V}_R = \bigcup_{p=1}^{|\mathcal V_R|}\mathcal V_R^{(p)}, \mathcal V_R^{(p)} = \{\,v_R^{(p,i)}\,\}_{i=1}^{\mathbf v_R[p]} \text{~~for~~} 1 \leq p \leq |\mathcal{V}_R|,\\
            & \mathbf{b}^{\text{expanded}}[p, i] = \mathbf{b}[p] \text{~~for~~} 1 \leq i \leq \mathbf{v}_L[p], 1 \leq p \leq |\mathcal{V}_L|, \\
            & \mathbf{F}_L^{\text{expanded}}[p, i] = \mathbf{F}_L[p] \text{~~for~~} 1 \leq i \leq \mathbf{v}_L[p], 1 \leq p \leq |\mathcal{V}_L|, \\
            & \mathbf{F}_R^{\text{expanded}}[p, i] = \mathbf{F}_R[p] \text{~~for~~} 1 \leq i \leq \mathbf{v}_R[p], 1 \leq p \leq |\mathcal{V}_R|.
        \end{split}
    \end{equation}
    The edge set $\tilde{\mathcal{E}}$ includes all the cluster interconnecting edges:
    $\{e_{\{p, i; q, j\}} \mid e_{\{p, q\}} \in \mathcal{E}, v_L^{(p, i)} \in \mathcal{V}_L^{(p)}, v_R^{(q, j)} \in \mathcal{V}_R^{(q)}\}$.
\end{definition}

\begin{remark}
    Expansion thus acts as a duplication operation: vertices and hyperedges are copied, with each child inheriting all incident edges of its parent.
\end{remark}

\begin{definition}[Bipartite graph refinement] \label{def::refinement}
    Given a bipartite graph $\tilde{B} = (\tilde{\mathcal{V}}_L, \tilde{\mathcal{V}}_R, \tilde{\mathcal{E}}, \mathbf{b}, \mathbf{F}_L^{\text{expanded}}, \mathbf{F}_R^{\text{expanded}})$, an edge selection vector $\mathbf{e} \in \{0, 1\}^{|\tilde{\mathcal{E}}|}$, a node budget split vector $\mathbf{f} \in [0, 1]^{|\tilde{\mathcal{V}}_L|}$, satisfying $\sum_{v \in \mathcal{V}_L^{(p)}} \mathbf{f}[v] = 1$ for all cluster $\mathcal{V}_L^{(p)}$ in $\tilde{\mathcal{V}}_L$, and two feature refinement vectors $\mathbf{F}_L^{\text{refine}}$ and $\mathbf{F}_R^{\text{refine}}$ with the same dimensions as $\mathbf{F}_L^{\text{expanded}}$ and $\mathbf{F}_R^{\text{expanded}}$, let $B(\tilde{B}, \mathbf{e}, \mathbf{f}, \mathbf{F}_L^{\text{refine}}, \mathbf{F}_R^{\text{refine}}) = (\tilde{\mathcal{V}}_L, \tilde{\mathcal{V}}_R, \mathcal{E}, \lfloor \mathbf{b} \odot \mathbf{f} \rceil, \mathbf{F}_L^{\text{refine}}, \mathbf{F}_R^{\text{refine}})$ denote the refinement of $\tilde{B}$, where $\mathcal{E} \subseteq \tilde{\mathcal{E}}$ such that the $i$-th edge $e_{(i)} \in \mathcal{E}$ if and only if $\mathbf{e}[i] = 1$.
\end{definition}

\begin{remark}
    Edges are selectively removed based on the binary indicator vector $\mathbf{e}$, and features are updated with new predictions.
    Node budgets are divided among child nodes according to the split proportions specified by the vector $\mathbf{f}$.
    Since node budgets must be integers, the resulting values are rounded across clusters using the largest remainder method.   
\end{remark}

\subsection{Probabilistic Modeling}
\label{sec:prob}

We now present a formalization of our learning framework, generalizing \citep{bergmeister2024efficient,gailhard2024hygenediffusionbasedhypergraphgeneration}.
Let $\{H^{(1)}, \ldots, H^{(N)}\}$ denote a set of \textit{i.i.d.} hypergraph instances. Our objective is to approximate the unknown generative process by learning a distribution $p(H)$. We model the marginal likelihood of each hypergraph $H$ as a sum over the likelihoods of its bipartite representation's expansion sequences $p(H) = p(B) = \sum_{\varpi \in \Pi(B)} p(\varpi)$,
where $\Pi(B)$ denotes the set of valid expansion sequences from a minimal bipartite graph to the full bipartite representation $B$ corresponding to $H$. Each intermediate $B^{(l-1)}$ is generated by expanding and refining its predecessor, in accordance with Definitions \ref{def::expansion} and \ref{def::refinement}.

To simplify notations, we drop the superscript for $\mathbf{F}^{\text{refine}}$ and simply write $\mathbf{F}$.
Assuming a Markovian generative structure, and after various assumptions, we model the distribution of expansion and refinement sequences as:
\begin{equation} \label{eq:target_learning}
\begin{aligned}
    p(\mathbf{v}_L^{(l)}, \mathbf{v}_R^{(l)} | B^{(l)}) & p(\mathbf{e}^{(l)}, \mathbf{f}^{(l)}, \mathbf{F}_L^{(l)}, \mathbf{F}_R^{(l)} | \tilde{B}^{(l)}) = \\
    & p(\mathbf{v}_L^{(l)}, \mathbf{v}_R^{(l)}, \mathbf{e}^{(l)}, \mathbf{f}^{(l)}, \mathbf{F}_L^{(l)}, \mathbf{F}_R^{(l)} | \tilde{B}^{(l)}),
\end{aligned}
\end{equation}
which corresponds to the combined expansion and refinement step that inverts coarsening at depth $l$.
The detailed rationale for this choice can be found in Appendix \ref{app:prob_modeling}.

During training, the model approximates the right-hand side of~\eqref{eq:target_learning}, 
learning to generate, for each expansion and refinement step, 
$\mathbf{v}_L$, $\mathbf{v}_R$, $\mathbf{e}$, $\mathbf{f}$, and 
$\mathbf{F}_L^{\text{refine}}, \mathbf{F}_R^{\text{refine}}$. 
Once trained, the model generates a hypergraph with $N$ nodes through successive expansion and refinement steps.
\begin{enumerate}[leftmargin=0.5cm]
    \itemsep0em
    \item \textit{Initialization}. Start from a minimal bipartite graph: $B^{(L)} = ( \{1\}, \{2\}, \{(1, 2)\}, (N))$, consisting of a single node on each side connected by one edge. The left-side node is assigned the full node budget, \ie $\mathbf{b} = [N]$.
    If node and hyperedge features need to be generated, $\mathbf{F}_L$ and $\mathbf{F}_R$ are initialized as zero matrices.
    \item \textit{Expansion and refinement}. Iteratively expand and refine the bipartite representation to add details until the target size is reached: $B^{(l)} \xrightarrow{\text{expand}} \tilde{B}^{(l-1)} \xrightarrow{\text{refine}} B^{(l-1)}$.
    \item \textit{Hypergraph reconstruction}. Once the final bipartite graph is generated, construct the hypergraph by collapsing each right-side node into a hyperedge connecting its adjacent left-side nodes.
\end{enumerate}

Full details of the sampling procedure are provided in Algorithm \ref{alg:sample_deterministic} in Appendix \ref{app::training_sampling}.

\subsection{Multi-scale graph OT coupling} \label{sec:ot_coupling}

Graphs and hypergraphs lack a canonical node ordering, making it challenging to align predictions with their targets.
After expansion, the model may generate a permuted version of the correct output, leading to a large loss despite an otherwise correct prediction and introducing noise into the learning signal.
To address this, we generalize \emph{minibatch OT coupling}~\citep{tong2024improvinggeneralizingflowbasedgenerative,pooladian2023multisampleflowmatchingstraightening} to align predictions and targets through a permutation that minimizes their matching cost.
Unlike standard minibatch OT coupling, which matches samples across a batch, our approach performs alignment within each graph by matching its constituent nodes.

Formally, for a given bipartite graph with $n$ nodes, adjacency matrix $\mathbf{A} \in \{0, 1\}^{n \times n}$, expanded node budgets $\mathbf{b} \in \mathbb{N}^n$ and expanded features $\mathbf{F} \in \mathbb{R}^{n \times d}$ where $d$ is the dimension of node or hyperedge features, $\mathbf{X} \in \mathbb{R}^{B\times d}$ and $\mathbf{Y} \in \mathbb{R}^{B\times d}$ correspond to prior samples and targets, respectively, the multi-scale graph OT coupling leads to the following optimization problem:
\begin{equation}\label{eq:perm_opt}
\begin{aligned}
\mathbf P^\ast
&= \argmin_{\mathbf P \in \Pi_n} \;\|\mathbf P\mathbf X - \mathbf Y\|^2 \\
\text{s.t.}\quad
&\mathbf P^\top \mathbf A \mathbf P = \mathbf A,\;
\mathbf P \mathbf b = \mathbf b,\;
\mathbf P \mathbf F = \mathbf F,
\end{aligned}
\end{equation}
where \(\mathbf{P}\) is a permutation matrix and \(\Pi_n\) denotes the set of all \(n\times n\) permutations.
To reduce overhead, we restrict permutations to children within a single cluster expansion, where equivalence naturally holds.
With two or three children per cluster, only two or six permutations are possible, making the operation lightweight and easily parallelizable with standard tensor operations.
We optimize \eqref{eq:perm_opt} using Algorithm \ref{alg:minibatch_ot} in Appendix \ref{sec:additional_tricks}.

\begin{proposition}
\label{prop:marginal_preservation}
Under the multi-scale graph OT coupling, the joint distribution $q(\mathbf{X}_0, \mathbf{X}_1)$ has marginals $q_0(\mathbf{X}_0), q_1(\mathbf{X}_1)$.
\end{proposition}

Proposition \ref{prop:marginal_preservation} ensures that the multi-scale graph OT coupling does not introduce bias in the learned distribution.
\begin{proposition}
\label{prop:variance_reduction}
Let $\mathcal{B} = \{(\mathbf{X}^0_i, \mathbf{X}^1_i)\}_{i=1}^N$ be a minibatch of size $N$ coupled by a permutation $\pi$.
Let $\Delta_\pi$ be the random variable representing the displacement vector $\mathbf{X}^1_i - \mathbf{X}^0_{\pi(i)}$ for an index $i$ drawn uniformly from the batch.
Multi-scale graph OT coupling (Algorithm~\ref{alg:minibatch_ot}) yields a permutation $\pi^*$ that reduces the total variance of this displacement:
\[
\tr(\operatorname{Var}(\Delta_{\pi^*})) \leq \tr(\operatorname{Var}(\Delta_{Id})).
\]
\end{proposition}
Proposition \ref{prop:variance_reduction} shows that multi-scale graph OT coupling reduces the variance of the target displacements compared with uncoupled targets, providing a more consistent learning signal.

In practice, restricting permutations to local cluster expansions keeps the additional computational overhead of multi-scale graph OT coupling negligible (approximately $4\%$).

\begin{table*}[t]
    \centering
    \caption{Evaluation for hypergraph datasets: SBM, Ego, Tree, ModelNet40 Bookshelf, and ModelNet40 Piano.}
    \label{tab:hypergraph_results}
    \setlength{\tabcolsep}{2pt}
    \resizebox{\textwidth}{!}{
    \begin{tabular}{@{}l cc cc cc cc cc@{}}
    \toprule
    \multirow{2}{*}{\textbf{Method}} & \multicolumn{2}{c}{\textbf{SBM Hypergraphs}} & \multicolumn{2}{c}{\textbf{Ego Hypergraphs}} & \multicolumn{2}{c}{\textbf{Tree Hypergraphs}} & \multicolumn{2}{c}{\textbf{ModelNet40 Bookshelf}} & \multicolumn{2}{c}{\textbf{ModelNet40 Piano}} \\
    & \multicolumn{2}{c}{($n_{avg} = 31.73, std = 0.55$)} & \multicolumn{2}{c}{($n_{avg} = 109.71, std = 10.23$)} & \multicolumn{2}{c}{($n_{avg} = 32, std = 0$)} & \multicolumn{2}{c}{($n_{avg} = 119.38, std = 68.20$)} & \multicolumn{2}{c}{($n_{avg} = 177.29, std = 57.11$)} \\
    \cmidrule(lr){2-3} \cmidrule(lr){4-5} \cmidrule(lr){6-7} \cmidrule(lr){8-9} \cmidrule(lr){10-11}
    & \textbf{Valid} (\%) $\uparrow$ & \textbf{Spectral} $\downarrow$ & \textbf{Valid} (\%) $\uparrow$ & \textbf{Spectral} $\downarrow$ & \textbf{Valid} (\%) $\uparrow$ & \textbf{Spectral} $\downarrow$ & \textbf{Node Num} $\downarrow$ & \textbf{Spectral} $\downarrow$ & \textbf{Node Num} $\downarrow$ & \textbf{Spectral} $\downarrow$ \\
    \midrule
    HyperPA & $2.5$ & $0.273$ & $0.0$ & $0.237$ & $0.0$ & $0.159$ & $8.025$ & $\underline{0.048}$ & $0.825$ & $\underline{0.067}$ \\
    VAE & $0.0$ & $0.024$ & $0.0$ & $0.133$ & $0.0$ & $0.124$ & $47.450$ & $0.190$ & $75.350$ & $0.396$ \\
    GAN & $0.0$ & $0.059$ & $0.0$ & $0.230$ & $0.0$ & $0.089$ & $\mathbf{0.000}$ & $0.476$ & $\mathbf{0.000}$ & $0.697$ \\
    Diffusion & $0.0$ & $0.031$ & $0.0$ & $0.190$ & $0.0$ & $0.127$ & $\mathbf{0.000}$ & $0.079$ & $\underline{0.050}$ & $0.113$ \\
    HYGENE & $\underline{65.0}$ & $\underline{0.010}$ & $\underline{90.0}$ & $\mathbf{0.004}$ & $\underline{77.5}$ & $\underline{0.012}$ & $69.730$ & $0.068$ & $42.520$ & $0.117$ \\
    \midrule
    \method & 
    $\mathbf{87.8 \vcenter{\hbox{\scriptsize $\boldsymbol{\pm 3.1}$}}}$ & 
    $\mathbf{0.006 \vcenter{\hbox{\scriptsize $\boldsymbol{\pm 0.004}$}}}$ & 
    $\mathbf{99.5 \vcenter{\hbox{\scriptsize $\boldsymbol{\pm 1.1}$}}}$ & 
    $\mathbf{0.004 \vcenter{\hbox{\scriptsize $\boldsymbol{\pm 0.003}$}}}$ & 
    $\mathbf{89.7 \vcenter{\hbox{\scriptsize $\boldsymbol{\pm 6.0}$}}}$ & 
    $\mathbf{0.003 \vcenter{\hbox{\scriptsize $\boldsymbol{\pm 0.002}$}}}$ & 
    $0.135 \vcenter{\hbox{\scriptsize $\pm 0.276$}}$ & 
    $\mathbf{0.024 \vcenter{\hbox{\scriptsize $\boldsymbol{\pm 0.015}$}}}$ & 
    $0.846 \vcenter{\hbox{\scriptsize $\pm 1.009$}}$ & 
    $\mathbf{0.040 \vcenter{\hbox{\scriptsize $\boldsymbol{\pm 0.026}$}}}$ \\
    \bottomrule
    \end{tabular}
    }
\end{table*}

\begin{table*}[t]
    \centering
    \caption{Evaluation for graph datasets: SBM, Tree, Planar, Protein, and Point Cloud.}
    \label{tab:graphs_results}
    \setlength{\tabcolsep}{3pt}
    \resizebox{\textwidth}{!}{
    \begin{tabular}{@{}l cc cc cc cc cc@{}}
    \toprule
    \multirow{2}{*}{\textbf{Method}} & \multicolumn{2}{c}{\textbf{SBM graphs}} & \multicolumn{2}{c}{\textbf{Tree graphs}} & \multicolumn{2}{c}{\textbf{Planar graphs}} & \multicolumn{2}{c}{\textbf{Protein}} & \multicolumn{2}{c}{\textbf{Point cloud}} \\
    & \multicolumn{2}{c}{($n_{avg} = 105.99, std = 38.38$)} & \multicolumn{2}{c}{($n_{avg} = 64.0, std = 0$)} & \multicolumn{2}{c}{($n_{avg} = 64.0, std = 0$)} & \multicolumn{2}{c}{($n_{avg} = 261.5, std = 104.7$)} & \multicolumn{2}{c}{($n_{avg} = 1434.3, std = 1285.9$)} \\
    \cmidrule(lr){2-3} \cmidrule(lr){4-5} \cmidrule(lr){6-7} \cmidrule(lr){8-9} \cmidrule(lr){10-11}
    & \textbf{Valid} (\%) $\uparrow$ & \textbf{Ratio} $\downarrow$ & \textbf{Valid} (\%) $\uparrow$ & \textbf{Ratio} $\downarrow$ & \textbf{Valid} (\%) $\uparrow$ & \textbf{Ratio} $\downarrow$ & \textbf{Spectral} $\downarrow$ & \textbf{Ratio} $\downarrow$ & \textbf{Spectral} $\downarrow$ & \textbf{Ratio} $\downarrow$ \\
    \midrule
    HSpectre & $45.0$ & $10.2$ & $\mathbf{100.0}$ & $4.0$ & $95.0$ & $\underline{2.1}$ & $\mathbf{0.001}$ & $\underline{5.9}$ & $\mathbf{0.005}$ & $\underline{7.0}$ \\
    BwR & $7.5$ & $38.6$ & $0.0$ & $11.4$ & $0.0$ & $251.9$ & $0.070$ & $254.4$ & $0.291$ & $133.2$ \\
    DiGress & $\underline{60.0}$ & $\mathbf{1.7}$ & $90.0$ & $\mathbf{1.6}$ & $77.5$ & $5.1$ & $0.002$ & $18.0$ & OOM & OOM \\
    DeFoG & $\mathbf{90.0 \vcenter{\hbox{\scriptsize $\boldsymbol{\pm 5.1}$}}}$ & $4.9 \vcenter{\hbox{\scriptsize $\pm 1.3$}}$ & $96.5 \vcenter{\hbox{\scriptsize $\pm 2.6$}}$ & $\mathbf{1.6 \vcenter{\hbox{\scriptsize $\boldsymbol{\pm 0.4}$}}}$ & $\mathbf{99.5 \vcenter{\hbox{\scriptsize $\boldsymbol{\pm 1.0}$}}}$ & $\mathbf{1.6 \vcenter{\hbox{\scriptsize $\boldsymbol{\pm 0.04}$}}}$ & --- & --- & --- & --- \\
    \midrule
    \method & 
    $50.0 \vcenter{\hbox{\scriptsize $\pm 5.0$}}$ & 
    $\underline{4.8 \vcenter{\hbox{\scriptsize $\pm 0.7$}}}$ & 
    $\mathbf{100.0 \vcenter{\hbox{\scriptsize $\boldsymbol{\pm 0.0}$}}}$ & 
    $\underline{1.8 \vcenter{\hbox{\scriptsize $\pm 0.8$}}}$ & 
    $\underline{96.7 \vcenter{\hbox{\scriptsize $\pm 6.2$}}}$ & 
    $2.2 \vcenter{\hbox{\scriptsize $\pm 1.6$}}$ & 
    $\mathbf{0.001 \vcenter{\hbox{\scriptsize $\boldsymbol{\pm 0.001}$}}}$ & 
    $\mathbf{3.8 \vcenter{\hbox{\scriptsize $\boldsymbol{\pm 1.9}$}}}$ & 
    $\mathbf{0.005 \vcenter{\hbox{\scriptsize $\boldsymbol{\pm 0.001}$}}}$ & 
    $\mathbf{3.2 \vcenter{\hbox{\scriptsize $\boldsymbol{\pm 0.5}$}}}$ \\
    \bottomrule
    \end{tabular}
    }
\end{table*}

\subsection{Generalization to Graphs} \label{sec:generalization_graphs}

Our framework can be directly extended to generate standard graphs by adapting the coarsen–expand procedure in \cite{bergmeister2024efficient}.
As in the hypergraph setting, each node begins with a node budget of one, which is summed when nodes are merged, while features are aggregated through a weighted average with weights proportional to the node budgets. 
During expansion, clusters propagate their node budgets and features to child nodes, and the model predicts how to split the node budget among them while refining their features.
Graph generation, therefore, starts from a single node with a node budget equal to the desired size and proceeds through successive expansion and refinement steps.
Multi-scale graph OT coupling is applied to groups of expanded nodes exactly as in the hypergraph case.

\section{Implementation}

Our implementation relies on a conditional flow-matching framework \citep{lipman2023flowmatchinggenerativemodeling} combined with a \textit{local PPGN} backbone \citep{bergmeister2024efficient}. At each refinement scale, the model jointly predicts: (\textit{i}) node expansion decisions, (\textit{ii}) hyperedge expansion decisions, (\textit{iii}) incidence refinement variables, (\textit{iv}) budget split proportions, and (\textit{v}) node and hyperedge feature refinements. 

Following an endpoint flow-matching framework, the model learns a vector field transporting random initial states toward the target refinement variables. Expansion and incidence variables are initialized from Gaussian priors and transported toward target values in $\{-1, 0,1\}$, determining the corresponding discrete decision. Budget split proportions are constrained to the simplex and initialized from Dirichlet distributions, while node and hyperedge features are initialized from either Gaussian or Dirichlet priors, depending on the feature space. 

During training, each component is learned independently through a mean squared error objective between the predicted and target endpoints.
During sampling, the learned flow is numerically integrated from $t=0$ to $t=1$ using Heun's second-order method.

To focus the model on regions undergoing refinement, we leverage the node budget to perform graph inpainting. Nodes and hyperedges that are not expanded retain their connectivity and features, allowing the model to focus on the changing parts of the graph. Concretely, during training, inactive nodes are excluded from the loss to avoid learning trivial identity mappings. During inference, we enforce several deterministic constraints:
(\textit{i}) clusters with node budget $1$ cannot expand;
(\textit{ii}) clusters of size $2$ are split evenly;
(\textit{iii}) non-expanded nodes preserve their node budgets; and
(\textit{iv}) features of non-expanded nodes and hyperedges remain unchanged.

To ensure consistency of the generated node budgets, predicted split proportions are projected onto the simplex within each expanded cluster. Likewise, when simplex-valued node or hyperedge features are used, they are projected back onto the simplex after each refinement step. 

Graph inpainting is motivated by the following result:

\begin{proposition}
\label{prop:flow_focusing}
Let $\mathbf{X}^{(l)}$ be the upsampled node state at scale $l$ and $\mathbf{Y}^{(l)}$ be the ground-truth target. In hierarchical generation, the node set $\mathcal{V}$ is partitioned into static nodes $\mathcal{S} = \{v : \mathbf{Y}^{(l)}_v = \mathbf{X}^{(l)}_v\}$ and active nodes $\mathcal{A} = \{v : \mathbf{Y}^{(l)}_v \neq \mathbf{X}^{(l)}_v\}$. As hierarchy depth increases, the gradient signal $\nabla_\theta \mathcal{L}$ becomes concentrated around learning an identity mapping due to the asymptotic dominance of $\mathcal{S}$ ($|\mathcal{S}| \gg |\mathcal{A}|$). Restricting the loss support to the active nodes $\mathcal{A}$ recovers the generative signal.
\end{proposition}

Intuitively, as refinement progresses, most nodes remain unchanged while only a small fraction undergo refinement. Including all nodes in the loss can therefore bias the model toward learning an identity mapping. Masking unchanged nodes instead focuses the learning signal on the nodes being refined.

Finally, before interpolation in the flow-matching process, we apply the proposed multi-scale graph OT coupling to align duplicated nodes and hyperedges within each expanded cluster.

The hierarchical strategy ensures high efficiency: the total computational complexity of our approach scales as $\tilde{\mathcal{O}}((n + m + k)\log n)$ for a hypergraph with $n$ nodes, $m$ hyperedges, and $k$ incidences. Additional implementation details regarding training, sampling, self-conditioning, and complexity are provided in Appendices \ref{app:implementation}, \ref{app::training_sampling}, and \ref{app:complexity}.

\section{Experiments and Results} \label{sec:experiments}

In this section, we detail our experimental setup, covering datasets, evaluation metrics, baselines, results, and ablation studies. 
Full experimental details are provided in Appendix \ref{app:experiments}.
Visualizations of the generated samples are provided in Figure \ref{fig:samples}, with additional examples in Appendix \ref{app:visual_examples}.

\subsection{Datasets and Experimental Setup}

\begin{table*}[t]
    \begin{minipage}{0.5\textwidth}
        \centering
        \caption{Evaluation on the graph point cloud datasets.}
        \label{tab:point_clouds_results}
        \centering
        \setlength{\tabcolsep}{2pt}  
        \resizebox{\columnwidth}{!}{
        \begin{tabular}{@{}l*{3}{c}*{3}{c}@{}}
        \toprule
        \multirow{2}{*}{\textbf{Method}} & \multicolumn{3}{c}{\textbf{Airplane Point Clouds}} & \multicolumn{3}{c}{\textbf{Bench Point Clouds}} \\
        & \multicolumn{3}{c}{{($n_{avg} = $ 971.14, $std = $ 63.15)}} &\multicolumn{3}{c}{{($n_{avg} = $ 987.31, $std = $ 49.33)}} \\
        \cmidrule(l){2-4}
        \cmidrule(l){5-7}
        & 
        \textbf{ChamDist} $\downarrow$ & 
        \textbf{ChamCov} (\%) $\uparrow$ & 
        \textbf{ChamDiv} $\uparrow$ & 
        \textbf{ChamDist} $\downarrow$ & 
        \textbf{ChamCov} (\%) $\uparrow$ & 
        \textbf{ChamDiv} $\uparrow$ \\
        \midrule
        Flow-Matching & OOM & OOM & OOM & OOM & OOM & OOM \\
        DiGress & OOM & OOM & OOM & OOM & OOM & OOM \\
        DeFoG   & OOM & OOM & OOM & OOM & OOM & OOM \\
        \cdashline{1-7}\\[-0.8em]
        \method & $\mathbf{0.094 \vcenter{\hbox{\scriptsize $\boldsymbol{\pm 0.006}$}}}$ &
        $\mathbf{22.2 \vcenter{\hbox{\scriptsize $\boldsymbol{\pm 14.5}$}}}$ &
        $\mathbf{0.085 \vcenter{\hbox{\scriptsize $\boldsymbol{\pm 0.026}$}}}$ &
        $\mathbf{0.130 \vcenter{\hbox{\scriptsize $\boldsymbol{\pm 0.001}$}}}$ &
        $\mathbf{15.8 \vcenter{\hbox{\scriptsize $\boldsymbol{\pm 10.6}$}}}$ &
        $\mathbf{0.191 \vcenter{\hbox{\scriptsize $\boldsymbol{\pm 0.114}$}}}$ \\
        \bottomrule
        \end{tabular}
    }
    \end{minipage}
    \hfill
    \begin{minipage}{0.5\textwidth}
        \centering
        \caption{Evaluation on the hypergraph Manifold40 meshes.}
        \label{tab:Manifold40_summary}
        \setlength{\tabcolsep}{3pt}
        \resizebox{\columnwidth}{!}{
        \begin{tabular}{@{}l*{3}{c}*{3}{c}@{}}
        \toprule
        \multirow{2}{*}{\textbf{Method}} & \multicolumn{3}{c}{\textbf{Manifold40 Airplane (Mesh)}} & \multicolumn{3}{c}{\textbf{Manifold40 Bench (Mesh)}} \\
        & \multicolumn{3}{c}{{($n_{avg} = $ 51.86, $std = $ 0.74)}} &\multicolumn{3}{c}{{($n_{avg} = $ 48.90, $std = $ 3.52)}} \\
        \cmidrule(l){2-4}
        \cmidrule(l){5-7}
        & \textbf{ChamDist} $\downarrow$
        & \textbf{ChamCov} (\%) $\uparrow$
        & \textbf{ChamDiv} $\uparrow$
        & \textbf{ChamDist} $\downarrow$
        & \textbf{ChamCov} (\%) $\uparrow$
        & \textbf{ChamDiv} $\uparrow$ \\
        \midrule
        Sequential &
        $\underline{0.082 \vcenter{\hbox{\scriptsize $\pm 0.074$}}}$ &
        $\mathbf{34.8 \vcenter{\hbox{\scriptsize $\boldsymbol{\pm 0.0}$}}}$ &
        $\underline{0.129 \vcenter{\hbox{\scriptsize $\pm 0.011$}}}$ &
        $0.169 \vcenter{\hbox{\scriptsize $\pm 0.007$}}$ &
        $13.3 \vcenter{\hbox{\scriptsize $\pm 0.0$}}$ &
        $0.157 \vcenter{\hbox{\scriptsize $\pm 0.003$}}$ \\
        
        WGAN 
        & $0.124 \vcenter{\hbox{\scriptsize $\pm 0.021$}}$
        & $10.4 \vcenter{\hbox{\scriptsize $\pm 3.5$}}$
        & $0.037 \vcenter{\hbox{\scriptsize $\pm 0.016$}}$
        & $0.128 \vcenter{\hbox{\scriptsize $\pm 0.015$}}$
        & $17.3 \vcenter{\hbox{\scriptsize $\pm 3.3$}}$
        & $0.063 \vcenter{\hbox{\scriptsize $\pm 0.099$}}$ \\
        
        Flow-Matching 
        & $0.085 \vcenter{\hbox{\scriptsize $\pm 0.014$}}$
        & $20.9 \vcenter{\hbox{\scriptsize $\pm 10.0$}}$
        & $0.063 \vcenter{\hbox{\scriptsize $\pm 0.005$}}$
        & $\underline{0.108 \vcenter{\hbox{\scriptsize $\pm 0.019$}}}$
        & $\underline{30.7 \vcenter{\hbox{\scriptsize $\pm 2.5$}}}$
        & $0.112 \vcenter{\hbox{\scriptsize $\pm 0.007$}}$ \\
        
        VAE 
        & $0.231 \vcenter{\hbox{\scriptsize $\pm 0.208$}}$
        & $26.9 \vcenter{\hbox{\scriptsize $\pm 11.7$}}$
        & $\mathbf{0.191 \vcenter{\hbox{\scriptsize $\boldsymbol{\pm 0.130}$}}}$
        & $0.280 \vcenter{\hbox{\scriptsize $\pm 0.223$}}$
        & $27.5 \vcenter{\hbox{\scriptsize $\pm 9.9$}}$
        & $\underline{0.195 \vcenter{\hbox{\scriptsize $\pm 0.107$}}}$ \\
        
        \cdashline{1-7}\\[-0.8em]
        \method 
        & $\mathbf{0.048 \vcenter{\hbox{\scriptsize $\boldsymbol{\pm 0.003}$}}}$
        & $\underline{31.9 \vcenter{\hbox{\scriptsize $\pm 13.2$}}}$
        & $0.101 \vcenter{\hbox{\scriptsize $\pm 0.011$}}$
        & $\mathbf{0.064 \vcenter{\hbox{\scriptsize $\boldsymbol{\pm 0.005}$}}}$
        & $\mathbf{42.2 \vcenter{\hbox{\scriptsize $\boldsymbol{\pm 13.9}$}}}$
        & $\mathbf{0.243 \vcenter{\hbox{\scriptsize $\boldsymbol{\pm 0.037}$}}}$ \\
        
        \bottomrule
        \end{tabular}
        }
    \end{minipage}
\end{table*}

\begin{table*}[!t]
    \centering
    \caption{Ablation studies on the node budgets (Enc.) and multi-scale graph OT coupling (Coup.) for ModelNet and Manifold40.} 
    \setlength{\tabcolsep}{3pt}
    \resizebox{\textwidth}{!}{
    \begin{tabular}{@{}cc*{10}{c}@{}}
    \toprule
    \multirow{3}{*}{\makecell{\textbf{Enc.}}} &
 \multirow{3}{*}{\makecell{\textbf{Coup.}}} &
 \multicolumn{2}{c}{\textbf{ModelNet Bookshelf}} &
 \multicolumn{2}{c}{\textbf{ModelNet Piano}} &
 \multicolumn{3}{c}{\textbf{Manifold40 Airplane}} &
 \multicolumn{3}{c}{\textbf{Manifold40 Bench}} \\
    \cmidrule(l){3-4}     \cmidrule(l){5-6}     \cmidrule(l){7-9}     \cmidrule(l){10-12}
     &
 &
 \makecell{\textbf{Node Num} $\downarrow$} &
 \textbf{Spectral $\downarrow$} &
 \makecell{\textbf{Node Num} $\downarrow$} &
 \textbf{Spectral $\downarrow$} &
 \textbf{ChamDist $\downarrow$} &
 \textbf{ChamCov} (\%) $\uparrow$ &
 \textbf{ChamDiv $\uparrow$} &
 \textbf{ChamDist $\downarrow$} &
 \textbf{ChamCov} (\%) $\uparrow$ &
 \textbf{ChamDiv $\uparrow$} \\
    \midrule
    \cmark &
 \cmark &
 $\mathbf{0.135 \vcenter{\hbox{\scriptsize $\boldsymbol{\pm 0.276}$}}}$ &
 $\underline{0.024 \vcenter{\hbox{\scriptsize $\pm $0.015}}}$ &
 $\mathbf{0.846 \vcenter{\hbox{\scriptsize $\boldsymbol{\pm 1.009}$}}}$ &
 $0.040 \vcenter{\hbox{\scriptsize $\pm $0.026}}$ &
 $\mathbf{0.048 \vcenter{\hbox{\scriptsize $\boldsymbol{\pm 0.003}$}}}$ &
 $\underline{31.9 \vcenter{\hbox{\scriptsize $\pm 13.2$}}}$ &
 $\underline{0.101 \vcenter{\hbox{\scriptsize $\pm 0.011$}}}$ &
 $\mathbf{0.064 \vcenter{\hbox{\scriptsize $\boldsymbol{\pm 0.005}$}}}$ &
 $\mathbf{42.2 \vcenter{\hbox{\scriptsize $\boldsymbol{\pm 13.9}$}}}$ &
 $\mathbf{0.243 \vcenter{\hbox{\scriptsize $\boldsymbol{\pm 0.037}$}}}$ \\
    \xmark &
 \cmark &
 $0.940 \vcenter{\hbox{\scriptsize $\pm $0.917}}$ &
 $0.032 \vcenter{\hbox{\scriptsize $\pm $0.013}}$ &
 $3.622 \vcenter{\hbox{\scriptsize $\pm $1.822}}$ &
 $0.055 \vcenter{\hbox{\scriptsize $\pm $0.040}}$ &
 $0.079 \vcenter{\hbox{\scriptsize $\pm $0.019}}$ &
 $29.0 \vcenter{\hbox{\scriptsize $\pm 17.6$}}$ &
 $0.080 \vcenter{\hbox{\scriptsize $\pm 0.011$}}$ &
 $0.090 \vcenter{\hbox{\scriptsize $\pm $0.003}}$ &
 $37.8 \vcenter{\hbox{\scriptsize $\pm 16.8$}}$ &
 $0.232 \vcenter{\hbox{\scriptsize $\pm 0.048$}}$ \\
    \cmark &
 \xmark &
 $\underline{0.265 \vcenter{\hbox{\scriptsize $\pm $0.496}}}$ &
 $\mathbf{0.014 \vcenter{\hbox{\scriptsize $\boldsymbol{\pm 0.007}$}}}$ &
 $\underline{3.155 \vcenter{\hbox{\scriptsize $\pm $3.637}}}$ &
 $\mathbf{0.030 \vcenter{\hbox{\scriptsize $\boldsymbol{\pm 0.048}$}}}$ &
 $\underline{0.050 \vcenter{\hbox{\scriptsize $\pm $0.005}}}$ &
 $30.4 \vcenter{\hbox{\scriptsize $\pm 15.0$}}$ &
 $0.093 \vcenter{\hbox{\scriptsize $\pm 0.019$}}$ &
 $\underline{0.085 \vcenter{\hbox{\scriptsize $\pm $0.056}}}$ &
 $32.2 \vcenter{\hbox{\scriptsize $\pm 7.7$}}$ &
 $0.208 \vcenter{\hbox{\scriptsize $\pm 0.030$}}$ \\
    \xmark &
 \xmark &
 $1.325 \vcenter{\hbox{\scriptsize $\pm $1.631}}$ &
 $0.031 \vcenter{\hbox{\scriptsize $\pm $0.009}}$ &
 $5.490 \vcenter{\hbox{\scriptsize $\pm $8.847}}$ &
 $\underline{0.036 \vcenter{\hbox{\scriptsize $\pm $0.016}}}$ &
 $0.100 \vcenter{\hbox{\scriptsize $\pm $0.023}}$ &
 $\mathbf{44.9 \vcenter{\hbox{\scriptsize $\boldsymbol{\pm 10.0}$}}}$ &
 $\mathbf{0.114 \vcenter{\hbox{\scriptsize $\boldsymbol{\pm 0.031}$}}}$ &
 $0.098 \vcenter{\hbox{\scriptsize $\pm $0.024}}$ &
 $\mathbf{42.2 \vcenter{\hbox{\scriptsize $\boldsymbol{\pm 22.0}$}}}$ & 
 $\underline{0.235 \vcenter{\hbox{\scriptsize $\pm 0.053$}}}$ \\
    \bottomrule
    \end{tabular}
    }
    \label{tab:ablations-modelnet}
\end{table*}

\textbf{Datasets}. For featureless structures, we evaluate our method on five hypergraph datasets: Stochastic Block Model (\textit{SBM}) \citep{kim2018stochasticblockmodelhypergraphs}, \textit{Ego} \citep{comrie2021hypergraphegonetworkstemporalevolution}, \textit{Tree} \citep{NIEMINEN199935}, and ModelNet40 \textit{bookshelf} and \textit{piano} \citep{wu20153dshapenetsdeeprepresentation}.
We also evaluate \method~on five unfeatured graph datasets from \citep{bergmeister2024efficient}: \textit{SBM}, \textit{Tree}, \textit{Planar}, \textit{Protein}, and \textit{Point cloud}.
Regarding featured structures, we evaluate \method~on two featured hypergraph datasets consisting of 3D meshes: Manifold40 \textit{bench} and \textit{airplane} \citep{subdivnet-tog-2022}; and two graph point cloud datasets sampled from the same mesh categories.
For 3D mesh and point cloud datasets, node features are 3D positions, while edges and hyperedges do not have features.

\textbf{Metrics}.
For featureless hypergraphs, we follow the evaluation protocol of \cite{gailhard2024hygenediffusionbasedhypergraphgeneration}, including: \textit{i}) structural metrics such as \textit{Node Num} (node count difference), and \textit{ii}) topological metrics such as \textit{Spectral} computed as the maximum mean discrepancy (MMD) between spectral distributions.
When datasets impose structural constraints, we also report \textit{Valid}, the percentage of valid generated samples.
Lower values indicate better performance for all metrics except \textit{Valid}, where higher is better.
For 3D meshes and point clouds, we report \textit{ChamDist}, the minimum Chamfer distance to training samples, \textit{ChamCov}, the fraction of reference samples matched by at least one generated sample, and \textit{ChamDiv}, the average pairwise Chamfer distance between generated samples.
Lower values of \textit{ChamDist} indicate better performance, while higher values of \textit{ChamCov} and \textit{ChamDiv} indicate better coverage and diversity, respectively.
For graph datasets, we also report the ratio of several metrics between generated and test sets (see Appendix \ref{app:experiments} for details).
Additional metrics and full numerical results are provided in Appendix \ref{sec:detailed_results}.

\noindent \textbf{Baselines}. For the featureless hypergraphs, we compare~\method~against HYGENE \citep{gailhard2024hygenediffusionbasedhypergraphgeneration}, HyperPA \citep{Do_2020}, a Variational Autoencoder (VAE) \citep{kingma2022autoencodingvariationalbayes}, a Generative Adversarial Network (GAN) \citep{goodfellow2014generativeadversarialnetworks}, and a standard 2D diffusion model \citep{ho2020denoisingdiffusionprobabilisticmodels} trained on incidence matrix images, where hyperedge membership is represented by white pixels and absence by black pixels.
For 3D meshes, we compare \method~with Wasserstein GAN (WGAN) \cite{arjovsky2017wasserstein}, flow-matching, and VAE trained on the concatenation of the incidence matrix plus features, along with a sequential disjoint generation baseline where the hypergraph topology is generated first, followed by the feature generation.
For the graph datasets, we compare \method~against DeFoG \citep{qin2025defogdiscreteflowmatching} and DiGress \citep{vignac2023digress}, two state-of-the-art flat models for graph generation, BwR \citep{diamant2023improving}, a recent graph bandwidth restriction method intended for scalability, and HSpectre \citep{bergmeister2024efficient}, a hierarchical generation method for unfeatured graphs.
All baseline results are taken from \citep{qin2025defogdiscreteflowmatching, bergmeister2024efficient}.

\subsection{Results and Discussion}

\begin{figure*}[!t]
    \centering

    \begin{subfigure}[b]{0.58\textwidth}
        \centering
        \begin{subfigure}[b]{0.31\textwidth}
            \centering
            \includegraphics[width=\textwidth]{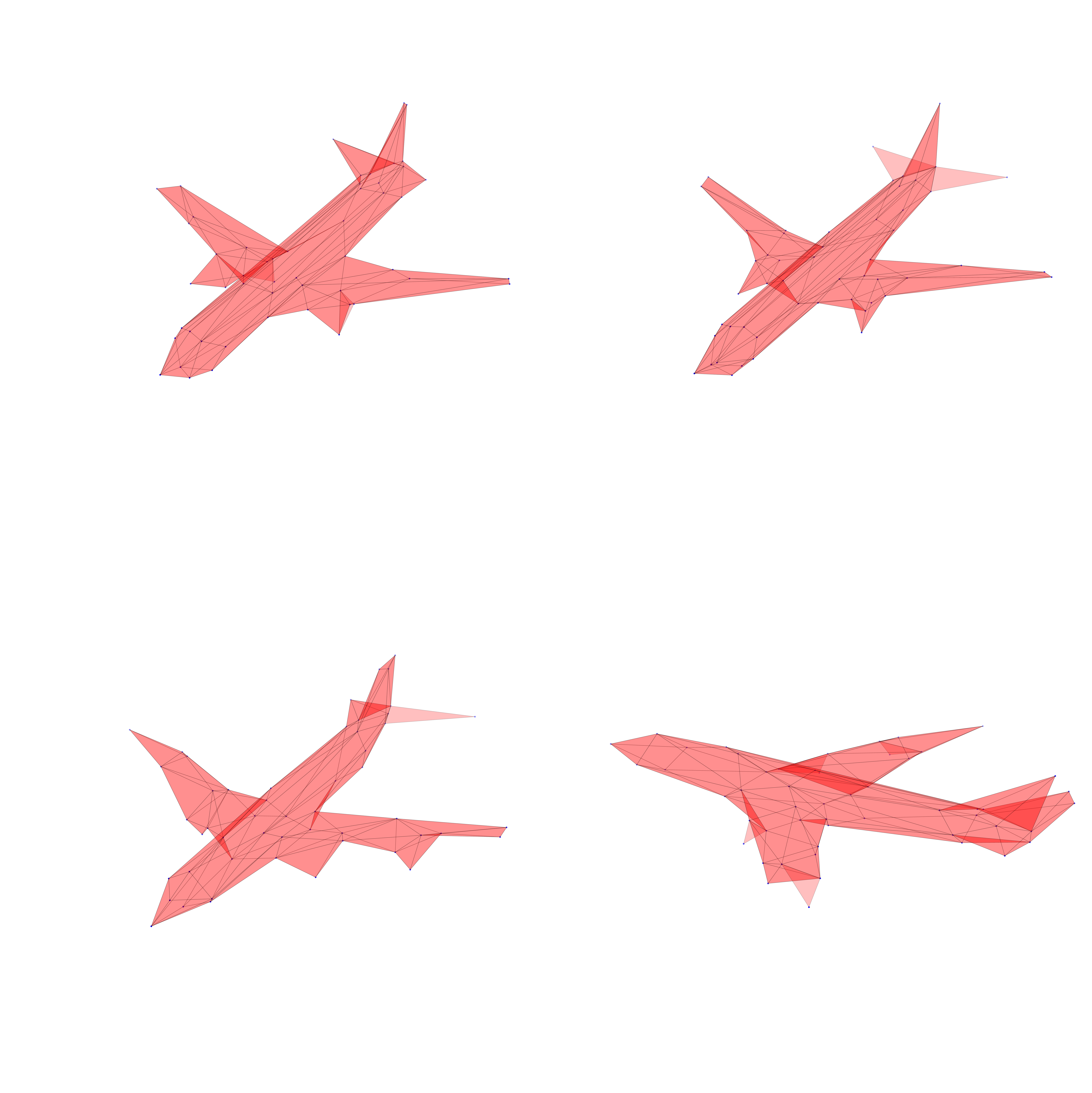}
            \caption*{Train samples}
        \end{subfigure}
        \hfill
        \begin{subfigure}[b]{0.31\textwidth}
            \centering
            \includegraphics[width=\textwidth]{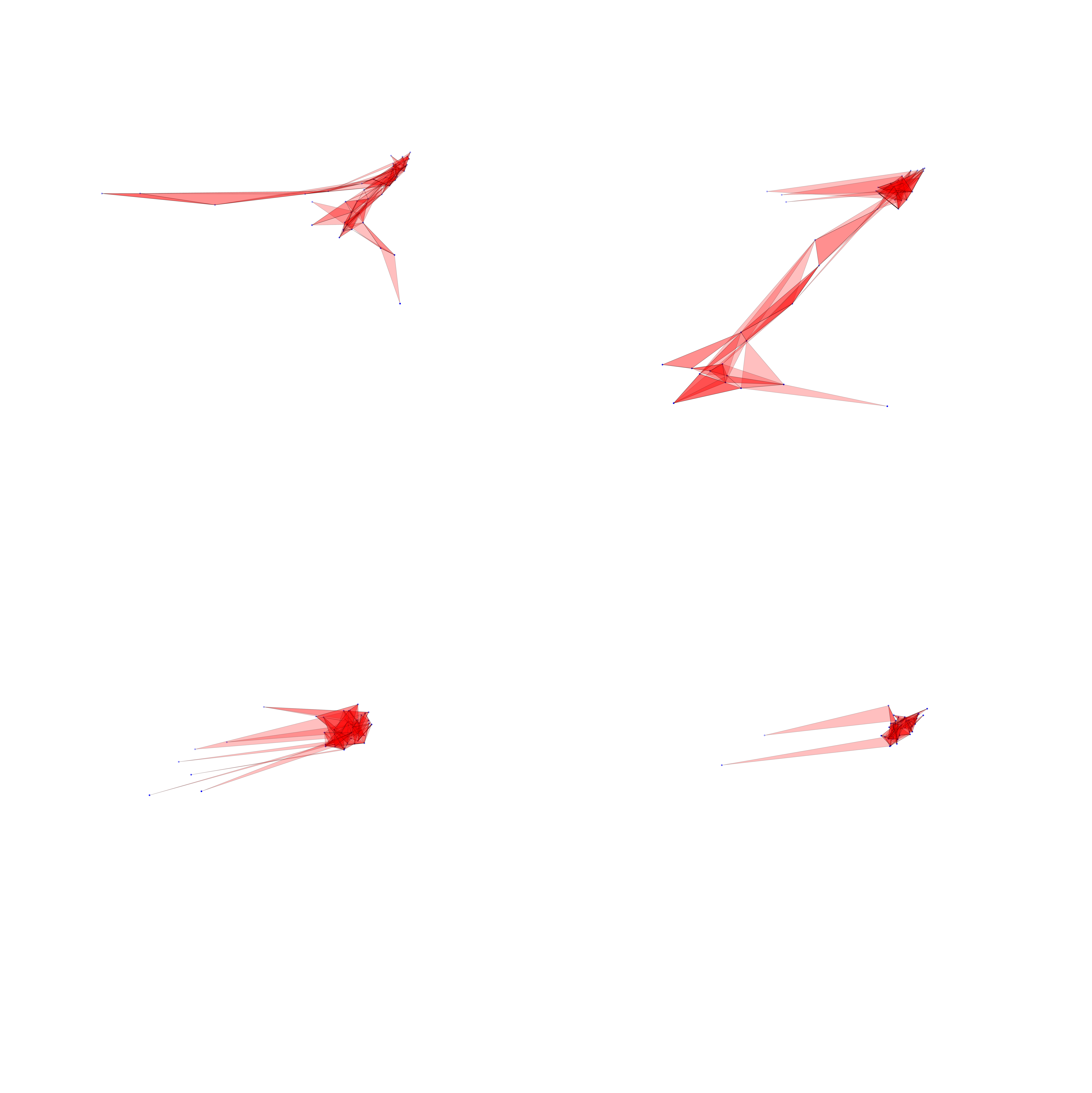}
            \caption*{Generated (sequential)}
        \end{subfigure}
        \hfill
        \begin{subfigure}[b]{0.31\textwidth}
            \centering
            \includegraphics[width=\textwidth]{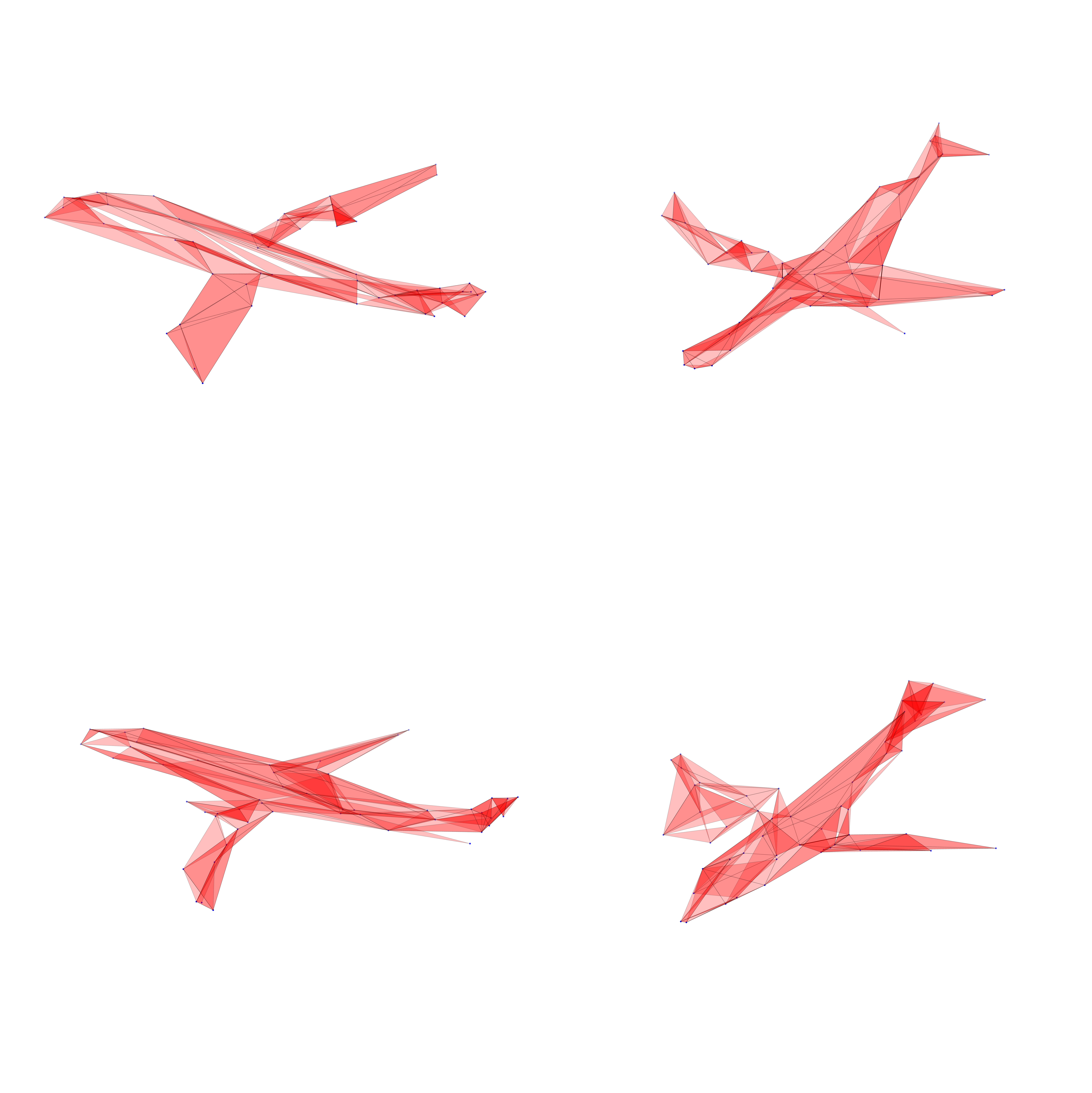}
            \caption*{Generated (joint)}
        \end{subfigure}
        \vspace{-0.15cm}
        \caption*{(\textit{i}) Airplane 3D meshes.}
    \end{subfigure}
    \hfill
    \begin{subfigure}[b]{0.39\textwidth}
        \centering
        \begin{subfigure}[b]{0.47\textwidth}
            \centering
            \includegraphics[width=\textwidth]{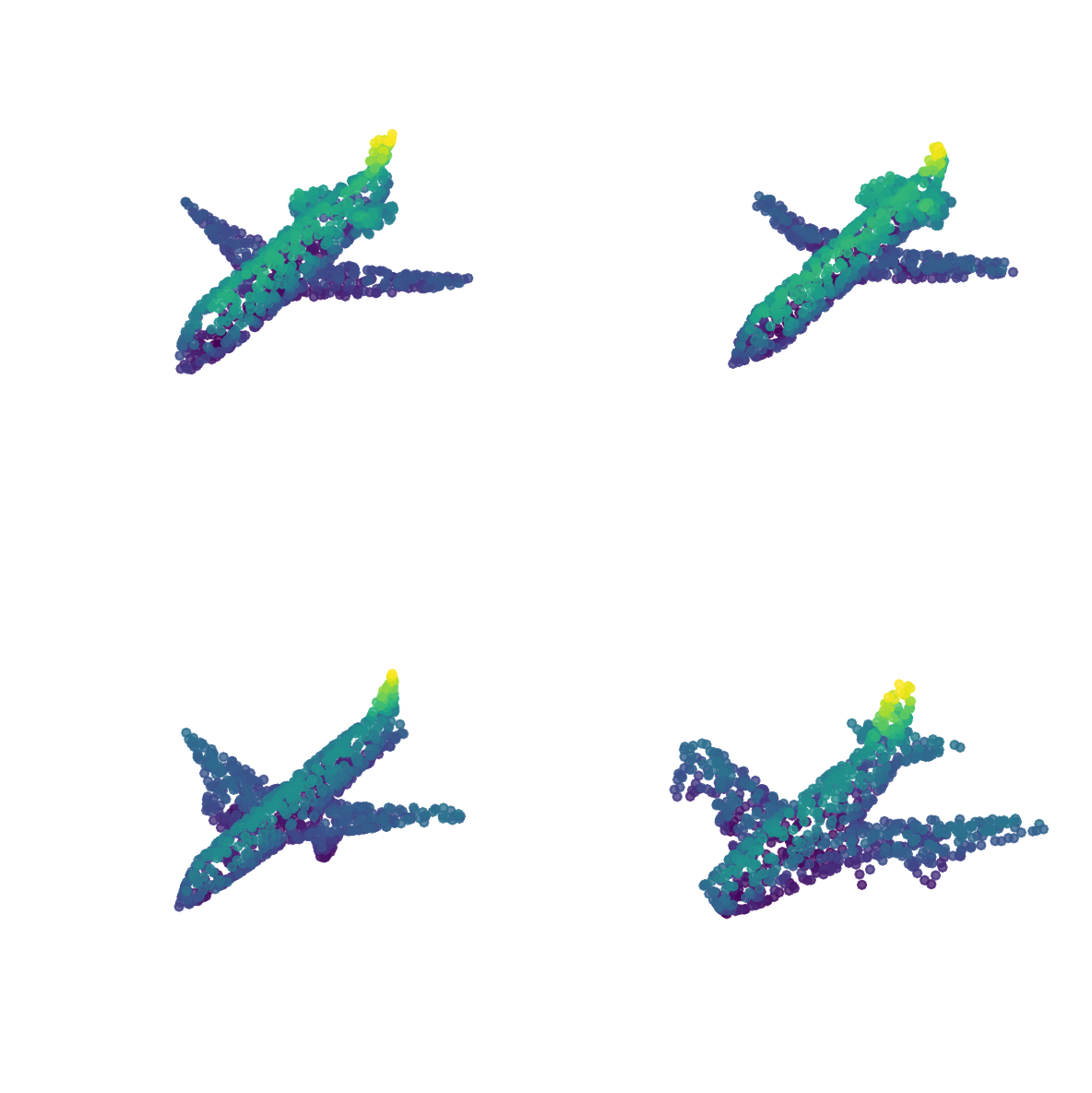}
            \caption*{Train samples}
        \end{subfigure}
        \hfill
        \begin{subfigure}[b]{0.47\textwidth}
            \centering
            \includegraphics[width=\textwidth]{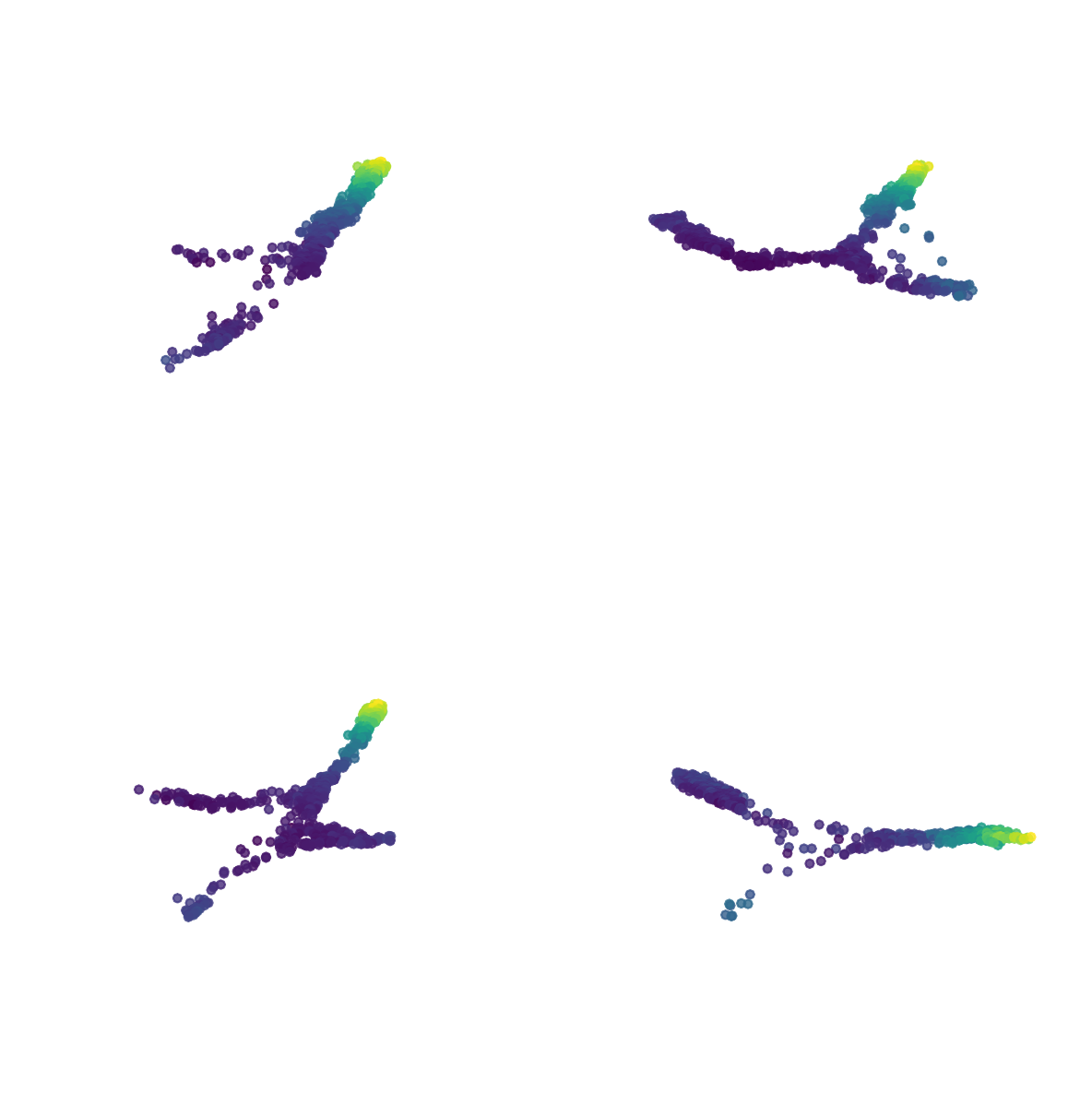}
            \caption*{Generated samples}
        \end{subfigure}
        \vspace{-0.15cm}
        \caption*{(\textit{ii}) Manifold40 Airplane point clouds.}
    \end{subfigure}

    \vspace{0.25cm}

    \begin{subfigure}[b]{0.58\textwidth}
        \centering
        \begin{subfigure}[b]{0.31\textwidth}
            \centering
            \includegraphics[width=\textwidth]{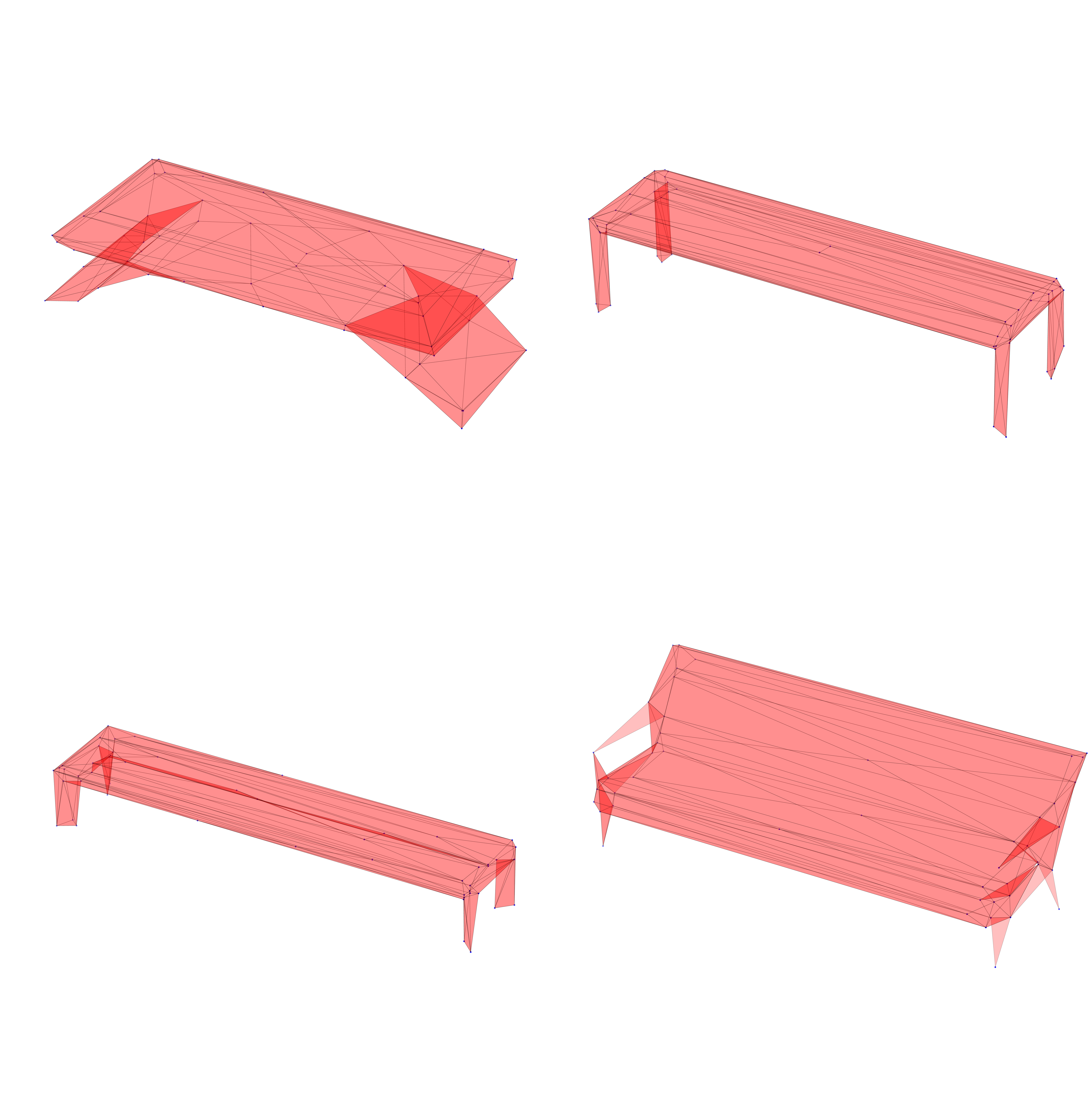}
            \caption*{Train samples}
        \end{subfigure}
        \hfill
        \begin{subfigure}[b]{0.31\textwidth}
            \centering
            \includegraphics[width=\textwidth]{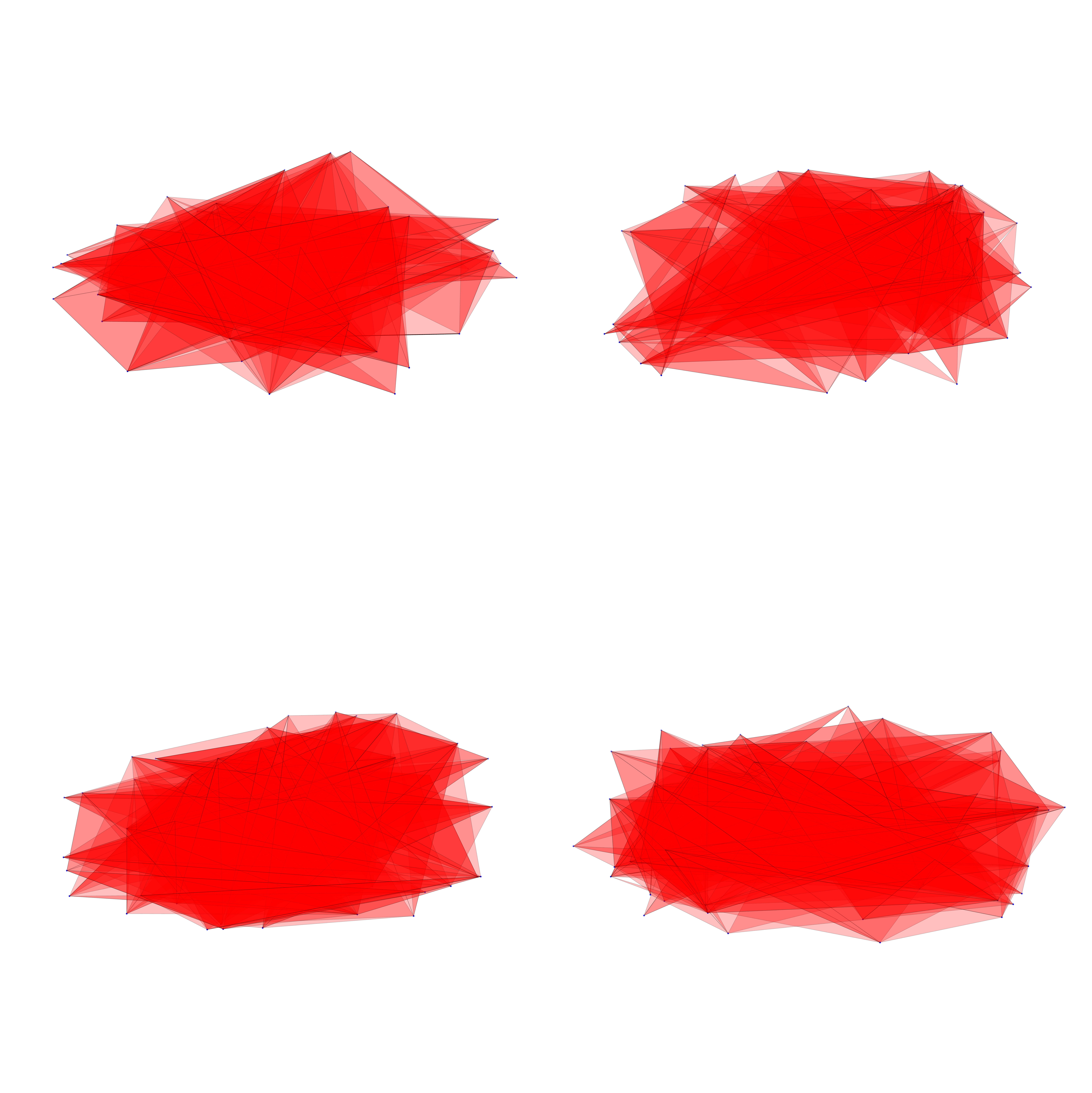}
            \caption*{Generated (sequential)}
        \end{subfigure}
        \hfill
        \begin{subfigure}[b]{0.31\textwidth}
            \centering
            \includegraphics[width=\textwidth]{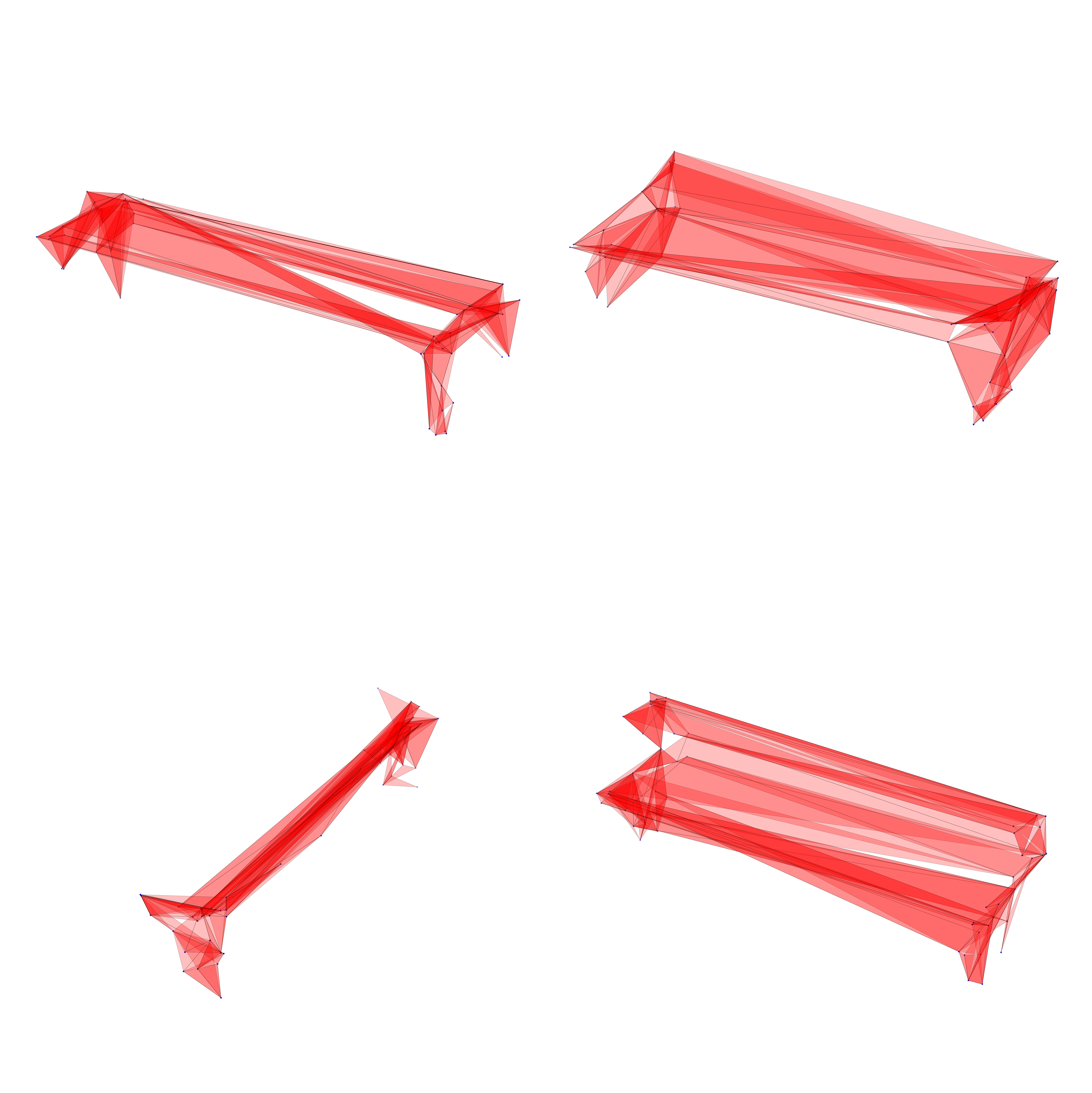}
            \caption*{Generated (joint)}
        \end{subfigure}
        \vspace{-0.15cm}
        \caption*{(\textit{iii}) Bench 3D meshes.}
    \end{subfigure}
    \hfill
    \begin{subfigure}[b]{0.39\textwidth}
        \centering
        \begin{subfigure}[b]{0.47\textwidth}
            \centering
            \includegraphics[width=\textwidth]{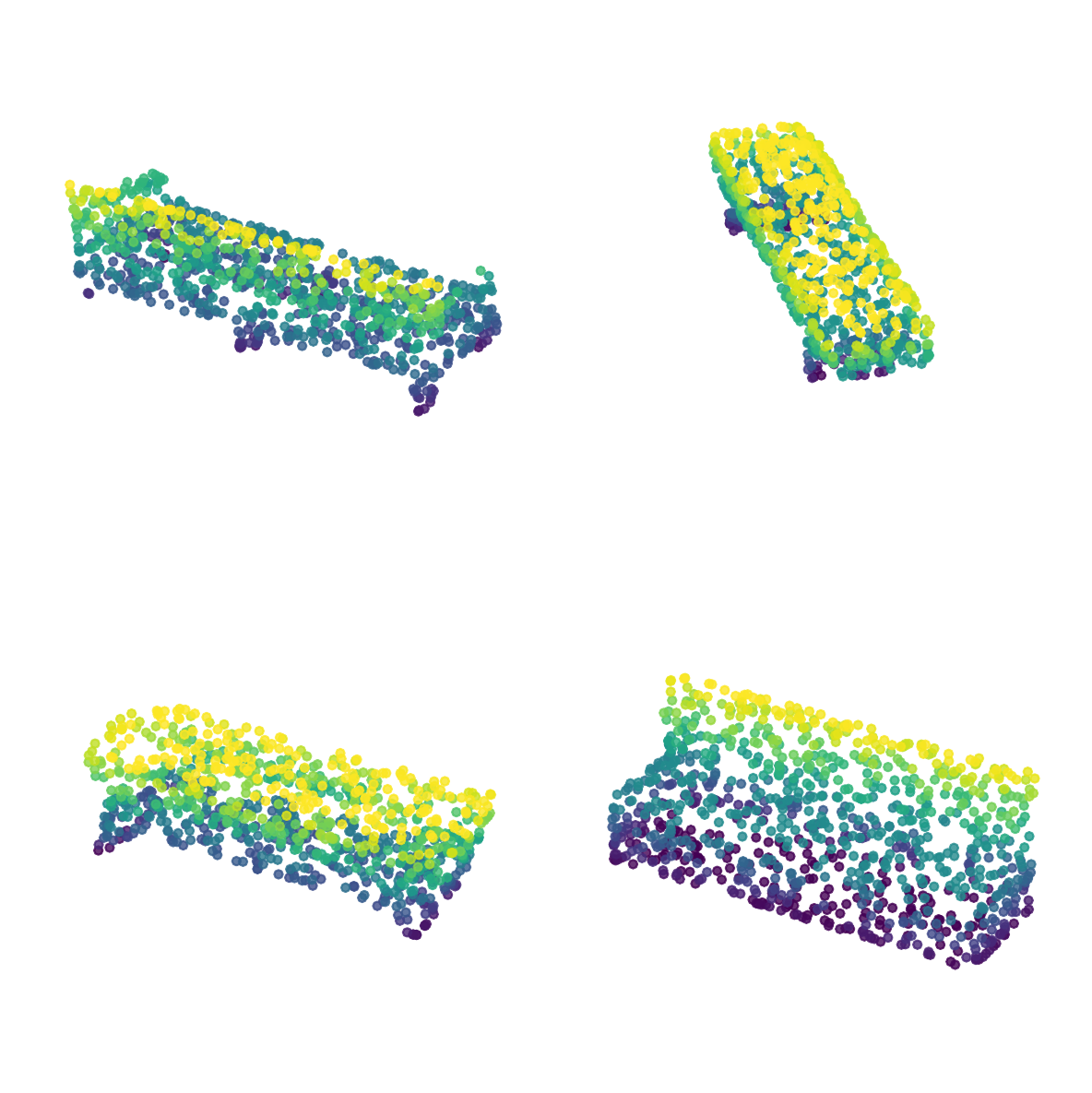}
            \caption*{Train samples}
        \end{subfigure}
        \hfill
        \begin{subfigure}[b]{0.47\textwidth}
            \centering
            \includegraphics[width=\textwidth]{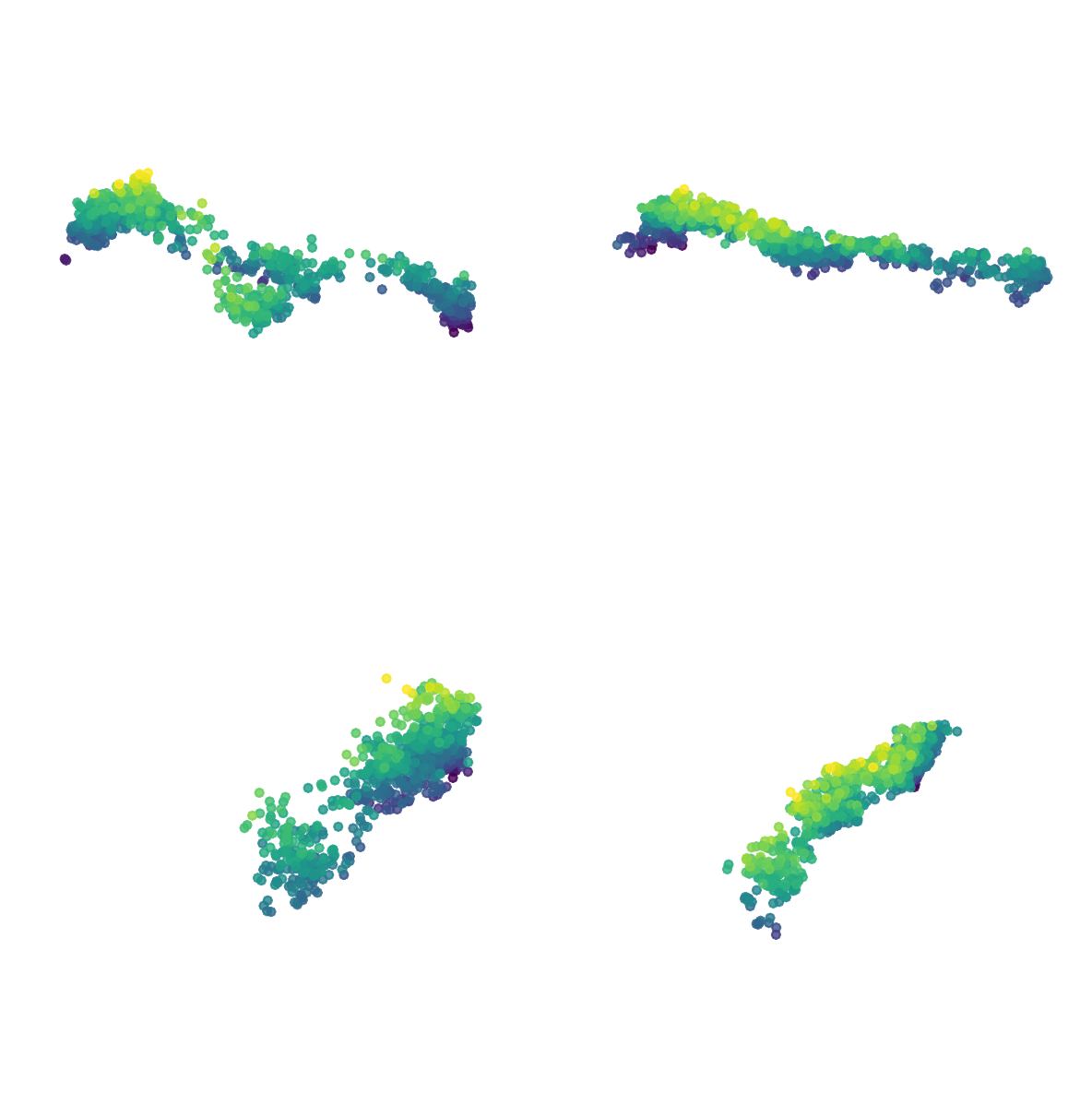}
            \caption*{Generated samples}
        \end{subfigure}
        \vspace{-0.15cm}
        \caption*{(\textit{iv}) Manifold40 Bench point clouds.}
    \end{subfigure}

    \caption{Examples of training samples and samples generated by \method~for the
    \textit{Airplane} and \textit{Bench} datasets, for both meshes and point clouds.
    Sequential generation---where topology is generated before features---produces
    poor-quality meshes, whereas joint generation yields substantially more coherent
    samples. For point clouds, the results remain encouraging despite the difficulty
    of generating large graphs with more than a thousand nodes. Generated airplanes tend to
    exhibit skeleton-like structures, while both airplanes and benches lack
    fine-grained details, resembling intermediate-scale representations rather than
    fully refined samples.}
    
    \label{fig:samples}
\end{figure*}

\textbf{Comparison with the baselines}.
Table \ref{tab:hypergraph_results} shows the comparisons for the featureless hypergraphs, where \textbf{bold} and \underline{underlined} indicate the best and second-best scores.
Our node budgets and multi-scale graph OT coupling consistently improve generation quality.
For the featureless graphs, Table \ref{tab:graphs_results} shows that \method~obtains competitive results compared to state-of-the-art flat methods on small-graph datasets, and that our new components improve generation quality in all datasets compared to HSpectre.
On featured point clouds (Table~\ref{tab:point_clouds_results}), our hierarchical approach is the only method that scales, highlighting its quasi-linear complexity compared to the quadratic cost of flat models.
Regarding the mesh-featured hypergraph datasets, Table \ref{tab:Manifold40_summary} and Figure \ref{fig:samples} show that \method~obtains better results in general than other baselines, such as the sequential generation or naive-joint approaches through concatenation (WGAN, Flow-Matching, and VAE).
This shows the necessity of jointly modeling topology and features in hypergraph generation.
Detailed numerical results are shown in Appendix \ref{sec:detailed_results}.

\subsection{Ablation Studies}
Table \ref{tab:ablations-modelnet} presents ablation studies of \method~for the node budgets and multi-scale graph OT coupling components.
We observe that using node budgets instead of concatenating the target size to each node embedding, like in \citep{bergmeister2024efficient, gailhard2024hygenediffusionbasedhypergraphgeneration}, improves generation quality.
The multi-scale graph OT coupling has a more nuanced effect, clearly improving quality on some datasets, like Manifold40 Airplane, while not changing much for others.
These two components are also essential for feature generation, as lacking one of them results in much worse results, as seen by the large increase in \textit{ChamDist} when one component is ablated.
We present an additional ablation regarding coarsening in Appendix \ref{sec:detailed_results}.

\subsection{Limitations}

While our node budget helps mitigate the issue of missing nodes, it does not fully resolve it, and our method still struggles when generating very large hypergraphs, such as ModelNet Bookshelf and Piano, or very large graphs, such as point cloud datasets. 
These cases highlight that scalability remains a challenge when both the number of nodes and the structural complexity grow substantially.
Moreover, our framework currently assumes continuous feature distributions (\textit{e.g.}, 3D coordinates), and extending it to domains with richer or heterogeneous node and hyperedge attributes (such as categorical or multi-modal features) may require additional modeling components. 
Finally, although FAHNES improves training stability through multi-scale graph OT coupling, training remains computationally expensive for large-scale datasets, which may limit its applicability in resource-constrained settings.

\section{Conclusion}
We presented FAHNES, to the best of our knowledge the first scalable hierarchical framework able to jointly generate topology and features for both hypergraphs and graphs.
By integrating features into coarse-to-fine structural modeling, FAHNES overcomes the limitations of flat or disjoint approaches and enables the generation of complex structures at larger scales.
Our main innovations include a node budget, which provides fine-grained control over local growth and can be leveraged to enhance the training signal, and multi-scale graph OT coupling, which improves alignment and stability during training. 
Extensive experiments on synthetic, protein, 3D mesh, and point cloud datasets show that FAHNES achieves strong performance in both topology and feature generation while scaling to graph and hypergraph sizes that are out of reach for prior models.

Our framework opens new directions for generative modeling of structured data, including applications in 3D geometry and circuit modeling, as well as extensions to richer feature modalities and larger-scale generation.

\section*{Impact Statement}
This paper presents work whose goal is to advance the field of machine learning. There are many potential societal consequences of our work, none of which we feel must be specifically highlighted here.

\section*{Acknowledgments}
The authors acknowledge the French National Research Agency (ANR) for its financial support of the System On Chip Design leveraging Artificial Intelligence (SODA) project under grant ANR-23-IAS3-0004 and the JCJC project DeSNAP ANR-24-CE23-1895-01. This work is also supported by Hi! PARIS and ANR/France 2030 program ANR-23-IACL-0005.


\bibliography{main}
\bibliographystyle{icml2026}

\newpage
\appendix
\onecolumn
\setcounter{proposition}{0}

The appendices offer additional technical details and formal proofs to complement the main paper.
The document is structured as follows:
Our detailed rationale for the probabilistic modeling is presented in Appendix \ref{app:prob_modeling}.
Formal proofs of the main propositions are presented in Appendix~\ref{app:proofs}.
Appendix~\ref{app::coarsening} describes our algorithm for sampling coarsening sequences.
Appendix~\ref{app:coarsening_example} provides illustrative examples of coarsening sequences.
Appendix~\ref{app:experiments} outlines the experimental setup and hyperparameter choices.
Appendix~\ref{app:implementation} discusses implementation details of \method.
Algorithms for model training and sampling featured hypergraphs are detailed in Appendix~\ref{app::training_sampling}.
Appendix~\ref{app:complexity} analyzes the algorithmic complexity of our approach. 
Appendix \ref{sec:detailed_results} presents detailed numerical results for the ablation studies.
Appendix~\ref{app:visual_examples} presents visual comparisons between training and generated samples.

\section{Probabilistic Modeling}
\label{app:prob_modeling}

In this section, we present a formalization of our learning framework, generalizing \citep{bergmeister2024efficient,gailhard2024hygenediffusionbasedhypergraphgeneration}.
Let $\{H^{(1)}, \ldots, H^{(N)}\}$ denote a set of \textit{i.i.d.} hypergraph instances. Our objective is to approximate the unknown generative process by learning a distribution $p(H)$. We model the marginal likelihood of each hypergraph $H$ as a sum over the likelihoods of its bipartite representation's expansion sequences $p(H) = p(B) = \sum_{\varpi \in \Pi(B)} p(\varpi)$,
where $\Pi(B)$ denotes the set of valid expansion sequences from a minimal bipartite graph to the full bipartite representation $B$ corresponding to $H$. Each intermediate $B^{(l-1)}$ is generated by expanding and refining its predecessor, in accordance with Definitions \ref{def::expansion} and \ref{def::refinement}.

To simplify notations, we drop the superscript for $\mathbf{F}^{\text{refine}}$ and simply write $\mathbf{F}$.
Assuming a Markovian generative structure, the likelihood of a specific expansion sequence $\varpi$ is factorized as:
\begin{equation}
p(\varpi) = \overbrace{p(B^{(L)})}^{1}  \prod_{l=L}^1 p(B^{(l-1)} | B^{(l)}) = \prod_{l=L}^1 p(\mathbf{e}^{(l-1)}, \mathbf{f}^{(l-1)}, \mathbf{F}_L^{(l-1)}, \mathbf{F}_R^{(l-1)} | \tilde{B}^{(l-1)})p(\mathbf{v}_L^{(l)}, \mathbf{v}_R^{(l)} | B^{(l)}).
\end{equation}
To simplify the modeling process and avoid learning two separate distributions $p(\mathbf{e}^{(l)}, \mathbf{f}^{(l)}, \mathbf{F}_L^{(l)}, \mathbf{F}_R^{(l)} | \tilde{B}^{(l)})$ and $p(\mathbf{v}_L^{(l)}, \mathbf{v}_R^{(l)} | B^{(l)})$, we rearrange terms as follows:
\begin{equation}
p(\varpi) = p(\mathbf{e}^{(0)}, \mathbf{f}^{(0)}, \mathbf{F}_L^{(0)}, \mathbf{F}_R^{(0)} | \tilde{B}^0) p(\mathbf{v}_L^{(L)}, \mathbf{v}_R^{(L)}) \left[ \prod_{l=L-1}^1 p(\mathbf{v}_L^{(l)}, \mathbf{v}_R^{(l)} | B^{(l)})p(\mathbf{e}^{(l)}, \mathbf{f}^{(l)}, \mathbf{F}_L^{(l)}, \mathbf{F}_R^{(l)} | \tilde{B}^{(l)}) \right],
\end{equation}
where $p(\mathbf{v}_L^{(L)}, \mathbf{v}_R^{(L)}) = p(\mathbf{v}_L^{(L)}, \mathbf{v}_R^{(L)} | B^{(L)})$.

We assume that the variables $\mathbf{v}_L^{(l)}$ and $\mathbf{v}_R^{(l)}$ are conditionally independent of $\tilde{B}^{(l)}$ when conditioned on $B^{(l)}$:
\begin{equation}
p(\mathbf{v}_L^{(l)}, \mathbf{v}_R^{(l)} | B^{(l)}, \tilde{B}^{(l)}) = p(\mathbf{v}_L^{(l)}, \mathbf{v}_R^{(l)} | B^{(l)}).
\end{equation}
This allows us to write the combined likelihood as:
\begin{equation}
p(\mathbf{v}_L^{(l)}, \mathbf{v}_R^{(l)} | B^{(l)})p(\mathbf{e}^{(l)}, \mathbf{f}^{(l)}, \mathbf{F}_L^{(l)}, \mathbf{F}_R^{(l)} | \tilde{B}^{(l)}) = p(\mathbf{v}_L^{(l)}, \mathbf{v}_R^{(l)}, \mathbf{e}^{(l)}, \mathbf{f}^{(l)}, \mathbf{F}_L^{(l)}, \mathbf{F}_R^{(l)} | \tilde{B}^{(l)}),
\end{equation}
which corresponds to the combined expansion and refinement step that inverts coarsening at depth $l$.

\section{Proofs} \label{app:proofs}

\subsection{Optimal Downsampled Features}

\begin{proposition*}
Let $\mathbf{X}$ be the node feature matrix and $\mathcal{P} = \{\mathcal{C}_k\}_{k=1}^K$ a partition of the vertices into clusters. The optimal downsampled features minimizing the reconstruction error are the cluster-wise barycenters:
\begin{equation*}
\mathbf{X}^*_{\mathcal{C}_k} = \frac{1}{|\mathcal{C}_k|} \sum_{v \in \mathcal{C}_k} \mathbf{X}_v \quad \text{for $1 \leq k \leq K$}.
\end{equation*}
\end{proposition*}

\begin{proof}
Let \( H = (\mathcal{V}, \mathcal{E}) \) be a hypergraph with node set \( \mathcal{V} \) and hyperedge set \( \mathcal{E} \), and let \( H_l = (\mathcal{V}_l, \mathcal{E}_l) \) denote the lifted hypergraph obtained by expanding each cluster \( \mathcal{C} \subseteq \mathcal{V} \) into a \(|C|\)-clique. By construction, there exists a bijection \( \phi : \mathcal{V} \to \mathcal{V}_l \) mapping each original node to its corresponding lifted node.
Suppose each node \( v \in \mathcal{V} \) is associated with a feature vector \( \mathbf{X}_v \in \mathbb{R}^d \), and each cluster \( \mathcal{C} \subseteq \mathcal{V} \) is assigned a cluster feature vector \( \mathbf{X}_\mathcal{C} \in \mathbb{R}^d \), which is inherited by all lifted nodes \( \phi(v) \) for \( v \in \mathcal{C} \).
Define the mean squared error between the original node features and the cluster features in the lifted hypergraph as:
\begin{equation}
\mathrm{MSE} = \sum_{v \in \mathcal{V}} \| \mathbf{X}_v - \mathbf{X}_{C(v)} \|^2,
\end{equation}
where \( C(v) \) denotes the cluster containing node \( v \).

To minimize the MSE, it is enough to minimize, for each cluster \( \mathcal{C} \),
\begin{equation}
J_\mathcal{C}(\mathbf{X}_\mathcal{C}) = \sum_{v \in \mathcal{C}} \| \mathbf{X}_v - \mathbf{X}_\mathcal{C} \|^2.
\end{equation}
Since \( J_\mathcal{C} \) is a convex quadratic function in \( \mathbf{X}_\mathcal{C} \), we find its minimum by setting the gradient to zero:
\begin{equation}
\nabla_{\mathbf{X}_\mathcal{C}} J_\mathcal{C}(\mathbf{X}_C) = \sum_{v \in \mathcal{C}} 2 (\mathbf{X}_\mathcal{C} - \mathbf{X}_v) = 2 |\mathcal{C}| \mathbf{X}_\mathcal{C} - 2 \sum_{v \in \mathcal{C}} \mathbf{X}_v = \mathbf{0}.
\end{equation}
Solving for \( \mathbf{X}_\mathcal{C} \), we obtain
\begin{equation}
\mathbf{X}_\mathcal{C} = \frac{1}{|\mathcal{C}|} \sum_{v \in \mathcal{C}} \mathbf{X}_v,
\end{equation}
which is the arithmetic mean of the feature vectors in the cluster \( \mathcal{C} \).
\end{proof}

\subsection{Multi-scale graph OT coupling}

Let $B$ be a bipartite graph.
Let $\mathbf{b}$ be its current node budget, $\mathbf{F}_L$ its node features, and $\mathbf{F}_R$ its hyperedge features.
Denote $\mathbf{X} = (\mathbf{v}_L, \mathbf{v}_R, \mathbf{e}, \mathbf{f}, \mathbf{F}_L^{\text{refine}}, \mathbf{F}_R^{\text{refine}})$, \ie the predictions of the model. We will denote $\mathbf{X}_0$ the initial noise samples and $\mathbf{X}_1$ the targets.
In the following, all distributions are conditioned on $\mathbf{b}$, $\mathbf{F}_L$ and $\mathbf{F}_R$.

\begin{proposition*}
Under the OT coupling of Algorithm~\ref{alg:minibatch_ot}, the joint distribution:
\[
q(\mathbf{X}_0, \mathbf{X}_1)
\]
has marginals:
\[
q_0(\mathbf{X}_0), \quad
q_1(\mathbf{X}_1).
\]
\end{proposition*}

\begin{proof}
Let $(\mathbf{X}_0,\mathbf{X}_1)$ be sampled independently from
$q_0$ and $q_1$, and let
$\sigma_{\mathbf{X}_0,\mathbf{X}_1}$ denote the reindexing selected by
Algorithm~\ref{alg:minibatch_ot}.
The coupled source variable is then
\[
\widetilde{\mathbf{X}}_0
=
\sigma_{\mathbf{X}_0,\mathbf{X}_1}(\mathbf{X}_0).
\]

By construction, Algorithm~\ref{alg:minibatch_ot} only permutes nodes
within an expanded cluster, which preserves the
conditioning graph, \ie its topology, node budgets, and features.
Consequently,
\[
(B,\mathbf{X}_0)
\cong
\bigl(B,\sigma_{\mathbf{X}_0,\mathbf{X}_1}(\mathbf{X}_0)\bigr).
\]

Let $f$ be any bounded measurable function on bipartite graphs.
Since graphs are invariant to node reindexing, $f$ is invariant under
the above isomorphism, and therefore
\[
f\!\left(B,\widetilde{\mathbf{X}}_0\right)
=
f\!\left(B,\mathbf{X}_0\right)
\]
for every realization $(\mathbf{X}_0,\mathbf{X}_1)$.
Taking expectations gives
\begin{align}
\mathbb{E}_{q}[f(B,\mathbf{X}_0)]
&=
\mathbb{E}_{q_0(\mathbf{X}_0)q_1(\mathbf{X}_1)}
\left[
f\!\left(
B,
\sigma_{\mathbf{X}_0,\mathbf{X}_1}(\mathbf{X}_0)
\right)
\right] \\
&=
\mathbb{E}_{q_0(\mathbf{X}_0)q_1(\mathbf{X}_1)}
\left[f(B,\mathbf{X}_0)\right] \\
&=
\mathbb{E}_{q_0(\mathbf{X}_0)}
\left[f(B,\mathbf{X}_0)\right].
\end{align}
Hence, the source marginal induced by the coupling is $q_0$ on the
space of bipartite graphs.

The target samples $\mathbf{X}_1$ are not modified by the coupling;
only their pairing with source samples changes. Therefore, for any
bounded measurable function $g$,
\[
\mathbb{E}_{q}[g(B,\mathbf{X}_1)]
=
\mathbb{E}_{q_1(\mathbf{X}_1)}[g(B,\mathbf{X}_1)],
\]
and the target marginal remains $q_1$.

Thus, the joint distribution induced by
Algorithm~\ref{alg:minibatch_ot} has marginals $q_0$ and $q_1$.
\end{proof}

\begin{proposition*}
Let $\mathcal{B} = \{(\mathbf{X}^0_i, \mathbf{X}^1_i)\}_{i=1}^N$ be a minibatch of size $N$ coupled by a permutation $\pi$.
Let $\Delta_\pi$ be the random variable representing the displacement vector $\mathbf{X}^1_i - \mathbf{X}^0_{\pi(i)}$ for an index $i$ drawn uniformly from the batch.
Multi-scale graph OT coupling (Algorithm~\ref{alg:minibatch_ot}) yields a permutation $\pi^*$ that reduces the total variance of this displacement:
\[
\operatorname{tr}(\operatorname{Var}(\Delta_{\pi^*})) \leq \operatorname{tr}(\operatorname{Var}(\Delta_{Id})).
\]
\end{proposition*}

\begin{proof}
The total variance decomposes into:
\begin{equation}
    \operatorname{Tr}(\operatorname{Var}(\Delta_\pi)) = \mathbb{E}[\|\Delta_\pi\|^2] - \|\mathbb{E}[\Delta_\pi]\|^2.
\end{equation}
First, by the linearity of expectation, we have:
\[
\mathbb{E}[\Delta_\pi] = \mathbb{E}[\mathbf{X}^1] - \mathbb{E}[\mathbf{X}^0_\pi].
\]
Since the permutation $\pi$ only reorders the samples within the minibatch without changing the set of values, the marginal expectations $\mathbb{E}[\mathbf{X}^1]$ and $\mathbb{E}[\mathbf{X}^0_\pi] = \mathbb{E}[\mathbf{X}^0]$ do not depend on the coupling. Thus, this term is constant.

Second, we can rewrite:
\[
\mathbb{E}[\|\Delta_\pi\|^2] \propto \sum_{i=1}^N \|\mathbf{X}^1_i - \mathbf{X}^0_{\pi(i)}\|^2 = \sum_{\text{clusters } \mathcal{C}} \sum_{i \in \mathcal{C}} \|\mathbf{X}^1_i - \mathbf{X}^0_{\pi(i)}\|^2.
\]
Algorithm~\ref{alg:minibatch_ot} is constructed to minimize the $\sum_{i \in \mathcal{C}} \|\mathbf{X}^1_i - \mathbf{X}^0_{\pi(i)}\|^2$ locally within each cluster.
Consequently, we have:
\[
\mathbb{E}[\|\Delta_{\pi^*}\|^2] \leq \mathbb{E}[\|\Delta_{Id}\|^2].
\]
Since the first term is reduced and the second term is constant, the total variance $\operatorname{Tr}(\operatorname{Var}(\Delta_{\pi^*}))$ is reduced.
\end{proof}

\subsection{Preservation of the Gradient Signal}

\begin{proposition*}
Let $\mathbf{X}^{(l)}$ be the upsampled node state at scale $l$ and $\mathbf{Y}^{(l)}$ be the ground-truth target. In hierarchical generation, the node set $\mathcal{V}$ is partitioned into static nodes $\mathcal{S} = \{v : \mathbf{Y}^{(l)}_v = \mathbf{X}^{(l)}_v\}$ and active nodes $\mathcal{A} = \{v : \mathbf{Y}^{(l)}_v \neq \mathbf{X}^{(l)}_v\}$. As hierarchy depth increases, the gradient signal $\nabla_\theta \mathcal{L}$ becomes concentrated around learning an identity mapping due to the asymptotic dominance of $\mathcal{S}$ ($|\mathcal{S}| \gg |\mathcal{A}|$). Restricting the loss support to the active nodes $\mathcal{A}$ recovers the generative signal.
\end{proposition*}

\begin{proof}
Let $\mathbf{X}^{(l)}$ be the upsampled state serving as input for scale $l$, and $\mathbf{Y}^{(l)}$ be the ground truth target. We partition the node set $\mathcal{V}$ into active nodes $\mathcal{A}$ (where refinement is required, $\mathbf{Y}_v \neq \mathbf{X}_v$) and static nodes $\mathcal{S}$ (where the coarse approximation is exact, $\mathbf{Y}_v = \mathbf{X}_v$).

A standard global objective minimizes the MSE over the entire graph:
\begin{equation}
    \mathcal{L}_{\text{total}} = \overbrace{\sum_{v \in \mathcal{S}} \| f_\theta(\mathbf{X}_t, \mathbf{X}_v) - \mathbf{X}_v \|_2^2}^{\text{Identity Reconstruction}} + \overbrace{\sum_{v \in \mathcal{A}} \| f_\theta(\mathbf{X}_t, \mathbf{X}_v) - \mathbf{Y}_v \|_2^2}^{\text{Structural Refinement}},
\end{equation}
where $\mathbf{X}_t$ is the denoised sample at time $t$.
As generation deepens, $|\mathcal{S}| \to N$ while $|\mathcal{A}|$ becomes small. Consequently, the Identity Reconstruction term dominates the loss landscape. Minimizing this term encourages the model to learn the trivial solution $f_\theta(\mathbf{X}_t, \mathbf{X}) \approx \mathbf{X}$ globally.
Thus, the gradients required to learn the complex refinement in $\mathcal{A}$ are overwhelmed by the gradients enforcing stationarity in $\mathcal{S}$.

By masking inactive nodes, the objective reduces to:
\begin{equation}
    \mathcal{L}_{\text{masked}} = \sum_{v \in \mathcal{A}} \| f_\theta(\mathbf{X}_t, \mathbf{X}_v) - \mathbf{Y}_v \|_2^2.
\end{equation}
This removes the identity bias entirely, ensuring that the capacity of the model and gradient updates are driven solely by active regions.
\end{proof}

\section{Coarsening Sequence Sampling} \label{app::coarsening}

This section outlines our methodology for sampling a coarsening sequence $\pi \in \Pi_F(H)$ for a given hypergraph $H$. The full procedure is detailed in Algorithm~\ref{alg:hypergraph_coarsening}.
At each coarsening step $l$, let $H^{(l)}$ denote the current hypergraph, $B^{(l)}$ its bipartite representation, and $C^{(l)}$ its weighted clique expansion. We begin by sampling a target reduction fraction $\text{red\_frac} \sim \mathcal{U}([\rho_\text{min}, \rho_\text{max}])$. We then evaluate all possible contraction sets $F(C^{(l-1)})$ using a cost function $c$, where lower cost indicates higher preference.
We employ a greedy randomized strategy that processes contraction sets in order of increasing cost. For each set:
\begin{itemize}[leftmargin=0.5cm]
    \itemsep0em 
    \item The set is stochastically rejected with probability $\lambda$.
    \item If not rejected:
    \begin{itemize}
        \item \textbf{Overlap check:} If the contraction set overlaps with any previously accepted contraction, it is discarded.
        \item \textbf{Coarsening attempt:} Otherwise, we compute tentative coarsened representations $C_\text{temp}$ and $B_\text{temp}$.
        \item \textbf{Cluster constraint check:} If all right-side clusters in $B_\text{temp}$ contain at most three nodes, the contraction is accepted.
        \item \textbf{Update step:} When a contraction is accepted, we:
        \begin{itemize}
            \item Sum the node budgets of the nodes in the contraction set to define the new cluster node budget.
            \item Compute the new cluster's node features as a weighted average of the original features, using node budgets as weights.
        \end{itemize}
    \end{itemize}
\end{itemize}

The loop terminates once the number of remaining nodes satisfies the stopping condition:
\[
|\mathcal{V}_L^{(l-1)}| - |\bar{\mathcal{V}}_L^{(l)}| > \text{red\_frac} \cdot |\mathcal{V}_L^{(l-1)}|,
\]
\ie when the number of nodes on the left side (corresponding to the original hypergraph nodes) has been reduced by the sampled fraction.
This framework allows a variety of cost functions $c$, contraction families $F$, reduction fraction ranges $[\rho_\text{min}, \rho_\text{max}]$, and randomization parameters $\lambda$.

\noindent \textbf{Practical considerations.} To avoid oversampling small graphs during training, we follow the heuristic of \cite{bergmeister2024efficient}: when the current graph has fewer than 16 nodes, we fix the reduction fraction to $\rho = \rho_\text{max}$.
Due to the constraint that no right-side cluster in $B^{(l)}$ may contain more than three nodes, achieving the target reduction fraction is not always feasible. However, we observe empirically that this is rarely problematic when $\rho_\text{max}$ is reasonably small.

During training, we sample a coarsening sequence for each dataset graph, but only retain a randomly selected intermediate graph from the sequence. Thus, our practical implementation of Algorithm~\ref{alg:hypergraph_coarsening} is designed to return a single coarsened graph with associated features and node budgets, rather than the full sequence $\pi$.

To improve efficiency, we adopt the caching mechanism of \cite{bergmeister2024efficient}. Once a coarsening sequence is generated, all its levels are cached. At each training step, a random level is sampled from the cache and then removed. A new coarsening sequence is generated only once all cached levels for a given graph have been used, avoiding unnecessary recomputation.

\noindent \textbf{Hyperparameters}. 
In all experiments described in Section \ref{sec:experiments}, we use the following settings:
\begin{itemize}[leftmargin=0.5cm]
    \itemsep0em 
    \item Contraction family: The set of all edges in the clique representation, \ie $F(C) = \mathcal{E}$, for a weighted clique expansion $C = (\mathcal{V}, \mathcal{E})$.
    \item Cost function: Local Variation Cost \cite{loukas2018graph} with a preserving eigenspace size of $k = 8$.
    \item Reduction fraction range: $[\rho_\text{min}, \rho_\text{max}] = [0.1, 0.3]$.
\end{itemize}

\begin{algorithm*}[hbtp]
\caption{\textbf{Hypergraph coarsening sequence sampling}: Randomized iterative coarsening of a hypergraph. At each step, contraction sets are selected based on cost, while ensuring that right-side clusters contain at most three nodes. Accepted contractions update the hypergraph structure, node budgets, and features.}
\label{alg:hypergraph_coarsening}
\begin{algorithmic}[1]

\STATE {\bfseries Parameters:} contraction family \(F\), cost function \(c\), reduction fraction range \([\rho_\text{min}, \rho_\text{max}]\), randomization parameter \(\lambda\)
\STATE {\bfseries Input:} hypergraph \(H\) with node features \(\mathbf{F}_L\) and hyperedge features \(\mathbf{F}_R\)
\STATE {\bfseries Output:} coarsening sequence \(\pi = (B^{(0)}, \ldots, B^{(L)})\)

\STATE Initialize \(H^{(0)} \gets H\)
\STATE \(C^{(0)} \gets \text{WeightedCliqueExpansion}(H^{(0)})\)
\STATE \(B^{(0)} \gets \text{FeaturedBipartiteRepresentation}(H^{(0)}, \mathbf{F}_L, \mathbf{F}_R)\)
\STATE Initialize all node and hyperedge budgets in \(B^{(0)}\) to \(1\)
\STATE \(\pi \gets (B^{(0)})\)
\STATE \(l \gets 0\)

\WHILE{\(|\mathcal{V}_L^{(l)}| > 1\)}
    \STATE \(\text{red\_frac} \sim \text{Uniform}([\rho_\text{min}, \rho_\text{max}])\)
    \STATE \(f \gets c(\cdot, C^{(l)})\)
    \STATE \(\text{accepted\_contractions} \gets \emptyset\)
    \STATE \(C_\text{next} \gets C^{(l)}\)
    \STATE \(B_\text{next} \gets B^{(l)}\)

    \FOR{\(S \in \text{SortedByCost}(F(C^{(l)}), f)\)}
        \IF{\(\text{Random}() > \lambda\)}
            \IF{\(S \cap
            \left(\bigcup_{P \in \text{accepted\_contractions}} P\right)
            = \emptyset\)}

                \STATE \(C_\text{temp}
                \gets \text{CoarsenCliqueExpansion}(C_\text{next}, S)\)

                \STATE \(B_\text{temp}
                \gets \text{CoarsenBipartite}(B_\text{next}, S)\)
                \COMMENT{Updates topology, budgets, and features}

                \IF{\(\forall \text{ right cluster } R
                \text{ induced by } B_\text{temp},\ |R| \leq 3\)}
                    \STATE \(\text{accepted\_contractions}
                    \gets \text{accepted\_contractions} \cup \{S\}\)
                    \STATE \(C_\text{next} \gets C_\text{temp}\)
                    \STATE \(B_\text{next} \gets B_\text{temp}\)
                \ENDIF
            \ENDIF
        \ENDIF

            \IF{\(|\mathcal{V}_L^{(l)}|
            - |\mathcal{V}_{L,\text{next}}|
            \geq \text{red\_frac}\,|\mathcal{V}_L^{(l)}|\)}
                \STATE \textbf{break}
            \ENDIF
    \ENDFOR

    \STATE \(l \gets l + 1\)
    \STATE \(C^{(l)} \gets C_\text{next}\)
    \STATE \(B^{(l)} \gets B_\text{next}\)
    \STATE \(\pi \gets \pi \cup (B^{(l)})\)

\ENDWHILE

\STATE \textbf{return} \(\pi\)

\end{algorithmic}
\end{algorithm*}

\newpage

\section{Examples of Coarsening Sequences} \label{app:coarsening_example}

\begin{figure}[h!]
    \centering
    \begin{subfigure}[b]{\textwidth}
        \centering
        \includegraphics[width=\textwidth]{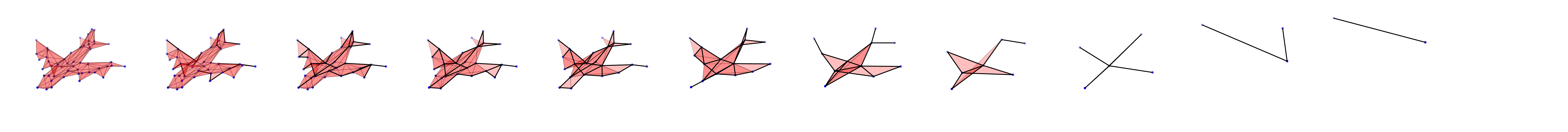}
    \end{subfigure}

    \vspace{0.21cm}
    
    \begin{subfigure}[b]{\textwidth}
        \centering
        \includegraphics[width=\textwidth]{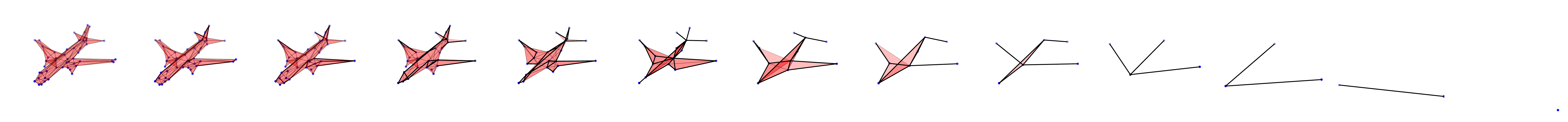}
    \end{subfigure}

    \vspace{0.21cm}
    
    \begin{subfigure}[b]{\textwidth}
        \centering
        \includegraphics[width=\textwidth]{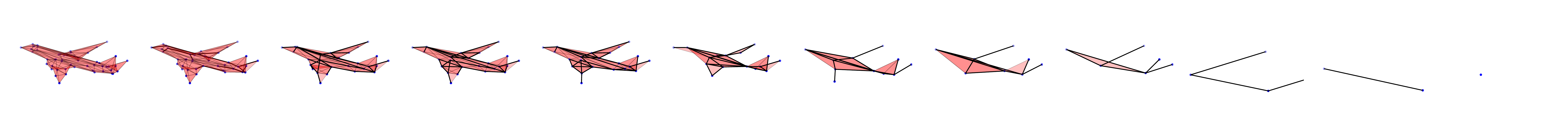}
    \end{subfigure}

    \vspace{0.21cm}
    
    \begin{subfigure}[b]{\textwidth}
        \centering
        \includegraphics[width=\textwidth]{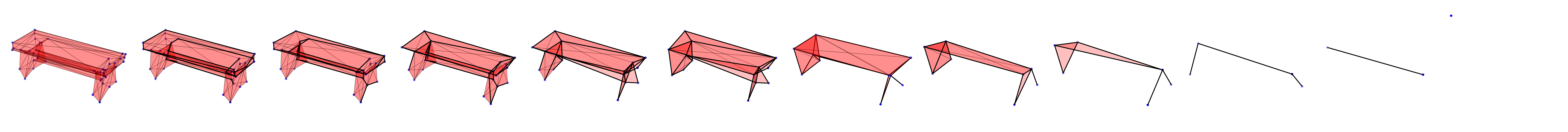}
    \end{subfigure}

    \vspace{0.21cm}
    
    \begin{subfigure}[b]{\textwidth}
        \centering
        \includegraphics[width=\textwidth]{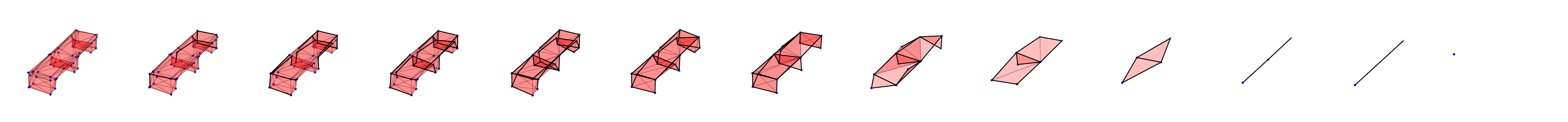}
    \end{subfigure}

    \vspace{0.21cm}
    
    \begin{subfigure}[b]{\textwidth}
        \centering
        \includegraphics[width=\textwidth]{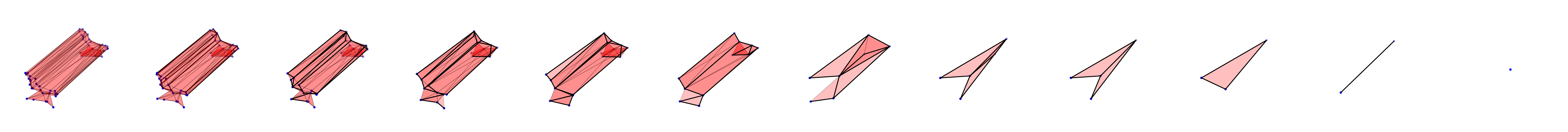}
    \end{subfigure}

    \vspace{0.21cm}
    
    \begin{subfigure}[b]{\textwidth}
        \centering
        \includegraphics[width=\textwidth]{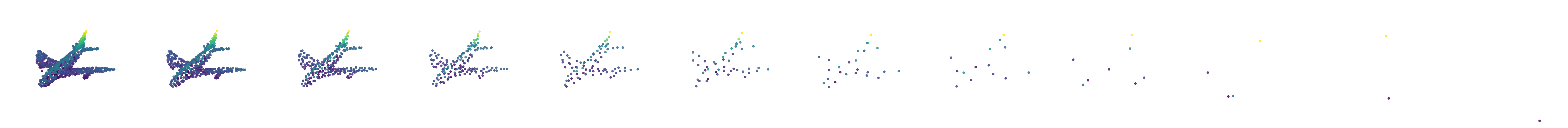}
    \end{subfigure}

    \vspace{0.21cm}
    
    \begin{subfigure}[b]{\textwidth}
        \centering
        \includegraphics[width=\textwidth]{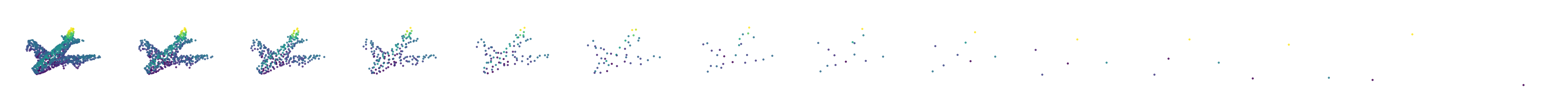}
    \end{subfigure}

    \vspace{0.21cm}
    
    \begin{subfigure}[b]{\textwidth}
        \centering
        \includegraphics[width=\textwidth]{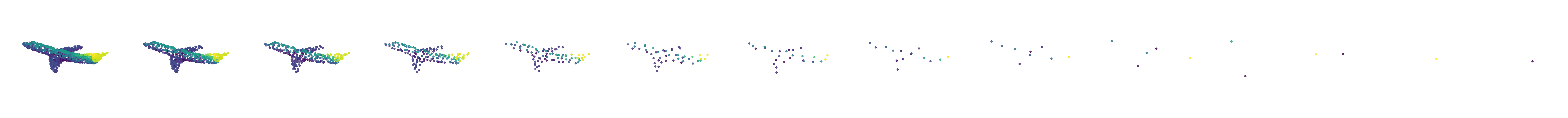}
    \end{subfigure}

    \vspace{0.21cm}
    
    \begin{subfigure}[b]{\textwidth}
        \centering
        \includegraphics[width=\textwidth]{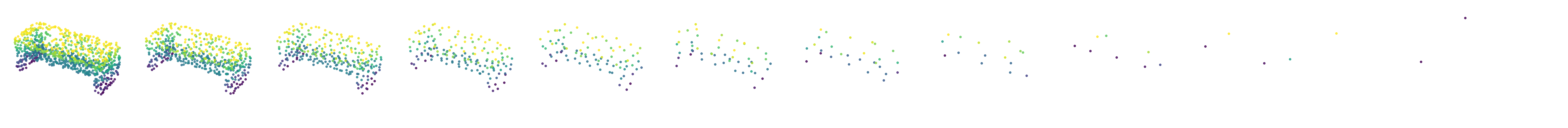}
    \end{subfigure}

    \vspace{0.21cm}
    
    \begin{subfigure}[b]{\textwidth}
        \centering
        \includegraphics[width=\textwidth]{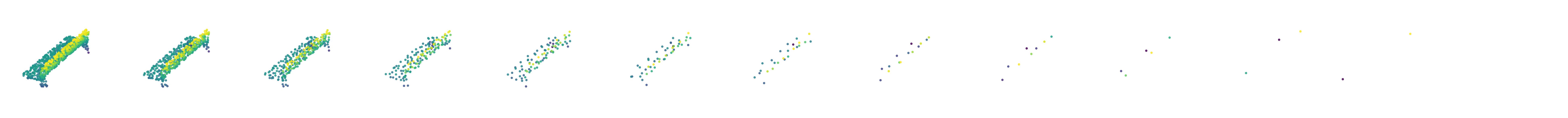}
    \end{subfigure}

    \vspace{0.21cm}
    
    \begin{subfigure}[b]{\textwidth}
        \centering
        \includegraphics[width=\textwidth]{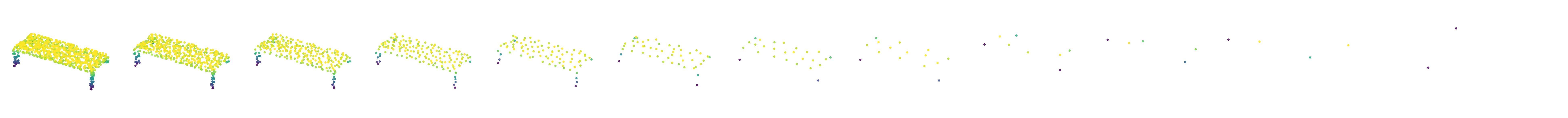}
    \end{subfigure}
\caption{Examples of coarsening sequence for various meshes and point clouds. Thick lines represent 2-edges. Edges are omitted in point clouds for clarity.}
\end{figure}

\clearpage

\section{Experimental Details} 
\label{app:experiments}
In this section, we detail all three types of experiments -- unfeatured hypergraphs, 3D meshes and graph point clouds -- individually, detailing their datasets, baselines, metrics and specific hyperparameters.

For all experiments, we use embeddings with 32 dimensions for edge selection vectors and node and hyperedge expansion numbers.
When they exist, features are embedded with 128 dimensions.
SignNet always has 5 layers and a hidden dimension of 128.
Positional encodings always have 32 dimensions.
We always use 10 layers of Local PPGN.
We use the AdamW \citep{loshchilov2017decoupled} optimizer with a learning rate of $1e-4$.
All experiments are run for 1M steps on a single L40S.
We use 8 CPU workers.

\subsection{Unfeatured Hypergraphs} \label{app:unfeatured_hypergraphs}

\noindent \textbf{Datasets.}  
Our experiments use five datasets: three synthetic and two real-world, consistent with those described in \cite{gailhard2024hygenediffusionbasedhypergraphgeneration}:
\begin{itemize}[leftmargin=0.5cm]
    \itemsep0em 
    \item \textbf{Stochastic Block Model (SBM) hypergraphs} \cite{kim2018stochasticblockmodelhypergraphs}\textbf{:} Constructed with 32 nodes split evenly into two groups. Every hyperedge connects three nodes. Hyperedges are sampled with probability 0.05 within groups and 0.001 between groups.
    \item \textbf{Ego hypergraphs} \cite{comrie2021hypergraphegonetworkstemporalevolution}\textbf{:} Created by generating an initial hypergraph of 150–200 nodes with 3000 randomly sampled hyperedges (up to 5 nodes each), then extracting an ego-centric subgraph by selecting a node and retaining only hyperedges that include it.    
    \item \textbf{Tree-structured hypergraphs} \cite{NIEMINEN199935}\textbf{:} A tree with 32 nodes is generated using \textit{networkx}, followed by grouping adjacent tree edges into hyperedges. Each hyperedge contains up to 5 nodes.
    
    \item \textbf{ModelNet40 meshes} \cite{wu20153dshapenetsdeeprepresentation}\textbf{:} Hypergraphs are derived from mesh topologies of selected ModelNet40 categories. To simplify computation, meshes are downsampled to fewer than 1000 vertices by iteratively merging nearby vertices. Duplicate triangles are removed, and the resulting low-poly mesh is converted into a hypergraph. We focus on the \textit{bookshelf} and \textit{piano} categories.
\end{itemize}

\noindent All datasets are divided into 128 training, 32 validation, and 40 testing hypergraphs.

\noindent \textbf{Evaluation Metrics.}  
We evaluate generated hypergraphs using the same suite of metrics as \cite{gailhard2024hygenediffusionbasedhypergraphgeneration}:
\begin{itemize}[leftmargin=0.5cm]
    \itemsep0em 
    \item \textbf{NodeNumDiff:} Average absolute difference in node count between generated and reference hypergraphs.
    \item \textbf{NodeDegreeDistrWasserstein:} Wasserstein distance between node degree distributions of generated and reference hypergraphs.
    \item \textbf{EdgeSizeDistrWasserstein:} Wasserstein distance between hyperedge size distributions.
    \item \textbf{Spectral:} Maximum Mean Discrepancy (MMD) between Laplacian spectra.
    \item \textbf{Uniqueness:} Fraction of generated hypergraphs that are non-isomorphic to one another.
    \item \textbf{Novelty:} Fraction of generated hypergraphs that are non-isomorphic to training samples.
    \item \textbf{CentralityCloseness, CentralityBetweenness, CentralityHarmonic:} Wasserstein distances computed between centrality distributions (on edges for $s=1$). For details see \cite{aksoy2020hypernetworksciencehighorderhypergraph}.
    \item \textbf{ValidEgo:} For the \textit{hypergraphEgo} dataset only, proportion of generated hypergraphs that contain a central node shared by all hyperedges.
    \item \textbf{ValidSBM:} For the \textit{hypergraphSBM} dataset only, proportion of generated graphs that retain the original intra- and inter-group connectivity patterns.
    \item \textbf{ValidTree:} For the \textit{hypergraphTree} dataset only, proportion of generated samples that preserve tree structure.
\end{itemize}

\noindent \textbf{Baselines.}  
We compare our method against the following baselines:
\begin{itemize}[leftmargin=0.5cm]
    \itemsep0em 
    \item \textbf{HyperPA} \cite{Do_2020}: A heuristic approach for hypergraph generation.
    
    \item \textbf{Image-based models:} We design three baseline models—Diffusion, GAN, and VAE—that operate on incidence matrix representations of hypergraphs:
    \begin{itemize}
        \item Each model is trained to produce binary images where white pixels signify node-hyperedge membership.
        \item Incidence matrices are permuted randomly and padded with black pixels to normalize input sizes.
        \item Generated images are thresholded to obtain binary incidence matrices.
    \end{itemize}
    
    \item \textbf{HYGENE} \cite{gailhard2024hygenediffusionbasedhypergraphgeneration}: A hierarchical diffusion-based generator using reduction, expansion, and refinement steps.
    
\end{itemize}

\noindent \textbf{Specific hyperparameters.} We use $\lambda = 0.3$, our Local PPGN layers have a dimension of 128, and the hidden dimension for our MLP is 256.
We use perturbed hypergraph expansion with a radius of 2 and dropout of 0.5.
The SignNet relies on the top $K = 2$ eigenvalues and eigenvectors of the graph.
We use 256 sampling steps during flow-matching.
Our model has 4M parameters.

\subsection{3D Meshes} 
\label{app:meshes}

\noindent \textbf{Datasets.}
Datasets for meshes are taken from Manifold40 \citep{subdivnet-tog-2022}, which is a reworked version of ModelNet40 \citep{wu20153dshapenetsdeeprepresentation} to obtain manifold and watertight meshes.
Meshes are subsequently coarsened to obtain low-poly versions of 50 vertices and 100 triangles.
We use two classes:
\begin{itemize}[leftmargin=0.5cm]
    \itemsep0em 
    \item \textit{Airplane} comprising 682 training samples, 21 validation samples, and 23 testing samples.
    \item \textit{Bench} comprising 144 training samples, 19 validation samples, and 30 testing samples.
\end{itemize}

\noindent \textbf{Metrics.}
We use the same metrics as for unfeatured hypergraphs.
To this, we add \textit{ChamDist}, which computes the minimal Chamfer distance between a point cloud sampled from the surface of the generated mesh and equivalent point clouds sampled from all train set meshes, \textit{ChamCov} (coverage), which measures the fraction of reference samples matched by at least one generated sample, and \textit{ChamDiv} (diversity), which measures the average pairwise Chamfer distance between generated samples.

\noindent \textbf{Baselines.} We compare against a simple sequential baseline:
\begin{enumerate}[leftmargin=0.5cm]
    \itemsep0em 
    \item Our model (4M parameters) is trained for 1M steps on the topology of meshes \textbf{without} learning to generate the features.
    \item A simple Local PPGN model (4M parameters) is trained for 20 epochs as a flow-matching model to learn to generate the 3D positions, with the topology fixed.
    \item We use the best checkpoint of the first model to generate the topology, then apply the second model on this topology to generate the 3D positions.
\end{enumerate}

Additionally, we compare our method to the following three simple baselines:
\begin{enumerate}
    \item A flow-matching model of 6M parameters, using self-conditioning.
    \item A Wasserstein GAN whose generator has 23M parameters and critic has 7M parameters.
    \item A VAE with 23M parameters.
\end{enumerate}

All models are trained for 100 epochs using the following framework:
\begin{itemize}
        \item Each model is trained to produce binary images where white pixels signify node-hyperedge membership. An additional set of 3 dimension is concatenated to the row for each node, containing the position of the node.
        \item To normalize input sizes, incidence matrices are permuted randomly and padded with black pixels.
        \item Generated images are thresholded to obtain binary incidence matrices. Predicted positions are kept as is.
    \end{itemize}

\noindent \textbf{Specific hyperparameters.} We use $\lambda = 0.1$, our Local PPGN layers have a dimension of 200, and the hidden dimension for our MLP is 300.
We use perturbed hypergraph expansion with a radius of 2 and dropout of 0.5.
We do not use SignNet's positional encodings. Instead, we take a random feature tensor for each node, which is replicated across all expanded nodes belonging to the same cluster.
We use 25 sampling steps during flow-matching.
Our model has 6M parameters.

\subsection{Unfeatured Graph Datasets} 
\label{app:graphs}
\noindent \textbf{Datasets.}  
We reuse the same datasets used in \citep{bergmeister2024efficient}:
\begin{itemize}[leftmargin=0.5cm]
    \itemsep0em
    \item \textbf{Planar graphs}\textbf{:}  
    This dataset consists of 200 planar graphs, each containing 64 nodes. Graphs are obtained by applying Delaunay triangulation to points sampled uniformly at random within the unit square.
    \item \textbf{SBM graphs}\textbf{:}  
    Comprising 200 samples, these graphs are generated from a stochastic block model with 2--5 communities. Each community contains 20--40 nodes. Edges are sampled with probability 0.3 between nodes in the same community and 0.05 otherwise.
    \item \textbf{Tree graphs}\textbf{:}  
    We generate 200 random trees with 64 nodes using \textit{networkx}.
    \item \textbf{Protein graphs}\textbf{:}  
    \cite{DOBSON2003771} provides a dataset where proteins are represented as graphs where nodes correspond to amino acids. An edge is introduced between two nodes if the Euclidean distance between the corresponding amino acids is less than 6\,\AA.
    \item \textbf{Point cloud graphs}\textbf{:}  
    This dataset, introduced by \cite{liao2019efficient}, contains 41 point clouds representing household objects \citep{neumann2013graph}. Each point cloud is converted into a graph, after which only the largest connected component is retained due to frequent disconnectedness in the raw graphs.
\end{itemize}
\noindent 20\% of samples are reserved for testing, while the remaining graphs are partitioned into 80\% training and 20\% validation.

\noindent \textbf{Evaluation Metrics.}  
We evaluate generated graphs using the same suite of metrics used in \citep{bergmeister2024efficient}. We compute the Maximum Mean Discrepancy (MMD) between generated and test graphs with respect to the following graph properties \textit{Degree distribution}, \textit{Clustering coefficient}, \textit{Orbit counts}, \textit{Spectrum}, \textit{Wavelet coefficients}.
\noindent As a reference, we also compute these metrics between the training and test sets and report the mean ratio across all applicable metrics. Certain statistics are omitted for datasets where they are ill-defined or degenerate: for the \textit{point cloud} dataset, the degree MMD is always zero due to the $k$-nearest-neighbor construction, and for the \textit{tree} dataset, both clustering coefficients and orbit counts are identically zero.
\noindent Finally, we include dataset-specific validity metrics:  
\begin{itemize}[leftmargin=0.5cm]
    \itemsep0em
    \item \textbf{ValidPlanar:} Proportion of generated graphs that remain planar.
    \item \textbf{ValidSBM:} Proportion consistent with the original SBM parameters.
    \item \textbf{ValidTree:} Proportion that preserve tree structure (\textit{i.e.}, cycle-free).
\end{itemize}

\noindent \textbf{Baselines.} We compare against two state-of-the-art flat generative models, one graph bandwidth reduction method intended for scalability, and one hierarchical generative model: Defog \cite{qin2025defogdiscreteflowmatching}, DiGress \cite{vignac2023digress}, BwR \cite{diamant2023improving}, and HSpectre \cite{bergmeister2024efficient}.

\noindent \textbf{Specific hyperparameters.} We use $\lambda = 0.3$, our Local PPGN layers have a dimension of 128, and the hidden dimension for our MLP is 256.
We use perturbed hypergraph expansion with a radius of 2 and dropout of 0.5.
The SignNet relies on the top $K = 2$ eigenvalues and eigenvectors of the graph for the \textit{Planar}, \textit{Point cloud} and \textit{Tree} datasets, and we use random positional encodings for the \textit{SBM} and \textit{Protein} datasets.
For the \textit{Point cloud} dataset, we use dropout with probability 0.1 during the computation of node embeddings and between each layer of our GNN.
We use 256 sampling steps during flow-matching.
Our model has 4M parameters.

\subsection{Graph Point Clouds} 
\label{app:point_clouds}

\noindent \textbf{Datasets.}
We reuse the same datasets as for 3D meshes.
Point clouds comprising 1024 nodes are sampled on the surface of each mesh, then each node is connected to its 3 nearest neighbors.
Only the largest connected component is kept.

\noindent \textbf{Metrics.}  
We use the same metrics as for unfeatured graphs.
To this, we add \textit{ChamDist}, which computes the minimal Chamfer distance between a generated point cloud and all train set point clouds, \textit{ChamCov} (coverage), which measures the fraction of reference point clouds matched by at least one generated sample, and \textit{ChamDiv} (diversity), which measures the average pairwise Chamfer distance between generated point clouds.

\noindent \textbf{Baselines.} We compare against two state-of-the-art flat generative models: Defog \cite{qin2025defogdiscreteflowmatching}, DiGress \cite{vignac2023digress}.
As these two methods are tailored for discrete data, we use a mixed-discrete continuous framework for continuous features.

\noindent \textbf{Specific hyperparameters.} We use $\lambda = 0.3$, our Local PPGN layers have a dimension of 128, and the hidden dimension for our MLP is 256.
We use perturbed graph expansion with a radius of 2 and dropout of 0.5.
We do not use SignNet's positional encodings. Instead, we take a random feature tensor for each node, which is replicated across all expanded nodes belonging to the same cluster.
We use 256 sampling steps during flow-matching.
Our model has 4M parameters.

\section{Implementation Details} \label{app:implementation}

\subsection{Flow-matching Framework} \label{app:flow-matching}

We use a flow-matching generative modeling framework~\cite{lipman2023flowmatchinggenerativemodeling}, with endpoint parameterization following~\citet{dunn2024mixedcontinuouscategoricalflow}, equivalent to denoising diffusion models~\cite{ho2020denoisingdiffusionprobabilisticmodels} using linear interpolation between the prior \( p_0 \) and target \( p_1 \) distributions.
The goal is to align two distributions by transporting samples from \( p_0 \) to \( p_1 \) through a learned time-dependent vector field \( \mathbf{f}(\mathbf{X}, t) \), governed by the ODE:
\begin{equation}
    \frac{\partial \mathbf{X}}{\partial t} = \mathbf{f}(\mathbf{X}, t), \quad \mathbf{X}_0 \sim p_0.
\end{equation}
Here, \( \mathbf{f}(\mathbf{X}, t) \) is trained such that integrating this ODE produces samples from \( p_1 \).

\noindent \textbf{Endpoint parameterization.}  
Instead of directly modeling the flow field, we learn the terminal point \( \hat{\mathbf{X}}_1(\mathbf{X}_t, 1) \) of the trajectory. The flow can be recovered using:
\begin{equation}
    \mathbf{f}(\mathbf{X}_t, t) = \frac{\hat{\mathbf{X}}_1(\mathbf{X}_t, 1) - \mathbf{X}_t}{1 - t}.
\end{equation}

\noindent \textbf{Training objective}.  
In our flow-matching framework, the model is trained to minimize the expected squared error between the predicted and true endpoints:
\begin{equation}
    \mathcal{L} = \mathbb{E}_{t, \: \mathbf{X}_0 \sim p_0, \: \mathbf{X}_1 \sim p_1} 
    \left[ \left\| \hat{\mathbf{X}}_1(\mathbf{X}_t, 1) - \mathbf{X}_1 \right\|^2 \right],
\end{equation}
where \( \mathbf{X}_t = (1 - t) \mathbf{X}_0 + t \mathbf{X}_1 \) is a linear interpolation between samples from the prior and target distributions.
For the hypergraphs, 3D mesh and point cloud datasets, during training, $t$ is sampled using a log-normal distribution, which has been shown to improve performance \cite{esser2024scalingrectifiedflowtransformers}. For regular unfeatured graph datasets, we use a standard uniform sampling.

In our setting, each loss term is masked so that it is evaluated only on the relevant nodes or hyperedges. 
Specifically, we retain: \textit{(i)} expansion losses only for nodes with node budget greater than one ; 
\textit{(ii)} node budget losses only for expanded nodes whose parent has a node budget greater than two; and 
\textit{(iii)} feature losses only for expanded nodes. 
For a given sample, let
\begin{align*}
    \mathcal{L}_{\text{node-expansion}} 
        &= \left\| \hat{\mathbf{v}}_L - \mathbf{v}_L \right\|^2 \,\mathbf{1}[\mathbf{b} > 1], \\
    \mathcal{L}_{\text{node budget}} 
        &= \left\| \hat{\mathbf{f}} - \mathbf{f} \right\|^2 \,\mathbf{1}[\mathbf{b} > 2 \ \wedge\ \text{child of expanded cluster}], \\
    \mathcal{L}_{\text{node-feature}} 
        &= \left\| \hat{\mathbf{F}}_L - \mathbf{F}_{L}^{\text{refine}} \right\|^2 \,\mathbf{1}[\text{child of expanded cluster}], \\
    \mathcal{L}_{\text{hyperedge-expansion}} 
        &= \left\| \hat{\mathbf{v}}_R - \mathbf{v}_R \right\|^2, \\
    \mathcal{L}_{\text{hyperedge-feature}} 
        &= \left\| \hat{\mathbf{F}}_R - \mathbf{F}_{R}^{\text{refine}} \right\|^2 \,\mathbf{1}[\text{child of expanded cluster}], \\
    \mathcal{L}_{\text{incidence}} 
        &= \left\| \hat{\mathbf{e}} - \mathbf{e} \right\|^2,
\end{align*}
where $\mathbf{1}[\cdot]$ denotes the indicator function.
The total loss is the sum of all components, averaged across samples:
\begin{equation*}
    \mathcal{L} = 
    \mathbb{E}\!\left[
        \mathcal{L}_{\text{node-expansion}}
        + \mathcal{L}_{\text{node budget}}
        + \mathcal{L}_{\text{node-feature}}
        + \mathcal{L}_{\text{hyperedge-expansion}}
        + \mathcal{L}_{\text{hyperedge-feature}}
        + \mathcal{L}_{\text{incidence}}
    \right].
\end{equation*}

\noindent \textbf{Graph inpainting}.
During generation, we use inpainting to enforce constraints: \textit{i}) node budget splits are fixed to 1 for unexpanded clusters or those of size $1$, \textit{ii}) equal splits are enforced for expanded clusters with size 2, \textit{iii}) clusters of size one are not allowed to expand; and \textit{iv}) features of non-expanded clusters are copied unchanged.

\newpage

\noindent \textbf{Prior distributions and targets}.  
We use different prior distributions depending on the task:
\begin{itemize}[leftmargin=0.5cm]
    \itemsep0em 
    \item \textbf{Node and edge predictions:} Prior samples \( p_0 \) are drawn from a Gaussian distribution. Targets are either \( -1 \) or \( 1 \), indicating binary decisions (\eg whether a node is expanded or an edge is retained).
    
    \item \textbf{Hyperedge expansion:} Similar to node and edge predictions, with targets \( -1 \), \( 0 \), or \( 1 \), encoding the number of expansions (none, one, or two).
    
    \item \textbf{node budget fractions:} Prior samples \( p_0 \) are drawn from a Dirichlet distribution with concentration parameter \( \alpha = 1.5 \), linearly mapped to \( [-1, 1] \) via \( 2x - 1 \).
    The target is the node budget fraction of each child node, linearly mapped to \( [-1, 1] \) via \( 2x - 1 \). 
    Parent cluster node budgets are encoded using sinusoidal positional encodings~\cite{vaswani2023attentionneed} (dimension 32, base frequency \( 10^{-4} \)).
    
    \item \textbf{Feature generation:} Following~\cite{ren2024flowarscalewiseautoregressiveimage}, we draw prior features from a Gaussian and predict true node features, conditioned on the parent node's feature using a FiLM layer~\cite{perez2017filmvisualreasoninggeneral}.
\end{itemize}

\noindent \textbf{Simplex projection via Von Neumann method}.
When modeling node budget fractions, predictions must lie on the probability simplex. 
To ensure this, during generation, we project the model's outputs using the Von Neumann projection~\cite{projectionsimplex}, which finds the closest point (in Euclidean distance) on the simplex:
\begin{equation}
\Delta^K = \left\{ \mathbf{X} \in \mathbb{R}^K \mid x_i \geq 0, \sum_{i=1}^K x_i = 1 \right\}.
\end{equation}

\begin{enumerate}[label=\roman*), itemsep=0.3em, font=\itshape]
    \item Sort \( \mathbf{z} \in \mathbb{R}^K \) into a descending vector \( \mathbf{u} \), such that \( u_1 \geq u_2 \geq \dots \geq u_K \).
    
    \item Find the smallest index \( \rho \in \{1, \dots, K\} \) such that:
    \begin{equation}
    u_\rho - \frac{1}{\rho} \left( \sum_{j=1}^{\rho} u_j - 1 \right) > 0.
    \end{equation}
    
    \item Compute the threshold:
    \begin{equation}
    \tau = \frac{1}{\rho} \left( \sum_{j=1}^{\rho} u_j - 1 \right).
    \end{equation}
    
    \item The projection is then:
    \begin{equation}
    \mathbf{X}^* = \max(\mathbf{z} - \tau, 0).
    \end{equation}
\end{enumerate}

\subsection{Model Architecture} \label{sec::model_architecture}

Our method represents the expansion numbers for left and right nodes, along with edge presence, as attributes of the bipartite graph.
To model the distribution $p(\mathbf{v}^{(l)}_L, \mathbf{v}^{(l)}_R, \mathbf{e}^{(l)}, \mathbf{f}, \mathbf{F}_L^{\text{refine}} \mid \tilde{B}^{(l)})$, we adopt an endpoint-parameterized flow-matching framework \cite{lipman2023flowmatchinggenerativemodeling}.
Within this framework, the attributes—namely, the expansion vectors and edge indicators—are corrupted with noise, and a denoising network is trained to reconstruct the original values.

The denoising network is structured as follows:
\begin{enumerate}[leftmargin=0.5cm]
    \itemsep0em 
    \item \textbf{Positional encoding:} 
    Node positions within the graph are encoded using SignNet \cite{lim2022signbasisinvariantnetworks}. These encodings are replicated according to the respective expansion numbers.
    \item \textbf{Attribute embedding:}
    Five separate linear layers are used to embed the bipartite graph attributes: left node features, right node features, edge features, node-specific features, and hyperedge-specific features.
    FiLM conditioning \cite{perez2017filmvisualreasoninggeneral} is applied to incorporate contextual information into node and hyperedge features.
    Node budgets are embedded using sinusoidal positional encodings \cite{vaswani2023attentionneed}.

    \newpage
    
    \item \textbf{Feature concatenation:}
    \begin{itemize}
        \item For each left and right node, embeddings are concatenated with positional encodings and the desired reduction fraction. Left nodes also receive the node budget embedding.
        \item If node features are present, they are appended to the left nodes. Likewise, hyperedge features are appended to the right nodes when available.
        \item For edges, embeddings include edge features, concatenated positional encodings of the incident nodes, and the reduction fraction.
    \end{itemize}
    \item \textbf{Graph processing:}
    A succession of sparse PPGN layers process the bipartite graph, following the architecture from \cite{bergmeister2024efficient}.
    \item \textbf{Output prediction:}
    The final graph representations are passed through three linear projection heads to generate outputs.
    \begin{itemize}
        \item Left node head: Predicts expansion values, node budget splits, and refined node features.
        \item Right node head: Predicts hyperedge expansions and refined hyperedge features.
        \item Edge head: Predicts edge existence.
    \end{itemize}
\end{enumerate}

\subsection{Additional Details}
\label{sec:additional_tricks}

\noindent \textbf{Perturbed expansion.}
Following \cite{bergmeister2024efficient, gailhard2024hygenediffusionbasedhypergraphgeneration}, we augment Definitions \ref{def::expansion} and \ref{def::refinement} with additional randomness to improve generation quality. 
This modification is especially useful in low-data regimes where overfitting is a concern. Specifically, we supplement the set of edges $\tilde{\mathcal{E}}$ by randomly adding edges between node pairs on opposite sides of the bipartite graph that are within a fixed distance in $B$.
The following definition extends the expansion process (Definition \ref{def::expansion}) to include this stochastic perturbation.

\begin{definition}[Perturbed hypergraph expansion]
\label{def::perturbed_expansion}
Let $B = (\mathcal{V}_L, \mathcal{V}_R, \mathcal{E})$ be a bipartite graph, and let $\mathbf{v}_L \in \mathbb{N}^{|\mathcal{V}_L|}$ and $\mathbf{v}_R \in \mathbb{N}^{|\mathcal{V}_R|}$ denote the left and right cluster size vectors. For a given radius $r \in \mathbb{N}$ and probability $0 \leq p \leq 1$, we construct $\tilde{B}$ as in Definition \ref{def::expansion}. Additionally, for each pair of distinct nodes $\mathbf{v}_L{(p)} \in \tilde{\mathcal{V}_L}$ and $\mathbf{v}_R{(q)} \in \tilde{\mathcal{V}}_R$ that are within a distance of at most $2r+1$ in $B$, we independently add an edge $e_{\{p_i,q_j\}}$ to $\tilde{\mathcal{E}}$ with probability $p$.
\end{definition}

\noindent \textbf{Spectral conditioning.} 
In line with \cite{martinkus2022spectrespectralconditioninghelps, bergmeister2024efficient}, we incorporate spectral information—specifically, the principal eigenvalues and eigenvectors of the \textit{normalized} Laplacian—as a form of conditioning during the generative process. This technique has been shown to improve the quality of generated graphs. 
To generate $B^{(l)}$ from its coarser form $B^{(l+1)}$, we leverage the approximate spectral preservation under coarsening. We compute the $k$ smallest non-zero eigenvalues and their corresponding eigenvectors from the normalized Laplacian matrix $\mathbf{\mathcal{L}}^{(l+1)}$ of $B^{(l+1)}$. These eigenvectors are processed using SignNet \cite{lim2022signbasisinvariantnetworks} to produce node embeddings for $B^{(l+1)}$. These embeddings are then copied to the expanded nodes of $B^{(l)}$, which helps to identify clusters. The hyperparameter $k$ controls the number of spectral components used.

\noindent \textbf{Multi-scale graph OT coupling.}
Minibatch OT coupling \cite{tong2024improvinggeneralizingflowbasedgenerative, pooladian2023multisampleflowmatchingstraightening} improves training by jointly sampling priors and targets and reindexing priors to minimize input–output distance, leading to smoother flows that require fewer inference steps.  
As detailed in Section~\ref{sec:ot_coupling}, we adapt this idea by treating each bipartite graph as a minibatch of nodes and applying OT coupling within them.  
To preserve the distribution, we restrict permutations to groups of \emph{equivalent nodes} (identical topology and features, differing only in the prior noise sample).  
For efficiency, we assume equivalence only among nodes from the same cluster expansion.  
Since each cluster produces two or three children, coupling reduces to the evaluation of two or six possible permutations per cluster, which can be efficiently parallelized with tensors (Algorithm~\ref{alg:minibatch_ot}).

\begin{algorithm*}[hbtp]
\caption{\textbf{Multi-scale graph OT coupling:} Align targets and prior samples during refinement, processing all clusters independently and in parallel.}
\label{alg:minibatch_ot}
\begin{algorithmic}[1]

\STATE {\bfseries Input:} Expanded bipartite representation \(B\) with clusters \(\{V^{(p)}\}\), each \(V^{(p)}\) containing at most 3 nodes; target samples \(\{x_i\}\)
\STATE {\bfseries Output:} Reindexed noise samples \(\{\tilde{z}_i\}\)

\STATE Sample noise \(\{z_i\}\) for each node in \(B\)
\STATE Initialize \(\tilde{z}_i \gets z_i\) for all nodes \(i\)

\FORALL{clusters \(V^{(p)} \in B\)}
    \IF{\(|V^{(p)}| > 1\)}
        \STATE Let \(I_p = (i_1, \ldots, i_k)\) be the node indices in \(V^{(p)}\), where \(k = |V^{(p)}| \in \{2,3\}\)
        \STATE Let \(\mathfrak{S}_k\) denote the set of all permutations of \(\{1,\ldots,k\}\)

        \STATE Select the minimum-cost permutation:
        \[
        \sigma^*
        \gets
        \arg\min_{\sigma \in \mathfrak{S}_k}
        \sum_{r=1}^{k}
        \left\|
            z_{i_{\sigma(r)}} - x_{i_r}
        \right\|^2
        \]

        \STATE Reindex the noise samples according to \(\sigma^*\):
        \[
        \tilde{z}_{i_r}
        \gets
        z_{i_{\sigma^*(r)}},
        \qquad r=1,\ldots,k
        \]
    \ENDIF
\ENDFOR

\STATE \textbf{return} \(\{\tilde{z}_i\}\)

\end{algorithmic}
\end{algorithm*}

\clearpage
\newpage

\section{Training and Sampling Procedures} \label{app::training_sampling}

In this section, we present the complete training and inference procedures, detailed in Algorithms~\ref{alg:train} and~\ref{alg:sample_deterministic}.
Both pipelines rely on node embeddings produced by Algorithm~\ref{alg:embeddings}.

\begin{algorithm*}[hbtp]
\caption{\textbf{Node embedding computation:} Compute left and right node embeddings for a bipartite representation of a hypergraph. Embeddings are replicated according to cluster size vectors.}
\label{alg:embeddings}
\begin{algorithmic}[1]

\STATE {\bfseries Parameters:} number of spectral features \(k\)
\STATE {\bfseries Input:} bipartite representation \(B = (\mathcal{V}_L, \mathcal{V}_R, \mathcal{E})\), spectral feature model \(\text{SignNet}_\theta\), cluster size vectors \(\mathbf{v}_L\) and \(\mathbf{v}_R\)
\STATE {\bfseries Output:} node embeddings computed for all nodes in \(\mathcal{V}_L\) and \(\mathcal{V}_R\), replicated according to \(\mathbf{v}_L\) and \(\mathbf{v}_R\)

\STATE \(n_B \gets |\mathcal{V}_L| + |\mathcal{V}_R|\)

\IF{\(k = 0\)}
    \STATE Sample random embeddings:
    \[
        \mathbf{H} = [h^{(1)}, \ldots, h^{(n_B)}], 
        \qquad h^{(i)} \overset{i.i.d.}{\sim} \mathcal{N}(0, I)
    \]
\ELSE
    \STATE \(k_{\mathrm{eff}} \gets \min(k, n_B - 2)\)
    \IF{\(k_{\mathrm{eff}} > 0\)}
        \STATE Compute the \(k_{\mathrm{eff}}\) smallest non-trivial spectral features:
        \[
            [\lambda_1, \ldots, \lambda_{k_{\mathrm{eff}}}], 
            [u_1, \ldots, u_{k_{\mathrm{eff}}}]
            \gets \text{EIG}(B)
        \]
    \ENDIF
    \IF{\(k_{\mathrm{eff}} < k\)}
        \STATE Pad remaining spectral features with zeros:
        \[
            [\lambda_{k_{\mathrm{eff}}+1}, \ldots, \lambda_k], 
            [u_{k_{\mathrm{eff}}+1}, \ldots, u_k]
            \gets [0,\ldots,0], [0,\ldots,0]
        \]
    \ENDIF
    \STATE Compute embeddings with SignNet:
    \[
        \mathbf{H} = [h^{(1)}, \ldots, h^{(n_B)}]
        \gets \text{SignNet}_\theta(
        [\lambda_1, \ldots, \lambda_k],
        [u_1, \ldots, u_k], B)
    \]
\ENDIF

\STATE Expand bipartite representation according to cluster sizes:
\[
    \tilde{B}
    \gets \tilde{B}(B, \mathbf{v}_L, \mathbf{v}_R)
\]

\STATE Replicate embeddings according to cluster sizes:
\[
\tilde{\mathbf{H}}[v_L^{(p,i)}]
=
\mathbf{H}[v_L^{(p)}],
\quad
\forall p \in [|\mathcal{V}_L|],\;
1 \leq i \leq \mathbf{v}_L[p]
\]
\[
\tilde{\mathbf{H}}[v_R^{(p,i)}]
=
\mathbf{H}[v_R^{(p)}],
\quad
\forall p \in [|\mathcal{V}_R|],\;
1 \leq i \leq \mathbf{v}_R[p]
\]

\STATE \textbf{return} \(\tilde{\mathbf{H}}\)

\end{algorithmic}
\end{algorithm*}

\begin{algorithm*}[htbp]
\caption{\textbf{End-to-end training procedure:} Complete training procedure for our model.}
\label{alg:train}
\begin{algorithmic}[1]

\STATE {\bfseries Parameters:} number of spectral features \(k\) for node embeddings
\STATE {\bfseries Input:} dataset \(\mathcal{D} = \{H_1, \ldots, H_N\}\), denoising model \(\text{GNN}_\theta\), spectral feature model \(\text{SignNet}_\theta\)
\STATE {\bfseries Output:} trained model parameters \(\theta\)

\WHILE{not converged}
    \STATE Sample a graph \(H \sim \text{Uniform}(\mathcal{D})\)
    \STATE Sample coarsening sequence:
    \[
        (B^{(0)}, \ldots, B^{(L)}) \gets
        \text{RndRedSeq}(H)
        \text{ (Algorithm \ref{alg:hypergraph_coarsening})}
    \]
    \STATE Sample level \(l \sim \text{Uniform}(\{0, \ldots, L\})\)

    \IF{\(l = 0\)}
        \STATE Set \(\mathbf{v}_L^{(0)} \gets 1\), \(\mathbf{v}_R^{(0)} \gets 1\)
    \ELSE
        \STATE Set \(\mathbf{v}_L^{(l)}, \mathbf{v}_R^{(l)}\) such that
        $
        \tilde{B}(B^{(l)}, \mathbf{v}_L^{(l)}, \mathbf{v}_R^{(l)})
        $
        has the same node set as \(B^{(l-1)}\)
    \ENDIF

    \IF{\(l = L\)}
        \STATE Initialize the terminal bipartite representation:
        \[
            B^{(L+1)}
            \gets
            (\{1\}, \{2\}, \{(1,2)\},
            \mathbf{b}=[|\mathcal{V}_L^{(0)}|],
            \mathbf{F}_L=0,
            \mathbf{F}_R=0)
        \]
        \STATE
        \(\mathbf{v}_L^{(L+1)} \gets 1\),
        \(\mathbf{v}_R^{(L+1)} \gets 1\)
    \ENDIF

    \STATE Set
    \(\mathbf{e}^{(l)}, \mathbf{f}^{(l)},
    \mathbf{F}_L^{\text{refine},(l)},
    \mathbf{F}_R^{\text{refine},(l)}\)
    such that
    $
        B\!\left(
        \tilde{B}(B^{(l+1)},
        \mathbf{v}_L^{(l+1)},
        \mathbf{v}_R^{(l+1)}),
        \mathbf{e}^{(l)},
        \mathbf{f}^{(l)},
        \mathbf{F}_L^{\text{refine},(l)},
        \mathbf{F}_R^{\text{refine},(l)}
        \right)
        =
        B^{(l)}
    $

    \STATE Compute node embeddings:
    \[
        \mathbf{H}^{(l)}
        \gets
        \text{Embeddings}
        \left(
        B^{(l+1)},
        \text{SignNet}_\theta,
        \mathbf{v}_L^{(l+1)},
        \mathbf{v}_R^{(l+1)}
        \right)
    \]

    \STATE Compute reduction fraction:
    \[
        \hat{\rho}^{(l)}
        \gets
        1 -
        \frac{|\mathcal{V}_L^{(l+1)}|}
             {|\mathcal{V}_L^{(l)}|}
    \]

    \STATE Compute denoising output:
    \[
        D_\theta
        \gets
        \text{GNN}_\theta
        \left(
        \cdot,\cdot,
        \tilde{B}^{(l)},
        \mathbf{H}^{(l)},
        |\mathcal{V}_L^{(0)}|,
        \hat{\rho}^{(l)}
        \right)
    \]

    \STATE Take gradient descent step on
    $
    \nabla_\theta
    \text{DiffusionLoss}
    \left(
    \mathbf{v}_L^{(l)},
    \mathbf{v}_R^{(l)},
    \mathbf{e}^{(l)},
    \mathbf{f}^{(l)},
    \mathbf{F}_L^{\text{refine},(l)},
    \mathbf{F}_R^{\text{refine},(l)},
    D_\theta
    \right)
    $
\ENDWHILE

\STATE \textbf{return} \(\theta\)

\end{algorithmic}
\end{algorithm*}

\begin{algorithm*}[htbp]
\caption{\textbf{End-to-end sampling procedure:}
This describes the sampling procedure, assuming maximum cluster sizes of 2 for nodes and 3 for hyperedges.}
\label{alg:sample_deterministic}
\begin{algorithmic}[1]

\STATE {\bfseries Parameters:} reduction fraction range
$[\rho_\text{min}, \rho_\text{max}]$
\STATE {\bfseries Input:} target hypergraph size $N$, denoising model
$\text{GNN}_\theta$, spectral feature model $\text{SignNet}_\theta$
\STATE {\bfseries Output:} sampled hypergraph
$H=(\mathcal{V},\mathcal{E})$ with $|\mathcal{V}|=N$

\STATE Initialize minimal bipartite graph:
\[
B=(\mathcal{V}_L,\mathcal{V}_R,\mathcal{E},
\mathbf{b},\mathbf{F}_L,\mathbf{F}_R)
\gets
(\{1\},\{2\},\{(1,2)\},[N],0,0)
\]
\STATE Initialize cluster size vectors:
$\mathbf{v}_L\gets[1]$, $\mathbf{v}_R\gets[1]$

\WHILE{$|\mathcal{V}_L|<N$}

    \STATE Compute node embeddings:
    \[
        \mathbf{H}
        \gets
        \text{Embeddings}
        (B,\text{SignNet}_\theta,\mathbf{v}_L,\mathbf{v}_R)
    \]

    \STATE Expand current bipartite representation:
    \[
        \tilde B
        \gets
        \tilde B(B,\mathbf{v}_L,\mathbf{v}_R)
    \]

    \STATE $n\gets|\tilde{\mathcal{V}}_L|$

    \STATE Sample reduction fraction:
    $\rho\sim\text{Uniform}([\rho_\text{min},\rho_\text{max}])$

    \STATE Compute target expanded size and number of nodes to add:
    \[
        n_{\mathrm{next}}
        \gets
        \min\!\left(
        \max\!\left(
        \left\lceil\frac{n}{1-\rho}\right\rceil,n+1
        \right),N
        \right),
        \qquad
        n^+\gets n_{\mathrm{next}}-n
    \]

    \STATE Actual reduction fraction:
    \[
        \hat{\rho}
        \gets
        1-\frac{n}{n_{\mathrm{next}}}
    \]

    \STATE Compute denoising output:
    \[
        D_\theta
        \gets
        \text{GNN}_\theta
        (\cdot,\cdot,\tilde B,\mathbf{H},N,\hat{\rho})
    \]

    \STATE Sample refinement predictions:
    \[
        (\mathbf{v}_L)_0,(\mathbf{v}_R)_0,
        (\mathbf{e})_0,\mathbf{f},
        \mathbf{F}_L^{\text{refine}},
        \mathbf{F}_R^{\text{refine}}
        \gets
        \text{Sample}(D_\theta)
    \]

    \STATE Update left-side cluster sizes by selecting the $n^+$ largest scores:
    \[
        \mathbf{v}_L[i]
        =
        \begin{cases}
            2 &
            \text{if }
            \left|
            \{j\mid(\mathbf{v}_L)_0[j]
            \geq(\mathbf{v}_L)_0[i]\}
            \right|
            \leq n^+ \\
            1 & \text{otherwise}
        \end{cases}
    \]

    \STATE Update right-side cluster sizes:
    \[
        \mathbf{v}_R[i]
        =
        \begin{cases}
            1 & (\mathbf{v}_R)_0[i] < 1.66 \\
            2 & 1.66 \leq (\mathbf{v}_R)_0[i] < 2.33 \\
            3 & \text{otherwise}
        \end{cases}
    \]

    \STATE Update incidences:
    \[
        \mathbf{e}[i]
        =
        \begin{cases}
            1 & (\mathbf{e})_0[i]>0.5 \\
            0 & \text{otherwise}
        \end{cases}
    \]

    \STATE Refine bipartite graph:
    \[
        B
        \gets
        B\!\left(
        \tilde B,\mathbf{e},\mathbf{f},
        \mathbf{F}_L^{\text{refine}},
        \mathbf{F}_R^{\text{refine}}
        \right)
    \]

\ENDWHILE

\STATE Build hypergraph $H$ from its bipartite representation $B$
\STATE \textbf{return} $H$

\end{algorithmic}
\end{algorithm*}

\newpage
\clearpage

\section{Complexity Analysis} \label{app:complexity}

In this section, we investigate the asymptotic complexity of our proposed algorithm, which extends the methodology introduced by \citet{bergmeister2024efficient} and \citet{gailhard2024hygenediffusionbasedhypergraphgeneration}.
To construct a hypergraph comprising $n$ nodes, $m$ hyperedges, and $k$ incidences, the algorithm sequentially produces a series of bipartite graphs
$B^{(0)} = (\{1\}, \{2\}, \{(1,2)\}), B^{(1)}, \dots, B^{(L)} = B$,
where the final graph $B$ corresponds to the bipartite representation of the generated hypergraph.
We use $n$, $m$, and $k$ to denote, respectively, the number of nodes, hyperedges, and incidences in the hypergraph, corresponding to the number of left-side nodes, right-side nodes, and edges in the corresponding bipartite graph.

For each level $1 \leq l \leq L$ of the sequence, the number of left-side nodes in $B^{(l)}$, denoted $n_l$, satisfies
$n_l \geq (1+\epsilon)n_{l-1}$
for some $\epsilon > 0$ (\eg $\epsilon = \texttt{reduction\_frac}/(1-\texttt{reduction\_frac})$).
This implies an upper bound on the number of steps in the expansion sequence:
$\lceil \log_{1+\epsilon} n \rceil \in \mathcal{O}(\log n)$.
Since the expansion process only increases node counts, all $B^{(l)}$ contain at most $n$ left-side and $m$ right-side nodes.
The number of edges, however, may temporarily exceed $k$, as intermediate bipartite graphs may include an excess of edges removed in later refinements.
Still, because coarsening during training consistently reduces incidences, the model is expected to learn accurate edge refinement and avoid this kind of accumulation.
Consequently, we assume $k_l \leq k$ and $m_l \leq m$ for all $0 \leq l \leq L$.

Next, we assess the computational cost of generating a single expansion step.
At level $l=0$, this consists of creating a pair of connected nodes, initializing features as matrices of zeros, initializing the node budget as the target node count, and predicting the initial expansion vectors $\mathbf{v}_L^{(0)}$ and $\mathbf{v}_R^{(0)}$---a process with constant complexity $\mathcal{O}(1)$.
For levels $0 \leq l < L$, given $B^{(l)}$ and expansion vectors $\mathbf{v}_L^{(l)}$, $\mathbf{v}_R^{(l)}$, the algorithm constructs the expanded bipartite graph
$\Tilde{B}(B^{(l)}, \mathbf{v}_L^{(l)}, \mathbf{v}_R^{(l)})$
in $\mathcal{O}(n+m)$ time.
It then samples $\mathbf{v}_L^{(l+1)}$, $\mathbf{v}_R^{(l+1)}$, $\mathbf{e}^{(l+1)}$, $\mathbf{f}$, $\mathbf{F}_L^{\text{refine}}$, and $\mathbf{F}_R^{\text{refine}}$, and constructs the refined graph
$B^{(l+1)} =
B(\Tilde{B}^{(l)}, \mathbf{e}^{(l+1)}, \mathbf{f},
\mathbf{F}_L^{\text{refine}}, \mathbf{F}_R^{\text{refine}})$.
Letting $v^L_{\max}$ and $v^R_{\max}$ be the maximum cluster sizes, the incidence count in the expanded graph is bounded by
$\tilde{k}_l \leq k_l v^L_{\max}v^R_{\max}$.

The sampling process queries a denoising model a constant number of times per step.
The complexity is thus governed by the architecture.
In our case, since bipartite graphs are triangle-free, the \textit{Local PPGN} model \cite{bergmeister2024efficient} has linear complexity $\mathcal{O}(n+m+k)$.
Embedding computation for $B^{(l)}$ similarly costs $\mathcal{O}(n+m+k)$.
This includes calculating the top $K$ eigenvalues/eigenvectors of the Laplacian via the method by \citet{vishnoi2013lx}, with complexity
$\tilde{\mathcal{O}}\left(Kk_l\right)$,
where $\tilde{\mathcal{O}}$ hides polylogarithmic factors.
Embedding via \textit{SignNet} is linear in graph size, \ie
$\mathcal{O}(K(n_l+m_l))$.

The final transformation from the bipartite graph to a hypergraph---by collapsing right-side nodes into hyperedges---has a cost of $\mathcal{O}(m+k)$.
Under these assumptions, the total complexity to generate a hypergraph $H$ with $n$ nodes, $m$ hyperedges, and $k$ incidences is
$\tilde{\mathcal{O}}((n+m+k)\log n)$,
again hiding polylogarithmic factors.

For graphs, the same reasoning can be used to obtain a total complexity of
$\tilde{\mathcal{O}}((n+k)\log n)$,
where $n$ and $k$ are respectively the number of nodes and the number of edges in the final graph.
However, one additional assumption is required: the graph contains at most $\mathcal{O}(k)$ triangles, which is needed to ensure linear complexity for the \textit{Local PPGN} layers.

\newpage

\section{Detailed Numerical Results}
\label{sec:detailed_results}

In this section, we present detailed numerical results for all datasets and metrics described in Appendix \ref{app:experiments}.
Reported values of the form $a \pm b$ indicate the mean $a$ and twice the standard deviation $b$ computed over 5 runs.
The best and second-best results are highlighted in \textbf{bold} and \underline{underlined}, respectively.

\subsection{Comparisons with the Baselines}
\begin{table*}[h]
\centering
\caption{Detailed numerical results for unfeatured hypergraphs.}
\setlength{\tabcolsep}{5pt}
\resizebox{\textwidth}{!}{
\begin{tabular}{@{}cccccccccccc@{}}
\toprule
\multicolumn{12}{c}{\textbf{SBM Hypergraphs}} \\
\midrule
\textbf{Method} &
 \textbf{Valid} (\%) $\uparrow$ &
 \textbf{NumDiff} $\downarrow$ &
 \textbf{Deg} $\downarrow$ &
 \textbf{EdgeSize} $\downarrow$ &
 \textbf{Spectral} $\downarrow$ &
 \textbf{Harmonic} $\downarrow$ &
 \textbf{Closeness} $\downarrow$ &
 \textbf{Betweenness} $\downarrow$ \\
\midrule
    HyperPA \cite{Do_2020}  &
 $2.5$ &
 $\underline{0.075}$ &
 $4.062$ &
 $0.407$ &
 $0.273$ &
 $77.840$ &
 $0.074$ &
 $0.008$ \\
    VAE \cite{kingma2022autoencodingvariationalbayes} &
 $0.0$ &
 $0.375$ &
 $1.280$ &
 $1.059$ &
 $0.024$ &
 $6.543$ &
 $\mathbf{0.007}$ &
 $0.006$ \\
    GAN \cite{goodfellow2014generativeadversarialnetworks} &
 $0.0$ &
 $1.200$ &
 $2.106$ &
 $1.203$ &
 $0.059$ &
 $10.700$ &
 $0.076$ &
 $0.012$ \\
    Diffusion \cite{ho2020denoisingdiffusionprobabilisticmodels} &
 $0.0$ &
 $0.150$ &
 $1.717$ &
 $1.390$ &
 $0.031$ &
 $13.940$ &
 $0.040$ &
 $0.004$ \\
    HYGENE \cite{gailhard2024hygenediffusionbasedhypergraphgeneration} &
 $\underline{65.0}$ &
 $0.525$ &
 $\mathbf{0.321}$ &
 $\mathbf{0.002}$ &
 $\underline{0.010}$ &
 $\mathbf{2.990}$ &
 $0.016$ &
 $\mathbf{0.000}$ \\
    \cdashline{1-12}\\[-0.8em]
    \method &
 $\mathbf{87.8 \vcenter{\hbox{\scriptsize $\boldsymbol{\pm 3.1}$}}}$ &
 $\mathbf{0.029 \vcenter{\hbox{\scriptsize $\boldsymbol{\pm 0.009}$}}}$ &
 $\underline{0.846 \vcenter{\hbox{\scriptsize $\pm 0.457$}}}$ &
 $\underline{0.005 \vcenter{\hbox{\scriptsize $\pm 0.003$}}}$ &
 $\mathbf{0.006 \vcenter{\hbox{\scriptsize $\boldsymbol{\pm 0.004}$}}}$ &
 $\underline{6.410 \vcenter{\hbox{\scriptsize $\pm 3.124$}}}$ &
 $\underline{0.009 \vcenter{\hbox{\scriptsize $\pm 0.006$}}}$ &
 $\underline{0.003 \vcenter{\hbox{\scriptsize $\pm 0.001$}}}$ \\
\midrule
\multicolumn{12}{c}{\textbf{Ego Hypergraphs}} \\
\midrule
\textbf{Method} &
 \textbf{Valid} (\%) $\uparrow$ &
 \textbf{NumDiff} $\downarrow$ &
 \textbf{Deg} $\downarrow$ &
 \textbf{EdgeSize} $\downarrow$ &
 \textbf{Spectral} $\downarrow$ &
 \textbf{Harmonic} $\downarrow$ &
 \textbf{Closeness} $\downarrow$ &
 \textbf{Betweenness} $\downarrow$ \\
    \midrule
    HyperPA \cite{Do_2020}  &
 $0.0$ &
 $35.830$ &
 $2.590$ &
 $0.423$ &
 $0.237$ &
 $143.000$ &
 $0.354$ &
 $0.002$ \\
    VAE \cite{kingma2022autoencodingvariationalbayes} &
 $0.0$ &
 $47.580$ &
 $0.803$ &
 $1.458$ &
 $0.133$ &
 $38.950$ &
 $0.558$ &
 $0.019$ \\
    GAN \cite{goodfellow2014generativeadversarialnetworks} &
 $0.0$ &
 $60.350$ &
 $0.917$ &
 $1.665$ &
 $0.230$ &
 $41.800$ &
 $0.612$ &
 $0.015$ \\
    Diffusion \cite{ho2020denoisingdiffusionprobabilisticmodels} &
 $0.0$ &
 $\underline{4.475}$ &
 $3.984$ &
 $2.985$ &
 $0.190$ &
 $6.911$ &
 $0.407$ &
 $0.009$ \\
    HYGENE \cite{gailhard2024hygenediffusionbasedhypergraphgeneration} &
 $\underline{90.0}$ &
 $12.550$ &
 $\mathbf{0.063}$ &
 $\underline{0.220}$ &
 $\mathbf{0.004}$ &
 $\underline{5.790}$ &
 $\underline{0.025}$ &
 $\mathbf{0.000}$ \\
    \cdashline{1-12}\\[-0.8em]
    \method &
 $\mathbf{99.5 \vcenter{\hbox{\scriptsize $\boldsymbol{\pm 1.1}$}}}$ &
 $\mathbf{0.128 \vcenter{\hbox{\scriptsize $\boldsymbol{\pm 0.171}$}}}$ &
 $\underline{0.124 \vcenter{\hbox{\scriptsize $\pm 0.086$}}}$ &
 $\mathbf{0.155 \vcenter{\hbox{\scriptsize $\boldsymbol{\pm 0.067}$}}}$ &
 $\mathbf{0.004 \vcenter{\hbox{\scriptsize $\boldsymbol{\pm 0.003}$}}}$ &
 $\mathbf{2.703 \vcenter{\hbox{\scriptsize $\boldsymbol{\pm 3.468}$}}}$ &
 $\mathbf{0.003 \vcenter{\hbox{\scriptsize $\boldsymbol{\pm 0.005}$}}}$ &
 $\mathbf{0.000 \vcenter{\hbox{\scriptsize $\boldsymbol{\pm 0.000}$}}}$ \\
\midrule
\multicolumn{12}{c}{\textbf{Tree Hypergraphs}} \\
\midrule
\textbf{Method} &
 \textbf{Valid} (\%) $\uparrow$ &
 \textbf{NumDiff} $\downarrow$ &
 \textbf{Deg} $\downarrow$ &
 \textbf{EdgeSize} $\downarrow$ &
 \textbf{Spectral} $\downarrow$ &
 \textbf{Harmonic} $\downarrow$ &
 \textbf{Closeness} $\downarrow$ &
 \textbf{Betweenness} $\downarrow$ \\
\midrule
    HyperPA \cite{Do_2020} &
 $0.0$ &
 $2.350$ &
 $0.315$ &
 $0.284$ &
 $0.159$ &
 $5.941$ &
 $0.477$ &
 $0.168$ \\
    VAE \cite{kingma2022autoencodingvariationalbayes} &
 $0.0$ &
 $9.700$ &
 $0.072$ &
 $0.480$ &
 $0.124$ &
 $3.869$ &
 $0.280$ &
 $0.139$ \\
    GAN \cite{goodfellow2014generativeadversarialnetworks} &
 $0.0$ &
 $6.000$ &
 $0.151$ &
 $0.469$ &
 $0.089$ &
 $2.198$ &
 $0.201$ &
 $0.124$ \\
    Diffusion \cite{ho2020denoisingdiffusionprobabilisticmodels} &
 $0.0$ &
 $2.225$ &
 $1.718$ &
 $1.922$ &
 $0.127$ &
 $8.565$ &
 $0.353$ &
 $0.139$ \\
    HYGENE \cite{gailhard2024hygenediffusionbasedhypergraphgeneration} &
 $\underline{77.5}$ &
 $\mathbf{0.000}$ &
 $\underline{0.059}$ &
 $\underline{0.108}$ &
 $\underline{0.012}$ &
 $\underline{1.099}$ &
 $\underline{0.041}$ &
 $\underline{0.016}$ \\
    \cdashline{1-12}\\[-0.8em]
    \method &
 $\mathbf{89.7 \vcenter{\hbox{\scriptsize $\boldsymbol{\pm 6.0}$}}}$ &
 $\mathbf{0.000 \vcenter{\hbox{\scriptsize $\boldsymbol{\pm 0.000}$}}}$ &
 $\mathbf{0.022 \vcenter{\hbox{\scriptsize $\boldsymbol{\pm 0.022}$}}}$ &
 $\mathbf{0.030 \vcenter{\hbox{\scriptsize $\boldsymbol{\pm 0.034}$}}}$ &
 $\mathbf{0.003 \vcenter{\hbox{\scriptsize $\boldsymbol{\pm 0.002}$}}}$ &
 $\mathbf{0.171 \vcenter{\hbox{\scriptsize $\boldsymbol{\pm 0.106}$}}}$ &
 $\mathbf{0.014 \vcenter{\hbox{\scriptsize $\boldsymbol{\pm 0.006}$}}}$ &
 $\mathbf{0.014 \vcenter{\hbox{\scriptsize $\boldsymbol{\pm 0.004}$}}}$ \\
\midrule
\multicolumn{12}{c}{\textbf{ModelNet Bookshelf}} \\
\midrule
\textbf{Method} &
 &
\textbf{NumDiff} $\downarrow$ &
 \textbf{Deg} $\downarrow$ &
 \textbf{EdgeSize} $\downarrow$ &
 \textbf{Spectral} $\downarrow$ &
 \textbf{Harmonic} $\downarrow$ &
 \textbf{Closeness} $\downarrow$ &
 \textbf{Betweenness} $\downarrow$ \\
\midrule
    HyperPA \cite{Do_2020} &
&
 $8.025$ &
 $7.562$ &
 $0.044$ &
 $\underline{0.048}$ &
 $877.500$ &
 $0.211$ &
 $0.005$ \\
    VAE \cite{kingma2022autoencodingvariationalbayes} &
&
 $47.450$ &
 $6.190$ &
 $1.520$ &
 $0.190$ &
 $113.600$ &
 $\underline{0.145}$ &
 $\underline{0.003}$ \\
    GAN \cite{goodfellow2014generativeadversarialnetworks} &
&
 $\mathbf{0.000}$ &
 $397.200$ &
 $46.300$ &
 $0.476$ &
 $670.100$ &
 $0.707$ &
 $0.007$ \\
    Diffusion \cite{ho2020denoisingdiffusionprobabilisticmodels} &
&
 $\mathbf{0.000}$ &
 $20.360$ &
 $2.346$ &
 $0.079$ &
 $264.100$ &
 $0.239$ &
 $0.006$ \\
    HYGENE \cite{gailhard2024hygenediffusionbasedhypergraphgeneration} &
&
 $69.730$ &
 $\mathbf{1.050}$ &
 $\underline{0.034}$ &
 $0.068$ &
 $\mathbf{27.400}$ &
 $0.204$ &
 $0.004$ \\
    \cdashline{1-12}\\[-0.8em]
    \method &
&
 $0.135 \vcenter{\hbox{\scriptsize $\pm 0.276$}}$ &
 $\underline{2.980 \vcenter{\hbox{\scriptsize $\pm 2.107$}}}$ &
 $\mathbf{0.020 \vcenter{\hbox{\scriptsize $\boldsymbol{\pm 0.025}$}}}$ &
 $\mathbf{0.024 \vcenter{\hbox{\scriptsize $\boldsymbol{\pm 0.015}$}}}$ &
 $\underline{46.614 \vcenter{\hbox{\scriptsize $\pm 58.803$}}}$ &
 $\mathbf{0.086 \vcenter{\hbox{\scriptsize $\boldsymbol{\pm 0.030}$}}}$ &
 $\mathbf{0.001 \vcenter{\hbox{\scriptsize $\boldsymbol{\pm 0.001}$}}}$ \\
\midrule
\multicolumn{12}{c}{\textbf{ModelNet Piano}} \\
\midrule
\textbf{Method} &
 &
\textbf{NumDiff} $\downarrow$ &
 \textbf{Deg} $\downarrow$ &
 \textbf{EdgeSize} $\downarrow$ &
 \textbf{Spectral} $\downarrow$ &
 \textbf{Harmonic} $\downarrow$ &
 \textbf{Closeness} $\downarrow$ &
 \textbf{Betweenness} $\downarrow$ \\
\midrule
    HyperPA \cite{Do_2020}  &
&
 $0.825$ &
 $9.254$ &
 $\mathbf{0.023}$ &
 $\underline{0.067}$ &
 $\mathbf{77.840}$ &
 $\underline{0.236}$ &
 $0.004$ \\
    VAE \cite{kingma2022autoencodingvariationalbayes} &
&
 $75.350$ &
 $8.060$ &
 $1.686$ &
 $0.396$ &
 $184.300$ &
 $0.241$ &
 $0.003$ \\
    GAN \cite{goodfellow2014generativeadversarialnetworks} &
&
 $\mathbf{0.000}$ &
 $409.000$ &
 $86.380$ &
 $0.697$ &
 $622.200$ &
 $0.738$ &
 $0.005$ \\
    Diffusion \cite{ho2020denoisingdiffusionprobabilisticmodels} &
&
 $\underline{0.050}$ &
 $20.900$ &
 $4.192$ &
 $0.113$ &
 $289.300$ &
 $0.303$ &
 $0.004$ \\
    HYGENE \cite{gailhard2024hygenediffusionbasedhypergraphgeneration} &
&
 $42.520$ &
 $\underline{6.290}$ &
 $\underline{0.027}$ &
 $0.117$ &
 $155.000$ &
 $0.285$ &
 $\mathbf{0.002}$ \\
    \cdashline{1-12}\\[-0.8em]
    \method &
&
 $0.846 \vcenter{\hbox{\scriptsize $\pm 1.009$}}$ &
 $\mathbf{3.265 \vcenter{\hbox{\scriptsize $\boldsymbol{\pm 1.954}$}}}$ &
 $0.042 \vcenter{\hbox{\scriptsize $\pm 0.056$}}$ &
 $\mathbf{0.040 \vcenter{\hbox{\scriptsize $\boldsymbol{\pm 0.026}$}}}$ &
 $\underline{119.158 \vcenter{\hbox{\scriptsize $\pm 81.023$}}}$ &
 $\mathbf{0.123 \vcenter{\hbox{\scriptsize $\boldsymbol{\pm 0.119}$}}}$ &
 $\mathbf{0.002 \vcenter{\hbox{\scriptsize $\boldsymbol{\pm 0.002}$}}}$  \\
\bottomrule
\end{tabular}
}
\end{table*}

\begin{table*}[h!]
\centering
\caption{Detailed numerical results for 3D meshes.}
\setlength{\tabcolsep}{5pt}
\resizebox{\textwidth}{!}{
\begin{tabular}{@{}cccccccccccccc@{}}
\toprule
\multicolumn{12}{c}{\textbf{Manifold40 Airplane}} \\
\midrule
\textbf{Method} &
\textbf{ChamDist} $\downarrow$ &
\textbf{ChamCov} (\%) $\uparrow$ &
\textbf{ChamDiv} $\uparrow$ &
\textbf{NumDiff} $\downarrow$ &
 \textbf{Deg} $\downarrow$ &
 \textbf{EdgeSize} $\downarrow$ &
 \textbf{Spectral} $\downarrow$ &
 \textbf{Harmonic} $\downarrow$ &
 \textbf{Closeness} $\downarrow$ &
 \textbf{Betweenness} $\downarrow$ \\
\midrule
    Sequential &
 $\underline{0.082 \vcenter{\hbox{\scriptsize $\pm 0.074$}}}$ &
 $\mathbf{34.8 \vcenter{\hbox{\scriptsize $\boldsymbol{\pm 0.0}$}}}$ &
 $\underline{0.129 \vcenter{\hbox{\scriptsize $\pm 0.011$}}}$ &
 $\underline{0.116 \vcenter{\hbox{\scriptsize $\pm 0.074$}}}$ &
 $\underline{0.258 \vcenter{\hbox{\scriptsize $\pm 0.009$}}}$ &
 $\underline{0.010 \vcenter{\hbox{\scriptsize $\pm 0.004$}}}$ &
 $\mathbf{0.010 \vcenter{\hbox{\scriptsize $\boldsymbol{\pm 0.001}$}}}$ &
 $\underline{4.491 \vcenter{\hbox{\scriptsize $\pm 0.663$}}}$ &
 $\underline{0.058 \vcenter{\hbox{\scriptsize $\pm 0.005$}}}$ &
 $\underline{0.005 \vcenter{\hbox{\scriptsize $\pm 0.000$}}}$ \\
    WGAN &
 $0.124 \vcenter{\hbox{\scriptsize $\pm 0.021$}}$ &
 $10.4 \vcenter{\hbox{\scriptsize $\pm 3.5$}}$ &
 $0.037 \vcenter{\hbox{\scriptsize $\pm 0.016$}}$ &
 $1.435 \vcenter{\hbox{\scriptsize $\pm 0.000$}}$ &
 $68.128 \vcenter{\hbox{\scriptsize $\pm 14.602$}}$ &
 $36.138 \vcenter{\hbox{\scriptsize $\pm 7.739$}}$ &
 $0.614 \vcenter{\hbox{\scriptsize $\pm 0.071$}}$ &
 $57.422 \vcenter{\hbox{\scriptsize $\pm 0.000$}}$ &
 $0.655 \vcenter{\hbox{\scriptsize $\pm 0.000$}}$ &
 $0.020 \vcenter{\hbox{\scriptsize $\pm 0.000$}}$ \\
    Flow-Matching &
 $0.085 \vcenter{\hbox{\scriptsize $\pm 0.014$}}$ &
 $20.9 \vcenter{\hbox{\scriptsize $\pm 10.0$}}$ &
 $0.063 \vcenter{\hbox{\scriptsize $\pm 0.005$}}$ &
 $0.574 \vcenter{\hbox{\scriptsize $\pm 0.216$}}$ &
 $0.620 \vcenter{\hbox{\scriptsize $\pm 0.130$}}$ &
 $1.089 \vcenter{\hbox{\scriptsize $\pm 0.052$}}$ &
 $0.040 \vcenter{\hbox{\scriptsize $\pm 0.004$}}$ &
 $10.782 \vcenter{\hbox{\scriptsize $\pm 3.903$}}$ &
 $0.156 \vcenter{\hbox{\scriptsize $\pm 0.028$}}$ &
 $0.009 \vcenter{\hbox{\scriptsize $\pm 0.001$}}$ \\
    VAE &
 $0.231 \vcenter{\hbox{\scriptsize $\pm 0.208$}}$ &
 $26.9 \vcenter{\hbox{\scriptsize $\pm 11.7$}}$ &
 $\mathbf{0.191 \vcenter{\hbox{\scriptsize $\boldsymbol{\pm 0.130}$}}}$ &
 $1.603 \vcenter{\hbox{\scriptsize $\pm 1.340$}}$ &
 $35.488 \vcenter{\hbox{\scriptsize $\pm 70.110$}}$ &
 $18.896 \vcenter{\hbox{\scriptsize $\pm 35.756$}}$ &
 $0.328 \vcenter{\hbox{\scriptsize $\pm 0.588$}}$ &
 $36.716 \vcenter{\hbox{\scriptsize $\pm 38.851$}}$ &
 $0.425 \vcenter{\hbox{\scriptsize $\pm 0.431$}}$ &
 $0.015 \vcenter{\hbox{\scriptsize $\pm 0.007$}}$ \\
    \cdashline{1-12}\\[-0.8em]
    \method &
 $\mathbf{0.048 \vcenter{\hbox{\scriptsize $\boldsymbol{\pm 0.003}$}}}$ &
 $\underline{31.9 \vcenter{\hbox{\scriptsize $\pm 13.2$}}}$ &
 $0.101 \vcenter{\hbox{\scriptsize $\pm 0.011$}}$ &
 $\mathbf{0.017 \vcenter{\hbox{\scriptsize $\boldsymbol{\pm 0.070}$}}}$ &
 $\mathbf{0.218 \vcenter{\hbox{\scriptsize $\boldsymbol{\pm 0.085}$}}}$ &
 $\mathbf{0.004 \vcenter{\hbox{\scriptsize $\boldsymbol{\pm 0.007}$}}}$ &
 $\mathbf{0.010 \vcenter{\hbox{\scriptsize $\boldsymbol{\pm 0.003}$}}}$ &
 $\mathbf{2.428 \vcenter{\hbox{\scriptsize $\boldsymbol{\pm 1.455}$}}}$ &
 $\mathbf{0.032 \vcenter{\hbox{\scriptsize $\boldsymbol{\pm 0.007}$}}}$ &
 $\mathbf{0.003 \vcenter{\hbox{\scriptsize $\boldsymbol{\pm 0.001}$}}}$ \\
\midrule
\multicolumn{12}{c}{\textbf{Manifold40 Bench}} \\
\midrule
\textbf{Method} &
\textbf{ChamDist} $\downarrow$ &
\textbf{ChamCov} (\%) $\uparrow$ &
\textbf{ChamDiv} $\uparrow$  &
\textbf{NumDiff} $\downarrow$ &
 \textbf{Deg} $\downarrow$ &
 \textbf{EdgeSize} $\downarrow$ &
 \textbf{Spectral} $\downarrow$ &
 \textbf{Harmonic} $\downarrow$ &
 \textbf{Closeness} $\downarrow$ &
 \textbf{Betweenness} $\downarrow$ \\
\midrule
    Sequential &
 $0.169 \vcenter{\hbox{\scriptsize $\pm 0.007$}}$ &
 $13.3 \vcenter{\hbox{\scriptsize $\pm 0.0$}}$ &
 $0.157 \vcenter{\hbox{\scriptsize $\pm 0.003$}}$ &
 $\mathbf{0.022 \vcenter{\hbox{\scriptsize $\boldsymbol{\pm 0.031}$}}}$ &
 $\mathbf{0.340 \vcenter{\hbox{\scriptsize $\boldsymbol{\pm 0.052}$}}}$ &
 $\underline{0.004 \vcenter{\hbox{\scriptsize $\pm 0.003$}}}$ &
 $\mathbf{0.007 \vcenter{\hbox{\scriptsize $\boldsymbol{\pm 0.000}$}}}$ &
 $\mathbf{3.694 \vcenter{\hbox{\scriptsize $\boldsymbol{\pm 0.510}$}}}$ &
 $\mathbf{0.046 \vcenter{\hbox{\scriptsize $\boldsymbol{\pm 0.009}$}}}$ &
 $\mathbf{0.004 \vcenter{\hbox{\scriptsize $\boldsymbol{\pm 0.000}$}}}$ \\
    WGAN &
 $0.128 \vcenter{\hbox{\scriptsize $\pm 0.015$}}$ &
 $17.3 \vcenter{\hbox{\scriptsize $\pm 3.3$}}$ &
 $0.063 \vcenter{\hbox{\scriptsize $\pm 0.099$}}$ &
 $2.267 \vcenter{\hbox{\scriptsize $\pm 0.000$}}$ &
 $69.748 \vcenter{\hbox{\scriptsize $\pm 20.988$}}$ &
 $36.352 \vcenter{\hbox{\scriptsize $\pm 10.914$}}$ &
 $0.621 \vcenter{\hbox{\scriptsize $\pm 0.066$}}$ &
 $56.108 \vcenter{\hbox{\scriptsize $\pm 0.000$}}$ &
 $0.640 \vcenter{\hbox{\scriptsize $\pm 0.000$}}$ &
 $0.019 \vcenter{\hbox{\scriptsize $\pm 0.000$}}$ \\
    Flow-Matching &
 $\underline{0.108 \vcenter{\hbox{\scriptsize $\pm 0.019$}}}$ &
 $\underline{30.7 \vcenter{\hbox{\scriptsize $\pm 2.5$}}}$ &
 $0.112 \vcenter{\hbox{\scriptsize $\pm 0.007$}}$ &
 $2.327 \vcenter{\hbox{\scriptsize $\pm 0.265$}}$ &
 $0.774 \vcenter{\hbox{\scriptsize $\pm 0.112$}}$ &
 $1.142 \vcenter{\hbox{\scriptsize $\pm 0.047$}}$ &
 $0.038 \vcenter{\hbox{\scriptsize $\pm 0.006$}}$ &
 $9.608 \vcenter{\hbox{\scriptsize $\pm 4.260$}}$ &
 $0.145 \vcenter{\hbox{\scriptsize $\pm 0.033$}}$ &
 $0.008 \vcenter{\hbox{\scriptsize $\pm 0.002$}}$ \\
    VAE &
 $0.280 \vcenter{\hbox{\scriptsize $\pm 0.223$}}$ &
 $27.5 \vcenter{\hbox{\scriptsize $\pm 9.9$}}$ &
 $\underline{0.195 \vcenter{\hbox{\scriptsize $\pm 0.107$}}}$ &
 $1.853 \vcenter{\hbox{\scriptsize $\pm 1.309$}}$ &
 $24.476 \vcenter{\hbox{\scriptsize $\pm 65.177$}}$ &
 $13.207 \vcenter{\hbox{\scriptsize $\pm 33.338$}}$ &
 $0.228 \vcenter{\hbox{\scriptsize $\pm 0.557$}}$ &
 $31.415 \vcenter{\hbox{\scriptsize $\pm 35.421$}}$ &
 $0.361 \vcenter{\hbox{\scriptsize $\pm 0.397$}}$ &
 $0.014 \vcenter{\hbox{\scriptsize $\pm 0.007$}}$ \\
    \cdashline{1-12}\\[-0.8em]
    \method &
$\mathbf{0.064 \vcenter{\hbox{\scriptsize $\boldsymbol{\pm 0.005}$}}}$ &
 $\mathbf{42.2 \vcenter{\hbox{\scriptsize $\boldsymbol{\pm 13.9}$}}}$ &
 $\mathbf{0.243 \vcenter{\hbox{\scriptsize $\boldsymbol{\pm 0.037}$}}}$ &
 $\underline{0.060 \vcenter{\hbox{\scriptsize $\pm 0.149$}}}$ &
 $\underline{0.349 \vcenter{\hbox{\scriptsize $\pm 0.454$}}}$ &
 $\mathbf{0.009 \vcenter{\hbox{\scriptsize $\boldsymbol{\pm 0.017}$}}}$ &
 $\underline{0.017 \vcenter{\hbox{\scriptsize $\pm 0.011$}}}$ &
 $\underline{5.069 \vcenter{\hbox{\scriptsize $\pm 8.171$}}}$ &
 $\mathbf{0.046 \vcenter{\hbox{\scriptsize $\boldsymbol{\pm 0.040}$}}}$ &
 $\mathbf{0.004 \vcenter{\hbox{\scriptsize $\boldsymbol{\pm 0.003}$}}}$ \\
\bottomrule
\end{tabular}
}
\end{table*}

\begin{table*}[h!]
\centering
\caption{Detailed numerical results for unfeatured graph datasets.}
\setlength{\tabcolsep}{5pt}
\resizebox{\textwidth}{!}{
\begin{tabular}{@{}ccccccccccc@{}}
\toprule
\multicolumn{11}{c}{\textbf{SBM Graphs}} \\
\midrule
\textbf{Method} &
 \textbf{Valid} (\%) $\uparrow$ &
 \textbf{Wavelet} $\downarrow$ &
 \textbf{Orbit} $\downarrow$ &
 \textbf{Clustering} $\downarrow$ &
 \textbf{Deg} $\downarrow$ &
 \textbf{Spectral} $\downarrow$ &
 \textbf{Ratio} $\downarrow$ \\
\midrule
    DiGress \cite{vignac2023digress} &
 $\underline{60.0}$ &
 $\mathbf{0.001}$ &
 \underline{$0.042$} &
 $\mathbf{0.049}$ &
 $\underline{0.002}$ &
 $\mathbf{0.004}$ &
 $\mathbf{1.7}$ \\
    DeFoG \cite{qin2025defogdiscreteflowmatching} &
 $\mathbf{90.0 \vcenter{\hbox{\scriptsize $\boldsymbol{\pm 5.1}$}}}$ &
 $\underline{0.008 \vcenter{\hbox{\scriptsize $\pm 0.002$}}}$ &
 $0.056 \vcenter{\hbox{\scriptsize $\pm 0.074$}}$ &
 $\underline{0.052 \vcenter{\hbox{\scriptsize $\pm 0.001$}}}$ &
 $\mathbf{0.001 \vcenter{\hbox{\scriptsize $\boldsymbol{\pm 0.002}$}}}$ &
 $0.005 \vcenter{\hbox{\scriptsize $\pm 0.001$}}$ &
 $4.9 \vcenter{\hbox{\scriptsize $\pm 1.3$}}$ \\
    HSpectre \cite{bergmeister2024efficient} &
 $45.0$ &
 $0.022$ &
 $0.067$ &
 $\underline{0.052}$ &
 $0.012$ &
 $0.007$ &
 $10.2$ \\
    BwR \citep{diamant2023improving} &
 $7.5$ &
 $0.089$ &
 $0.114$ &
 $0.064$ &
 $0.048$ &
 $0.0169$ &
 $38.6$ \\
    \cdashline{1-8}\\[-0.8em]
    \method &
 $50.0 \vcenter{\hbox{\scriptsize $\pm 5.0$}}$ &
 $\underline{0.008 \vcenter{\hbox{\scriptsize $\pm 0.003$}}}$ &
 $\mathbf{0.040 \vcenter{\hbox{\scriptsize $\boldsymbol{\pm 0.000}$}}}$ &
 $\underline{0.052 \vcenter{\hbox{\scriptsize $\pm 0.004$}}}$ &
 $0.007 \vcenter{\hbox{\scriptsize $\pm 0.000$}}$ &
 $\mathbf{0.004 \vcenter{\hbox{\scriptsize $\boldsymbol{\pm 0.002}$}}}$ &
 $\underline{4.8 \vcenter{\hbox{\scriptsize $\pm 0.7$}}}$ \\
\midrule
\multicolumn{11}{c}{\textbf{Tree Graphs}} \\
\midrule
\textbf{Method} &
 \textbf{Valid} (\%) $\uparrow$ &
 \textbf{Wavelet} $\downarrow$ &
 \textbf{Orbit} $\downarrow$ &
 \textbf{Clustering} $\downarrow$ &
 \textbf{Deg} $\downarrow$ &
 \textbf{Spectral} $\downarrow$ &
 \textbf{Ratio} $\downarrow$ \\
    \midrule
    DiGress \cite{vignac2023digress} &
 $90.0$ &
 $\mathbf{0.004}$ &
 $\mathbf{0.000}$ &
 $\mathbf{0.000}$ &
 $\mathbf{0.000}$ &
 $\mathbf{0.011}$ &
 $\mathbf{1.6}$ \\
    DeFoG \cite{qin2025defogdiscreteflowmatching} &
 $96.5 \vcenter{\hbox{\scriptsize $\pm 2.6$}}$ &
 $\underline{0.005 \vcenter{\hbox{\scriptsize $\pm 0.000$}}}$ &
 $\mathbf{0.000 \vcenter{\hbox{\scriptsize $\boldsymbol{\pm 0.000}$}}}$ &
 $\mathbf{0.000 \vcenter{\hbox{\scriptsize $\boldsymbol{\pm 0.000}$}}}$ &
 $\mathbf{0.000 \vcenter{\hbox{\scriptsize $\boldsymbol{\pm 0.000}$}}}$ &
 $\mathbf{0.011 \vcenter{\hbox{\scriptsize $\boldsymbol{\pm 0.003}$}}}$ &
 $\mathbf{1.6 \vcenter{\hbox{\scriptsize $\boldsymbol{\pm 0.4}$}}}$ \\
    HSpectre \cite{bergmeister2024efficient} &
 $\mathbf{100.0}$ &
 $\underline{0.005}$ &
 $\mathbf{0.000}$ &
 $\mathbf{0.000}$ &
 $\mathbf{0.000}$ &
 $0.012$ &
 $4.0$ \\
    BwR \citep{diamant2023improving} &
 $0.0$ &
 $0.039$ &
 $0.000$ &
 $0.124$ &
 $0.002$ &
 $0.048$ &
 $11.4$ \\
    \cdashline{1-8}\\[-0.8em]
    \method &
 $\mathbf{100.0 \vcenter{\hbox{\scriptsize $\boldsymbol{\pm 0.0}$}}}$ &
 $\underline{0.005 \vcenter{\hbox{\scriptsize $\pm 0.001$}}}$ &
 $\mathbf{0.000 \vcenter{\hbox{\scriptsize $\boldsymbol{\pm 0.000}$}}}$ &
 $\mathbf{0.000 \vcenter{\hbox{\scriptsize $\boldsymbol{\pm 0.000}$}}}$ &
 $\mathbf{0.000 \vcenter{\hbox{\scriptsize $\boldsymbol{\pm 0.000}$}}}$ &
 $0.012 \vcenter{\hbox{\scriptsize $\pm 0.004$}}$ &
 $1.8 \vcenter{\hbox{\scriptsize $\pm 0.8$}}$ \\
\midrule
\multicolumn{11}{c}{\textbf{Planar Graphs}} \\
\midrule
\textbf{Method} &
 \textbf{Valid} (\%) $\uparrow$ &
 \textbf{Wavelet} $\downarrow$ &
 \textbf{Orbit} $\downarrow$ &
 \textbf{Clustering} $\downarrow$ &
 \textbf{Deg} $\downarrow$ &
 \textbf{Spectral} $\downarrow$ &
 \textbf{Ratio} $\downarrow$ \\
\midrule
    DiGress \cite{vignac2023digress} &
 $77.5$ &
 $0.003$ &
 $0.008$ &
 $0.078$ &
 $0.001$ &
 $0.010$ &
 $5.1$ \\
    DeFoG \cite{qin2025defogdiscreteflowmatching} &
 $\mathbf{99.5 \vcenter{\hbox{\scriptsize $\boldsymbol{\pm 1.0}$}}}$ &
 $\mathbf{0.001 \vcenter{\hbox{\scriptsize $\boldsymbol{\pm 0.000}$}}}$ &
 $\mathbf{0.001 \vcenter{\hbox{\scriptsize $\boldsymbol{\pm 0.000}$}}}$ &
 $\underline{0.050 \vcenter{\hbox{\scriptsize $\pm 0.015$}}}$ &
 $\mathbf{0.001 \vcenter{\hbox{\scriptsize $\boldsymbol{\pm 0.000}$}}}$ &
 $\mathbf{0.007 \vcenter{\hbox{\scriptsize $\boldsymbol{\pm 0.001}$}}}$ &
 $\mathbf{1.6 \vcenter{\hbox{\scriptsize $\boldsymbol{\pm 0.4}$}}}$ \\
    HSpectre \cite{bergmeister2024efficient} &
 $95.0$ &
 $\mathbf{0.001}$ &
 $\underline{0.002}$ &
 $0.063$ &
 $\mathbf{0.001}$ &
 $\underline{0.007}$ &
 $\underline{2.1}$ \\
    BwR \citep{diamant2023improving} &
 $0.0$ &
 $0.131$ &
 $0.547$ &
 $0.260$ &
 $0.023$ &
 $0.044$ &
 $251.9$ \\
    \cdashline{1-8}\\[-0.8em]
    \method &
 $\underline{96.7 \vcenter{\hbox{\scriptsize $\pm 6.2$}}}$ &
 $\mathbf{0.001 \vcenter{\hbox{\scriptsize $\boldsymbol{\pm 0.001}$}}}$ &
 $0.003 \vcenter{\hbox{\scriptsize $\pm 0.005$}}$ &
 $\mathbf{0.042 \vcenter{\hbox{\scriptsize $\boldsymbol{\pm 0.002}$}}}$ &
 $\mathbf{0.000 \vcenter{\hbox{\scriptsize $\boldsymbol{\pm 0.000}$}}}$ &
 $\underline{0.008 \vcenter{\hbox{\scriptsize $\pm 0.003$}}}$ &
 $2.2 \vcenter{\hbox{\scriptsize $\pm 1.6$}}$ \\
\midrule
\multicolumn{11}{c}{\textbf{Protein Graphs}} \\
\midrule
\textbf{Method} &
 &
 \textbf{Wavelet} $\downarrow$ &
 \textbf{Orbit} $\downarrow$ &
 \textbf{Clustering} $\downarrow$ &
 \textbf{Deg} $\downarrow$ &
 \textbf{Spectral} $\downarrow$ &
 \textbf{Ratio} $\downarrow$ \\
\midrule
    DiGress \cite{vignac2023digress} &
 &
 $0.006$ &
 $0.129$ &
 $0.049$ &
 $0.004$ &
 $0.002$ &
 $18.0$ \\
    HSpectre \cite{bergmeister2024efficient} &
 &
 $\underline{0.003}$ &
 $\mathbf{0.005}$ &
 $\mathbf{0.031}$ &
 $\underline{0.003}$ &
 $\mathbf{0.001}$ &
 $\underline{5.9}$ \\
    BwR \citep{diamant2023improving} &
 &
 $0.120$ &
 $0.494$ &
 $0.420$ &
 $0.126$ &
 $0.070$ &
 $245.4$ \\
    \cdashline{1-8}\\[-0.8em]
    \method &
 &
 $\mathbf{0.001 \vcenter{\hbox{\scriptsize $\boldsymbol{\pm 0.001}$}}}$ &
 $\underline{0.013 \vcenter{\hbox{\scriptsize $\pm 0.004$}}}$ &
 $\underline{0.039 \vcenter{\hbox{\scriptsize $\pm 0.005$}}}$ &
 $\mathbf{0.000 \vcenter{\hbox{\scriptsize $\boldsymbol{\pm 0.001}$}}}$ &
 $\mathbf{0.001 \vcenter{\hbox{\scriptsize $\boldsymbol{\pm 0.001}$}}}$ &
 $\mathbf{3.8 \vcenter{\hbox{\scriptsize $\boldsymbol{\pm 1.9}$}}}$ \\
\midrule
\multicolumn{11}{c}{\textbf{Point Cloud (Unfeatured) Graphs}} \\
\midrule
\textbf{Method} &
  &
 \textbf{Wavelet} $\downarrow$ &
 \textbf{Orbit} $\downarrow$ &
 \textbf{Clustering} $\downarrow$ &
 \textbf{Deg} $\downarrow$ &
 \textbf{Spectral} $\downarrow$ &
 \textbf{Ratio} $\downarrow$ \\
\midrule
    DiGress \cite{vignac2023digress} &
 &
 OOM &
 OOM &
 OOM &
 OOM &
 OOM &
 OOM \\
    HSpectre \cite{bergmeister2024efficient} &
 &
 $\underline{0.019}$ &
 $\underline{0.078}$ &
 $\underline{0.578}$ &
 $\underline{0.014}$ &
 $\mathbf{0.005}$ &
 $\underline{7.0}$ \\
    BwR \citep{diamant2023improving} &
 &
 $0.592$ &
 $1.073$ &
 $0.469$ &
 $0.493$ &
 $0.291$ &
 $133.2$ \\
    \cdashline{1-8}\\[-0.8em]
    \method &
 &
 $\mathbf{0.018 \vcenter{\hbox{\scriptsize $\boldsymbol{\pm 0.005}$}}}$ &
 $\mathbf{0.009 \vcenter{\hbox{\scriptsize $\boldsymbol{\pm 0.001}$}}}$ &
 $\mathbf{0.441 \vcenter{\hbox{\scriptsize $\boldsymbol{\pm 0.073}$}}}$ &
 $\mathbf{0.006 \vcenter{\hbox{\scriptsize $\boldsymbol{\pm 0.012}$}}}$ &
 $\mathbf{0.006 \vcenter{\hbox{\scriptsize $\boldsymbol{\pm 0.001}$}}}$ &
 $\mathbf{3.2 \vcenter{\hbox{\scriptsize $\boldsymbol{\pm 0.5}$}}}$ \\
\bottomrule
\end{tabular}
}
\end{table*}

\clearpage

\begin{table*}[h!]
\centering
\caption{Detailed numerical results for graph point cloud datasets.}
\setlength{\tabcolsep}{5pt}
\resizebox{\textwidth}{!}{
\begin{tabular}{@{}cccccccccccccc@{}}
\toprule
\multicolumn{12}{c}{\textbf{Airplane Point Clouds}} \\
\midrule
\textbf{Method} &
 \textbf{ChamDist} $\downarrow$ &
\textbf{ChamCov} (\%) $\uparrow$ &
\textbf{ChamDiv} $\uparrow$ &
 \textbf{NumDiff} $\downarrow$ &
 \textbf{Wavelet} $\downarrow$ &
 \textbf{Orbit} $\downarrow$ &
 \textbf{Clustering} $\downarrow$ &
 \textbf{Deg} $\downarrow$ &
 \textbf{Spectral} $\downarrow$ &
 \textbf{Ratio} $\downarrow$ \\
\midrule
    DiGress \cite{vignac2023digress} &
 OOM &
 OOM &
 OOM &
 OOM &
 OOM &
 OOM &
 OOM &
 OOM &
 OOM &
 OOM  \\
    DeFoG \cite{qin2025defogdiscreteflowmatching} &
 OOM &
 OOM &
 OOM &
 OOM &
 OOM &
 OOM &
 OOM &
 OOM &
 OOM &
 OOM  \\
    \cdashline{1-12}\\[-0.8em]
    \method &
 $\mathbf{0.094 \vcenter{\hbox{\scriptsize $\boldsymbol{\pm 0.006}$}}}$ &
 $\mathbf{22.2 \vcenter{\hbox{\scriptsize $\boldsymbol{\pm 14.5}$}}}$ &
 $\mathbf{0.085 \vcenter{\hbox{\scriptsize $\boldsymbol{\pm 0.026}$}}}$ &
 $\mathbf{0.000 \vcenter{\hbox{\scriptsize $\boldsymbol{\pm 0.000}$}}}$ &
 $\mathbf{0.004 \vcenter{\hbox{\scriptsize $\boldsymbol{\pm 0.001}$}}}$ &
 $\mathbf{0.074 \vcenter{\hbox{\scriptsize $\boldsymbol{\pm 0.104}$}}}$ &
 $\mathbf{0.264 \vcenter{\hbox{\scriptsize $\boldsymbol{\pm 0.254}$}}}$ &
 $\mathbf{0.002 \vcenter{\hbox{\scriptsize $\boldsymbol{\pm 0.002}$}}}$ &
 $\mathbf{0.005 \vcenter{\hbox{\scriptsize $\boldsymbol{\pm 0.004}$}}}$ &
 $\mathbf{67.3 \vcenter{\hbox{\scriptsize $\boldsymbol{\pm 44.9}$}}}$ \\
\midrule
\multicolumn{12}{c}{\textbf{Bench Point Clouds}} \\
\midrule
\textbf{Method} &
 \textbf{ChamDist} $\downarrow$ &
\textbf{ChamCov} (\%) $\uparrow$ &
\textbf{ChamDiv} $\uparrow$ &
 \textbf{NumDiff} $\downarrow$ &
 \textbf{Wavelet} $\downarrow$ &
 \textbf{Orbit} $\downarrow$ &
 \textbf{Clustering} $\downarrow$ &
 \textbf{Deg} $\downarrow$ &
 \textbf{Spectral} $\downarrow$ &
 \textbf{Ratio} $\downarrow$ \\
    \midrule
    DiGress \cite{vignac2023digress} &
 OOM &
 OOM &
 OOM &
 OOM &
 OOM &
 OOM &
 OOM &
 OOM &
 OOM &
 OOM  \\
    DeFoG \cite{qin2025defogdiscreteflowmatching} &
 OOM &
 OOM &
 OOM &
 OOM &
 OOM &
 OOM &
 OOM &
 OOM &
 OOM &
 OOM  \\
    \cdashline{1-12}\\[-0.8em]
    \method &
 $\mathbf{0.130 \vcenter{\hbox{\scriptsize $\boldsymbol{\pm 0.001}$}}}$ &
 $\mathbf{15.8 \vcenter{\hbox{\scriptsize $\boldsymbol{\pm 10.6}$}}}$ &
 $\mathbf{0.191 \vcenter{\hbox{\scriptsize $\boldsymbol{\pm 0.114}$}}}$ &
 $\mathbf{0.000 \vcenter{\hbox{\scriptsize $\boldsymbol{\pm 0.000}$}}}$ &
 $\mathbf{0.003 \vcenter{\hbox{\scriptsize $\boldsymbol{\pm 0.000}$}}}$ &
 $\mathbf{0.013 \vcenter{\hbox{\scriptsize $\boldsymbol{\pm 0.004}$}}}$ &
 $\mathbf{0.229 \vcenter{\hbox{\scriptsize $\boldsymbol{\pm 0.071}$}}}$ &
 $\mathbf{0.000 \vcenter{\hbox{\scriptsize $\boldsymbol{\pm 0.000}$}}}$ &
 $\mathbf{0.004 \vcenter{\hbox{\scriptsize $\boldsymbol{\pm 0.000}$}}}$ &
 $\mathbf{73.3 \vcenter{\hbox{\scriptsize $\boldsymbol{\pm 22.2}$}}}$ \\
\bottomrule
\end{tabular}
}
\end{table*}

\vspace{2em}

\subsection{Ablation Studies}

\begin{table*}[h!]
\centering
\caption{Detailed numerical results for ablation studies on unfeatured hypergraphs.}
\setlength{\tabcolsep}{5pt}
\resizebox{\textwidth}{!}{
\begin{tabular}{@{}cccccccccccc@{}}
\toprule
\multicolumn{12}{c}{\textbf{SBM Hypergraphs}} \\
\midrule
\textbf{node budget} &
 \textbf{Graph OT Coupling} &
 \textbf{Valid} (\%) $\uparrow$ &
 \textbf{NumDiff} $\downarrow$ &
 \textbf{Deg} $\downarrow$ &
 \textbf{EdgeSize} $\downarrow$ &
 \textbf{Spectral} $\downarrow$ &
 \textbf{Harmonic} $\downarrow$ &
 \textbf{Closeness} $\downarrow$ &
 \textbf{Betweenness} $\downarrow$ \\
\midrule
\cmark &
 \cmark &
 $\mathbf{87.8 \vcenter{\hbox{\scriptsize $\boldsymbol{\pm 3.1}$}}}$ &
 $\mathbf{0.029 \vcenter{\hbox{\scriptsize $\boldsymbol{\pm 0.009}$}}}$ &
 $\mathbf{0.846 \vcenter{\hbox{\scriptsize $\boldsymbol{\pm 0.457}$}}}$ &
 $\mathbf{0.005 \vcenter{\hbox{\scriptsize $\boldsymbol{\pm 0.003}$}}}$ &
 $\mathbf{0.006 \vcenter{\hbox{\scriptsize $\boldsymbol{\pm 0.004}$}}}$ &
 $\underline{6.410 \vcenter{\hbox{\scriptsize $\pm $3.124}}}$ &
 $\mathbf{0.009 \vcenter{\hbox{\scriptsize $\boldsymbol{\pm 0.006}$}}}$ &
 $\mathbf{0.003 \vcenter{\hbox{\scriptsize $\boldsymbol{\pm 0.001}$}}}$ \\
\xmark &
 \cmark &
 $85.3 \vcenter{\hbox{\scriptsize $\pm $5.9}}$ &
 $0.044 \vcenter{\hbox{\scriptsize $\pm $0.078}}$ &
 $\underline{0.856 \vcenter{\hbox{\scriptsize $\pm $0.636}}}$ &
 $0.023 \vcenter{\hbox{\scriptsize $\pm $0.022}}$ &
 $\mathbf{0.006 \vcenter{\hbox{\scriptsize $\boldsymbol{\pm 0.004}$}}}$ &
 $\mathbf{6.210 \vcenter{\hbox{\scriptsize $\boldsymbol{\pm 4.649}$}}}$ &
 $\mathbf{0.009 \vcenter{\hbox{\scriptsize $\boldsymbol{\pm 0.013}$}}}$ &
 $\mathbf{0.003 \vcenter{\hbox{\scriptsize $\boldsymbol{\pm 0.003}$}}}$ \\
\cmark &
 \xmark &
 $\underline{86.7 \vcenter{\hbox{\scriptsize $\pm $6.7}}}$ &
 $\underline{0.039 \vcenter{\hbox{\scriptsize $\pm $0.014}}}$ &
 $0.910 \vcenter{\hbox{\scriptsize $\pm $0.363}}$ &
 $\underline{0.006 \vcenter{\hbox{\scriptsize $\pm $0.004}}}$ &
 $\mathbf{0.006 \vcenter{\hbox{\scriptsize $\boldsymbol{\pm 0.004}$}}}$ &
 $6.815 \vcenter{\hbox{\scriptsize $\pm $1.913}}$ &
 $0.010 \vcenter{\hbox{\scriptsize $\pm $0.005}}$ &
 $0.004 \vcenter{\hbox{\scriptsize $\pm $0.001}}$ \\
\xmark &
 \xmark &
 $84.6 \vcenter{\hbox{\scriptsize $\pm $4.6}}$ &
 $0.049 \vcenter{\hbox{\scriptsize $\pm $0.045}}$ &
 $0.916 \vcenter{\hbox{\scriptsize $\pm $0.749}}$ &
 $0.047 \vcenter{\hbox{\scriptsize $\pm $0.070}}$ &
 $0.007 \vcenter{\hbox{\scriptsize $\pm $0.007}}$ &
 $6.665 \vcenter{\hbox{\scriptsize $\pm $5.199}}$ &
 $0.012 \vcenter{\hbox{\scriptsize $\pm $0.014}}$ &
 $0.004 \vcenter{\hbox{\scriptsize $\pm $0.004}}$ \\
\midrule
\multicolumn{12}{c}{\textbf{Ego Hypergraphs}} \\
\midrule
\textbf{node budget} &
 \textbf{Graph OT Coupling} &
 \textbf{Valid} (\%) $\uparrow$ &
 \textbf{NumDiff} $\downarrow$ &
 \textbf{Deg} $\downarrow$ &
 \textbf{EdgeSize} $\downarrow$ &
 \textbf{Spectral} $\downarrow$ &
 \textbf{Harmonic} $\downarrow$ &
 \textbf{Closeness} $\downarrow$ &
 \textbf{Betweenness} $\downarrow$ \\
\midrule
\cmark &
 \cmark &
 $99.5 \vcenter{\hbox{\scriptsize $\pm $1.1}}$ &
 $0.128 \vcenter{\hbox{\scriptsize $\pm $0.171}}$ &
 $\underline{0.124 \vcenter{\hbox{\scriptsize $\pm $0.086}}}$ &
 $\underline{0.155 \vcenter{\hbox{\scriptsize $\pm $0.067}}}$ &
 $\mathbf{0.004 \vcenter{\hbox{\scriptsize $\boldsymbol{\pm 0.003}$}}}$ &
 $\mathbf{2.703 \vcenter{\hbox{\scriptsize $\boldsymbol{\pm 3.468}$}}}$ &
 $0.003 \vcenter{\hbox{\scriptsize $\pm $0.005}}$ &
 $\mathbf{0.000 \vcenter{\hbox{\scriptsize $\boldsymbol{\pm 0.000}$}}}$ \\
\xmark &
 \cmark &
 $\mathbf{100.0 \vcenter{\hbox{\scriptsize $\boldsymbol{\pm 0.0}$}}}$ &
 $0.118 \vcenter{\hbox{\scriptsize $\pm $0.158}}$ &
 $0.140 \vcenter{\hbox{\scriptsize $\pm $0.104}}$ &
 $\mathbf{0.133 \vcenter{\hbox{\scriptsize $\boldsymbol{\pm 0.106}$}}}$ &
 $0.005 \vcenter{\hbox{\scriptsize $\pm $0.003}}$ &
 $3.426 \vcenter{\hbox{\scriptsize $\pm $2.629}}$ &
 $\mathbf{0.000 \vcenter{\hbox{\scriptsize $\boldsymbol{\pm 0.000}$}}}$ &
 $\mathbf{0.000 \vcenter{\hbox{\scriptsize $\boldsymbol{\pm 0.000}$}}}$ \\
\cmark &
 \xmark &
 $\underline{99.9 \vcenter{\hbox{\scriptsize $\pm $0.4}}}$ &
 $\mathbf{0.073 \vcenter{\hbox{\scriptsize $\boldsymbol{\pm 0.050}$}}}$ &
 $0.237 \vcenter{\hbox{\scriptsize $\pm $0.465}}$ &
 $0.193 \vcenter{\hbox{\scriptsize $\pm $0.175}}$ &
 $0.007 \vcenter{\hbox{\scriptsize $\pm $0.012}}$ &
 $4.521 \vcenter{\hbox{\scriptsize $\pm $5.629}}$ &
 $\mathbf{0.000 \vcenter{\hbox{\scriptsize $\boldsymbol{\pm 0.002}$}}}$ &
 $\mathbf{0.000 \vcenter{\hbox{\scriptsize $\boldsymbol{\pm 0.000}$}}}$ \\
\xmark &
 \xmark &
 $99.5 \vcenter{\hbox{\scriptsize $\pm $1.1}}$ &
 $\underline{0.117 \vcenter{\hbox{\scriptsize $\pm $0.155}}}$ &
 $\mathbf{0.110 \vcenter{\hbox{\scriptsize $\boldsymbol{\pm 0.111}$}}}$ &
 $0.189 \vcenter{\hbox{\scriptsize $\pm $0.084}}$ &
 $\mathbf{0.004 \vcenter{\hbox{\scriptsize $\boldsymbol{\pm 0.003}$}}}$ &
 $\underline{2.765 \vcenter{\hbox{\scriptsize $\pm $1.684}}}$ &
 $0.001 \vcenter{\hbox{\scriptsize $\pm $0.004}}$ &
 $\mathbf{0.000 \vcenter{\hbox{\scriptsize $\boldsymbol{\pm 0.000}$}}}$ \\
\midrule
\multicolumn{12}{c}{\textbf{Tree Hypergraphs}} \\
\midrule
\textbf{node budget} &
 \textbf{Graph OT Coupling} &
 \textbf{Valid} (\%) $\uparrow$ &
 \textbf{NumDiff} $\downarrow$ &
 \textbf{Deg} $\downarrow$ &
 \textbf{EdgeSize} $\downarrow$ &
 \textbf{Spectral} $\downarrow$ &
 \textbf{Harmonic} $\downarrow$ &
 \textbf{Closeness} $\downarrow$ &
 \textbf{Betweenness} $\downarrow$ \\
\midrule
\cmark &
 \cmark &
 $89.7 \vcenter{\hbox{\scriptsize $\pm $6.0}}$ &
 $\mathbf{0.000 \vcenter{\hbox{\scriptsize $\boldsymbol{\pm 0.000}$}}}$ &
 $\underline{0.022 \vcenter{\hbox{\scriptsize $\pm $0.022}}}$ &
 $\mathbf{0.030 \vcenter{\hbox{\scriptsize $\boldsymbol{\pm 0.034}$}}}$ &
 $\mathbf{0.003 \vcenter{\hbox{\scriptsize $\boldsymbol{\pm 0.002}$}}}$ &
 $\underline{0.171 \vcenter{\hbox{\scriptsize $\pm $0.106}}}$ &
 $\underline{0.014 \vcenter{\hbox{\scriptsize $\pm $0.006}}}$ &
 $0.014 \vcenter{\hbox{\scriptsize $\pm $0.004}}$ \\
\xmark &
 \cmark &
 $\underline{90.7 \vcenter{\hbox{\scriptsize $\pm $6.7}}}$ &
 $\mathbf{0.000 \vcenter{\hbox{\scriptsize $\boldsymbol{\pm 0.000}$}}}$ &
 $0.024 \vcenter{\hbox{\scriptsize $\pm $0.019}}$ &
 $0.067 \vcenter{\hbox{\scriptsize $\pm $0.016}}$ &
 $\underline{0.004 \vcenter{\hbox{\scriptsize $\pm $0.001}}}$ &
 $0.315 \vcenter{\hbox{\scriptsize $\pm $0.098}}$ &
 $0.018 \vcenter{\hbox{\scriptsize $\pm $0.012}}$ &
 $0.014 \vcenter{\hbox{\scriptsize $\pm $0.008}}$ \\
\cmark &
 \xmark &
 $89.3 \vcenter{\hbox{\scriptsize $\pm $3.6}}$ &
 $\mathbf{0.000 \vcenter{\hbox{\scriptsize $\boldsymbol{\pm 0.000}$}}}$ &
 $\mathbf{0.019 \vcenter{\hbox{\scriptsize $\boldsymbol{\pm 0.013}$}}}$ &
 $\underline{0.047 \vcenter{\hbox{\scriptsize $\pm $0.077}}}$ &
 $0.005 \vcenter{\hbox{\scriptsize $\pm $0.011}}$ &
 $\mathbf{0.169 \vcenter{\hbox{\scriptsize $\boldsymbol{\pm 0.106}$}}}$ &
 $0.041 \vcenter{\hbox{\scriptsize $\pm $0.082}}$ &
 $\underline{0.013 \vcenter{\hbox{\scriptsize $\pm $0.010}}}$ \\
\xmark &
 \xmark &
 $\mathbf{95.6 \vcenter{\hbox{\scriptsize $\boldsymbol{\pm 3.7}$}}}$ &
 $\mathbf{0.000 \vcenter{\hbox{\scriptsize $\boldsymbol{\pm 0.000}$}}}$ &
 $0.042 \vcenter{\hbox{\scriptsize $\pm $0.042}}$ &
 $0.077 \vcenter{\hbox{\scriptsize $\pm $0.091}}$ &
 $0.005 \vcenter{\hbox{\scriptsize $\pm $0.003}}$ &
 $0.217 \vcenter{\hbox{\scriptsize $\pm $0.169}}$ &
 $\mathbf{0.013 \vcenter{\hbox{\scriptsize $\boldsymbol{\pm 0.013}$}}}$ &
 $\mathbf{0.008 \vcenter{\hbox{\scriptsize $\boldsymbol{\pm 0.011}$}}}$ \\
\midrule
\multicolumn{12}{c}{\textbf{ModelNet Bookshelf}} \\
\midrule
\textbf{node budget} &
 \textbf{Graph OT Coupling} &
 &
 \textbf{NumDiff} $\downarrow$ &
 \textbf{Deg} $\downarrow$ &
 \textbf{EdgeSize} $\downarrow$ &
 \textbf{Spectral} $\downarrow$ &
 \textbf{Harmonic} $\downarrow$ &
 \textbf{Closeness} $\downarrow$ &
 \textbf{Betweenness} $\downarrow$ \\
\midrule
\cmark &
 \cmark &
&
 $\mathbf{0.135 \vcenter{\hbox{\scriptsize $\boldsymbol{\pm 0.276}$}}}$ &
 $\underline{2.980 \vcenter{\hbox{\scriptsize $\pm $2.107}}}$ &
 $\mathbf{0.020 \vcenter{\hbox{\scriptsize $\boldsymbol{\pm 0.025}$}}}$ &
 $\underline{0.024 \vcenter{\hbox{\scriptsize $\pm $0.015}}}$ &
 $\mathbf{46.614 \vcenter{\hbox{\scriptsize $\boldsymbol{\pm 58.803}$}}}$ &
 $\mathbf{0.086 \vcenter{\hbox{\scriptsize $\boldsymbol{\pm 0.030}$}}}$ &
 $\mathbf{0.001 \vcenter{\hbox{\scriptsize $\boldsymbol{\pm 0.001}$}}}$ \\
\xmark &
 \cmark &
&
 $0.940 \vcenter{\hbox{\scriptsize $\pm $0.917}}$ &
 $3.040 \vcenter{\hbox{\scriptsize $\pm $1.446}}$ &
 $0.089 \vcenter{\hbox{\scriptsize $\pm $0.099}}$ &
 $0.032 \vcenter{\hbox{\scriptsize $\pm $0.013}}$ &
 $59.300 \vcenter{\hbox{\scriptsize $\pm $108.101}}$ &
 $0.137 \vcenter{\hbox{\scriptsize $\pm $0.042}}$ &
 $0.003 \vcenter{\hbox{\scriptsize $\pm $0.001}}$ \\
\cmark &
 \xmark &
&
 $\underline{0.265 \vcenter{\hbox{\scriptsize $\pm $0.496}}}$ &
 $4.400 \vcenter{\hbox{\scriptsize $\pm $1.635}}$ &
 $\underline{0.025 \vcenter{\hbox{\scriptsize $\pm $0.021}}}$ &
 $\mathbf{0.014 \vcenter{\hbox{\scriptsize $\boldsymbol{\pm 0.007}$}}}$ &
 $\underline{54.558 \vcenter{\hbox{\scriptsize $\pm $30.276}}}$ &
 $\underline{0.102 \vcenter{\hbox{\scriptsize $\pm $0.020}}}$ &
 $\mathbf{0.001 \vcenter{\hbox{\scriptsize $\boldsymbol{\pm 0.001}$}}}$ \\
\xmark &
 \xmark &
&
 $1.325 \vcenter{\hbox{\scriptsize $\pm $1.631}}$ &
 $\mathbf{2.820 \vcenter{\hbox{\scriptsize $\boldsymbol{\pm 0.958}$}}}$ &
 $0.055 \vcenter{\hbox{\scriptsize $\pm $0.077}}$ &
 $0.031 \vcenter{\hbox{\scriptsize $\pm $0.009}}$ &
 $79.889 \vcenter{\hbox{\scriptsize $\pm $98.721}}$ &
 $0.135 \vcenter{\hbox{\scriptsize $\pm $0.012}}$ &
 $0.003 \vcenter{\hbox{\scriptsize $\pm $0.001}}$ \\
\midrule
\multicolumn{12}{c}{\textbf{ModelNet Piano}} \\
\midrule
\textbf{node budget} &
 \textbf{Graph OT Coupling} &
 &
 \textbf{NumDiff} $\downarrow$ &
 \textbf{Deg} $\downarrow$ &
 \textbf{EdgeSize} $\downarrow$ &
 \textbf{Spectral} $\downarrow$ &
 \textbf{Harmonic} $\downarrow$ &
 \textbf{Closeness} $\downarrow$ &
 \textbf{Betweenness} $\downarrow$ \\
\midrule
\cmark &
 \cmark &
&
 $\mathbf{0.846 \vcenter{\hbox{\scriptsize $\boldsymbol{\pm 1.009}$}}}$ &
 $\mathbf{3.265 \vcenter{\hbox{\scriptsize $\boldsymbol{\pm 1.954}$}}}$ &
 $\mathbf{0.042 \vcenter{\hbox{\scriptsize $\boldsymbol{\pm 0.056}$}}}$ &
 $0.040 \vcenter{\hbox{\scriptsize $\pm $0.026}}$ &
 $\underline{119.158 \vcenter{\hbox{\scriptsize $\pm $81.023}}}$ &
 $\mathbf{0.123 \vcenter{\hbox{\scriptsize $\boldsymbol{\pm 0.119}$}}}$ &
 $\mathbf{0.002 \vcenter{\hbox{\scriptsize $\boldsymbol{\pm 0.002}$}}}$ \\
\xmark &
 \cmark &
&
 $3.622 \vcenter{\hbox{\scriptsize $\pm $1.822}}$ &
 $\underline{3.358 \vcenter{\hbox{\scriptsize $\pm $1.772}}}$ &
 $\underline{0.054 \vcenter{\hbox{\scriptsize $\pm $0.080}}}$ &
 $0.055 \vcenter{\hbox{\scriptsize $\pm $0.040}}$ &
 $133.806 \vcenter{\hbox{\scriptsize $\pm $137.603}}$ &
 $0.155 \vcenter{\hbox{\scriptsize $\pm $0.058}}$ &
 $0.012 \vcenter{\hbox{\scriptsize $\pm $0.033}}$ \\
\cmark &
 \xmark &
&
 $\underline{3.155 \vcenter{\hbox{\scriptsize $\pm $3.637}}}$ &
 $5.250 \vcenter{\hbox{\scriptsize $\pm $1.087}}$ &
 $0.066 \vcenter{\hbox{\scriptsize $\pm $0.168}}$ &
 $\mathbf{0.030 \vcenter{\hbox{\scriptsize $\boldsymbol{\pm 0.048}$}}}$ &
 $154.211 \vcenter{\hbox{\scriptsize $\pm $383.979}}$ &
 $0.188 \vcenter{\hbox{\scriptsize $\pm $0.164}}$ &
 $\mathbf{0.002 \vcenter{\hbox{\scriptsize $\boldsymbol{\pm 0.002}$}}}$ \\
\xmark &
 \xmark &
&
 $5.490 \vcenter{\hbox{\scriptsize $\pm $8.847}}$ &
 $3.733 \vcenter{\hbox{\scriptsize $\pm $2.989}}$ &
 $0.107 \vcenter{\hbox{\scriptsize $\pm $0.215}}$ &
 $\underline{0.036 \vcenter{\hbox{\scriptsize $\pm $0.016}}}$ &
 $\mathbf{97.724 \vcenter{\hbox{\scriptsize $\boldsymbol{\pm 198.774}$}}}$ &
 $\underline{0.138 \vcenter{\hbox{\scriptsize $\pm $0.129}}}$ &
 $\mathbf{0.002 \vcenter{\hbox{\scriptsize $\boldsymbol{\pm 0.002}$}}}$ \\
\bottomrule
\end{tabular}
}
\end{table*}

\clearpage

\begin{table*}[h!]
\centering
\caption{Detailed numerical results for ablation studies on 3D meshes.}
\setlength{\tabcolsep}{5pt}
\resizebox{\textwidth}{!}{
\begin{tabular}{@{}cccccccccccccc@{}}
\toprule
\multicolumn{12}{c}{\textbf{Manifold40 Airplane}} \\
\midrule
\textbf{node budget} &
 \textbf{Graph OT Coupling} &
 \textbf{ChamDist} $\downarrow$ &
 \textbf{ChamCov} (\%) $\uparrow$ &
 \textbf{ChamDiv} $\uparrow$ &
 \textbf{NumDiff} $\downarrow$ &
 \textbf{Deg} $\downarrow$ &
 \textbf{EdgeSize} $\downarrow$ &
 \textbf{Spectral} $\downarrow$ &
 \textbf{Harmonic} $\downarrow$ &
 \textbf{Closeness} $\downarrow$ &
 \textbf{Betweenness} $\downarrow$ \\
\midrule
\cmark &
 \cmark &
 $\mathbf{0.048 \vcenter{\hbox{\scriptsize $\boldsymbol{\pm 0.003}$}}}$ &
 $\underline{31.9 \vcenter{\hbox{\scriptsize $\pm 13.2$}}}$ &
 $\underline{0.101 \vcenter{\hbox{\scriptsize $\pm 0.011$}}}$ &
 $\mathbf{0.017 \vcenter{\hbox{\scriptsize $\boldsymbol{\pm 0.070}$}}}$ &
 $\mathbf{0.218 \vcenter{\hbox{\scriptsize $\boldsymbol{\pm 0.085}$}}}$ &
 $\mathbf{0.004 \vcenter{\hbox{\scriptsize $\boldsymbol{\pm 0.007}$}}}$ &
 $\mathbf{0.010 \vcenter{\hbox{\scriptsize $\boldsymbol{\pm 0.003}$}}}$ &
 $\underline{2.428 \vcenter{\hbox{\scriptsize $\pm $1.455}}}$ &
 $0.032 \vcenter{\hbox{\scriptsize $\pm $0.007}}$ &
 $\mathbf{0.003 \vcenter{\hbox{\scriptsize $\boldsymbol{\pm 0.001}$}}}$ \\
\xmark &
 \cmark &
 $0.079 \vcenter{\hbox{\scriptsize $\pm $0.019}}$ &
 $29.0 \vcenter{\hbox{\scriptsize $\pm 17.6$}}$ &
 $0.080 \vcenter{\hbox{\scriptsize $\pm 0.011$}}$ &
 $0.426 \vcenter{\hbox{\scriptsize $\pm $0.820}}$ &
 $0.588 \vcenter{\hbox{\scriptsize $\pm $0.451}}$ &
 $0.024 \vcenter{\hbox{\scriptsize $\pm $0.021}}$ &
 $0.014 \vcenter{\hbox{\scriptsize $\pm $0.007}}$ &
 $2.429 \vcenter{\hbox{\scriptsize $\pm $1.394}}$ &
 $0.033 \vcenter{\hbox{\scriptsize $\pm $0.028}}$ &
 $\mathbf{0.003 \vcenter{\hbox{\scriptsize $\boldsymbol{\pm 0.002}$}}}$ \\
\cmark &
 \xmark &
 $\underline{0.050 \vcenter{\hbox{\scriptsize $\pm $0.005}}}$ &
 $30.4 \vcenter{\hbox{\scriptsize $\pm 15.0$}}$ &
 $0.093 \vcenter{\hbox{\scriptsize $\pm 0.019$}}$ & 
 $\underline{0.052 \vcenter{\hbox{\scriptsize $\pm $0.065}}}$ &
 $\underline{0.235 \vcenter{\hbox{\scriptsize $\pm $0.063}}}$ &
 $\underline{0.014 \vcenter{\hbox{\scriptsize $\pm $0.027}}}$ &
 $\underline{0.012 \vcenter{\hbox{\scriptsize $\pm $0.005}}}$ &
 $\mathbf{1.604 \vcenter{\hbox{\scriptsize $\boldsymbol{\pm 1.400}$}}}$ &
 $\mathbf{0.022 \vcenter{\hbox{\scriptsize $\boldsymbol{\pm 0.007}$}}}$ &
 $\mathbf{0.003 \vcenter{\hbox{\scriptsize $\boldsymbol{\pm 0.001}$}}}$ \\
\xmark &
 \xmark &
 $0.100 \vcenter{\hbox{\scriptsize $\pm $0.023}}$ &
 $\mathbf{44.9 \vcenter{\hbox{\scriptsize $\boldsymbol{\pm 10.0}$}}}$ &
 $\mathbf{0.114 \vcenter{\hbox{\scriptsize $\boldsymbol{\pm 0.031}$}}}$ &
 $0.304 \vcenter{\hbox{\scriptsize $\pm $0.437}}$ &
 $0.864 \vcenter{\hbox{\scriptsize $\pm $0.358}}$ &
 $0.020 \vcenter{\hbox{\scriptsize $\pm $0.016}}$ &
 $0.019 \vcenter{\hbox{\scriptsize $\pm $0.009}}$ &
 $4.184 \vcenter{\hbox{\scriptsize $\pm $2.176}}$ &
 $\underline{0.025 \vcenter{\hbox{\scriptsize $\pm $0.009}}}$ &
 $0.005 \vcenter{\hbox{\scriptsize $\pm $0.002}}$ \\
\midrule
\multicolumn{12}{c}{\textbf{Manifold40 Bench}} \\
\midrule
\textbf{node budget} &
 \textbf{Graph OT Coupling} &
 \textbf{ChamDist} $\downarrow$ &
 \textbf{ChamCov} (\%) $\uparrow$ &
 \textbf{ChamDiv} $\uparrow$ &
 \textbf{NumDiff} $\downarrow$ &
 \textbf{Deg} $\downarrow$ &
 \textbf{EdgeSize} $\downarrow$ &
 \textbf{Spectral} $\downarrow$ &
 \textbf{Harmonic} $\downarrow$ &
 \textbf{Closeness} $\downarrow$ &
 \textbf{Betweenness} $\downarrow$ \\
\midrule
\cmark &
 \cmark &
 $\mathbf{0.064 \vcenter{\hbox{\scriptsize $\boldsymbol{\pm 0.005}$}}}$ &
 $\mathbf{42.2 \vcenter{\hbox{\scriptsize $\boldsymbol{\pm 13.9}$}}}$ &
 $\mathbf{0.243 \vcenter{\hbox{\scriptsize $\boldsymbol{\pm 0.037}$}}}$ &
 $\underline{0.060 \vcenter{\hbox{\scriptsize $\pm $0.149}}}$ &
 $\underline{0.349 \vcenter{\hbox{\scriptsize $\pm $0.454}}}$ &
 $\underline{0.009 \vcenter{\hbox{\scriptsize $\pm $0.017}}}$ &
 $\underline{0.017 \vcenter{\hbox{\scriptsize $\pm $0.011}}}$ &
 $\underline{5.069 \vcenter{\hbox{\scriptsize $\pm $8.171}}}$ &
 $0.046 \vcenter{\hbox{\scriptsize $\pm $0.040}}$ &
 $\mathbf{0.004 \vcenter{\hbox{\scriptsize $\boldsymbol{\pm 0.003}$}}}$ \\
\xmark &
 \cmark &
 $0.090 \vcenter{\hbox{\scriptsize $\pm $0.003}}$ &
 $37.8 \vcenter{\hbox{\scriptsize $\pm 16.8$}}$ &
 $0.232 \vcenter{\hbox{\scriptsize $\pm 0.048$}}$ &
 $0.240 \vcenter{\hbox{\scriptsize $\pm $0.265}}$ &
 $1.089 \vcenter{\hbox{\scriptsize $\pm $0.346}}$ &
 $0.013 \vcenter{\hbox{\scriptsize $\pm $0.015}}$ &
 $0.018 \vcenter{\hbox{\scriptsize $\pm $0.009}}$ &
 $6.507 \vcenter{\hbox{\scriptsize $\pm $2.771}}$ &
 $\mathbf{0.014 \vcenter{\hbox{\scriptsize $\boldsymbol{\pm 0.005}$}}}$ &
 $\underline{0.005 \vcenter{\hbox{\scriptsize $\pm $0.002}}}$ \\
\cmark &
 \xmark &
 $\underline{0.085 \vcenter{\hbox{\scriptsize $\pm $0.056}}}$ &
 $32.2 \vcenter{\hbox{\scriptsize $\pm 7.7$}}$ &
 $0.208 \vcenter{\hbox{\scriptsize $\pm 0.030$}}$ &
 $\mathbf{0.020 \vcenter{\hbox{\scriptsize $\boldsymbol{\pm 0.032}$}}}$ &
 $\mathbf{0.221 \vcenter{\hbox{\scriptsize $\boldsymbol{\pm 0.160}$}}}$ &
 $\mathbf{0.006 \vcenter{\hbox{\scriptsize $\boldsymbol{\pm 0.004}$}}}$ &
 $\mathbf{0.013 \vcenter{\hbox{\scriptsize $\boldsymbol{\pm 0.003}$}}}$ &
 $\mathbf{1.864 \vcenter{\hbox{\scriptsize $\boldsymbol{\pm 2.828}$}}}$ &
 $0.028 \vcenter{\hbox{\scriptsize $\pm $0.019}}$ &
 $\underline{0.005 \vcenter{\hbox{\scriptsize $\pm $0.006}}}$ \\
\xmark &
 \xmark &
 $0.098 \vcenter{\hbox{\scriptsize $\pm $0.024}}$ &
 $\mathbf{42.2 \vcenter{\hbox{\scriptsize $\boldsymbol{\pm 22.0}$}}}$ & 
 $\underline{0.235 \vcenter{\hbox{\scriptsize $\pm 0.053$}}}$ &
 $0.267 \vcenter{\hbox{\scriptsize $\pm $0.319}}$ &
 $1.242 \vcenter{\hbox{\scriptsize $\pm $0.582}}$ &
 $0.013 \vcenter{\hbox{\scriptsize $\pm $0.011}}$ &
 $0.022 \vcenter{\hbox{\scriptsize $\pm $0.014}}$ &
 $7.619 \vcenter{\hbox{\scriptsize $\pm $4.165}}$ &
 $\underline{0.015 \vcenter{\hbox{\scriptsize $\pm $0.016}}}$ &
 $0.006 \vcenter{\hbox{\scriptsize $\pm $0.004}}$ \\
\bottomrule
\end{tabular}
}
\end{table*}

We additionally provide the following results, where we try different strategies for the coarsening process:
\begin{enumerate}[leftmargin=0.5cm]
    \setlength\itemsep{0pt}
    \item \textit{Random} corresponds to random mergings of nodes, where the local variation cost in Algorithm \ref{alg:hypergraph_coarsening} is replaced by a random value.
    \item \textit{Feature} corresponds to mergings that prioritize pairs of adjacent nodes with similar features, where the local variation cost in Algorithm \ref{alg:hypergraph_coarsening} is replaced by the squared Euclidean distance between features.
\end{enumerate}
Results are shown in Table \ref{tab:coarsening_strategies}.

\begin{table*}[h!]
\centering
\caption{Ablations on the coarsening strategy employed.}
\label{tab:coarsening_strategies}
\resizebox{0.65\textwidth}{!}{
\begin{tabular}{lccccc}
\toprule
\multicolumn{6}{c}{\textbf{Manifold40 Bench}} \\
\midrule
\textbf{Method} & \textbf{ChamDist} $\downarrow$ & \textbf{NumDiff} $\downarrow$ & \textbf{Deg} $\downarrow$ & \textbf{EdgeSize} $\downarrow$ & \textbf{Spectral} $\downarrow$ \\
\midrule
Random Mergings     & 0.158          & 2.300          & \textbf{0.324} & \textbf{0.002} & \underline{0.019}          \\
Feature-Preserving  & \underline{0.085} & \underline{0.200} & 0.750          & \underline{0.004} & 0.030 \\
Spectrum-Preserving & \textbf{0.064} & \textbf{0.060} & \underline{0.349} & 0.009          & \textbf{0.017} \\
\midrule
\multicolumn{6}{c}{\textbf{Manifold40 Airplane}} \\
\midrule
\textbf{Method} & \textbf{ChamDist} $\downarrow$ & \textbf{NumDiff} $\downarrow$ & \textbf{Deg} $\downarrow$ & \textbf{EdgeSize} $\downarrow$ & \textbf{Spectral} $\downarrow$ \\
\midrule
Random Mergings     & \underline{0.054}          & \textbf{0.000} & 0.423          & 0.009 & \underline{0.011} \\
Feature-Preserving  & 0.056 & \textbf{0.000} & \underline{0.241} & \textbf{0.002} & 0.014          \\
Spectrum-Preserving & \textbf{0.048} & \underline{0.017} & \textbf{0.218} & \underline{0.004}          & \textbf{0.010} \\
\bottomrule
\end{tabular}
}
\end{table*}

\clearpage

\section{Comparison between Training and Generated Samples} \label{app:visual_examples}

\begin{figure}[h!]
    \centering
    \begin{subfigure}[b]{0.47\textwidth}
        \centering
        \begin{subfigure}[b]{0.47\textwidth}
            \centering
            \includegraphics[width=\textwidth]{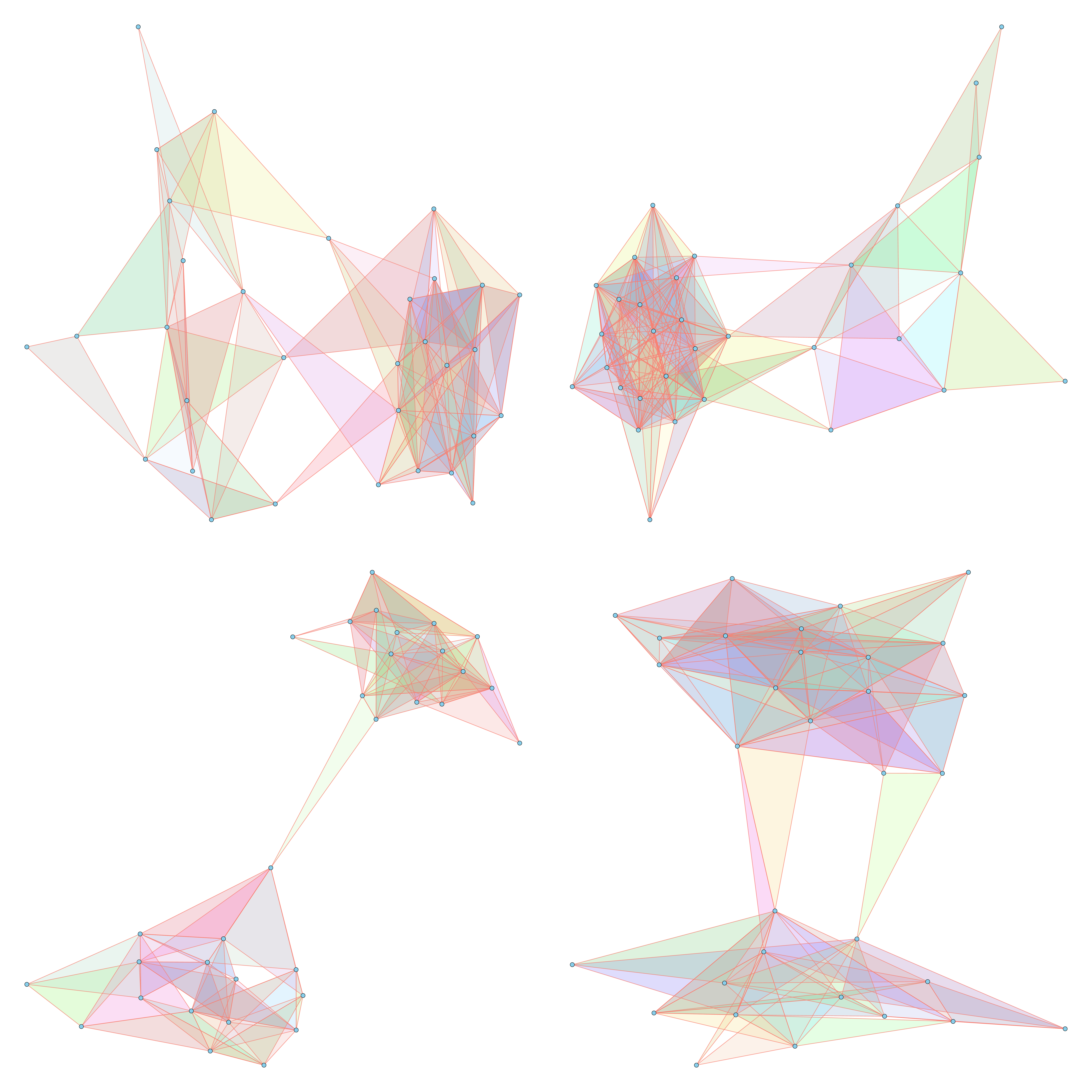}
            \caption*{Train samples}
        \end{subfigure}
        \hfill
        \begin{subfigure}[b]{0.47\textwidth}
            \centering
            \includegraphics[width=\textwidth]{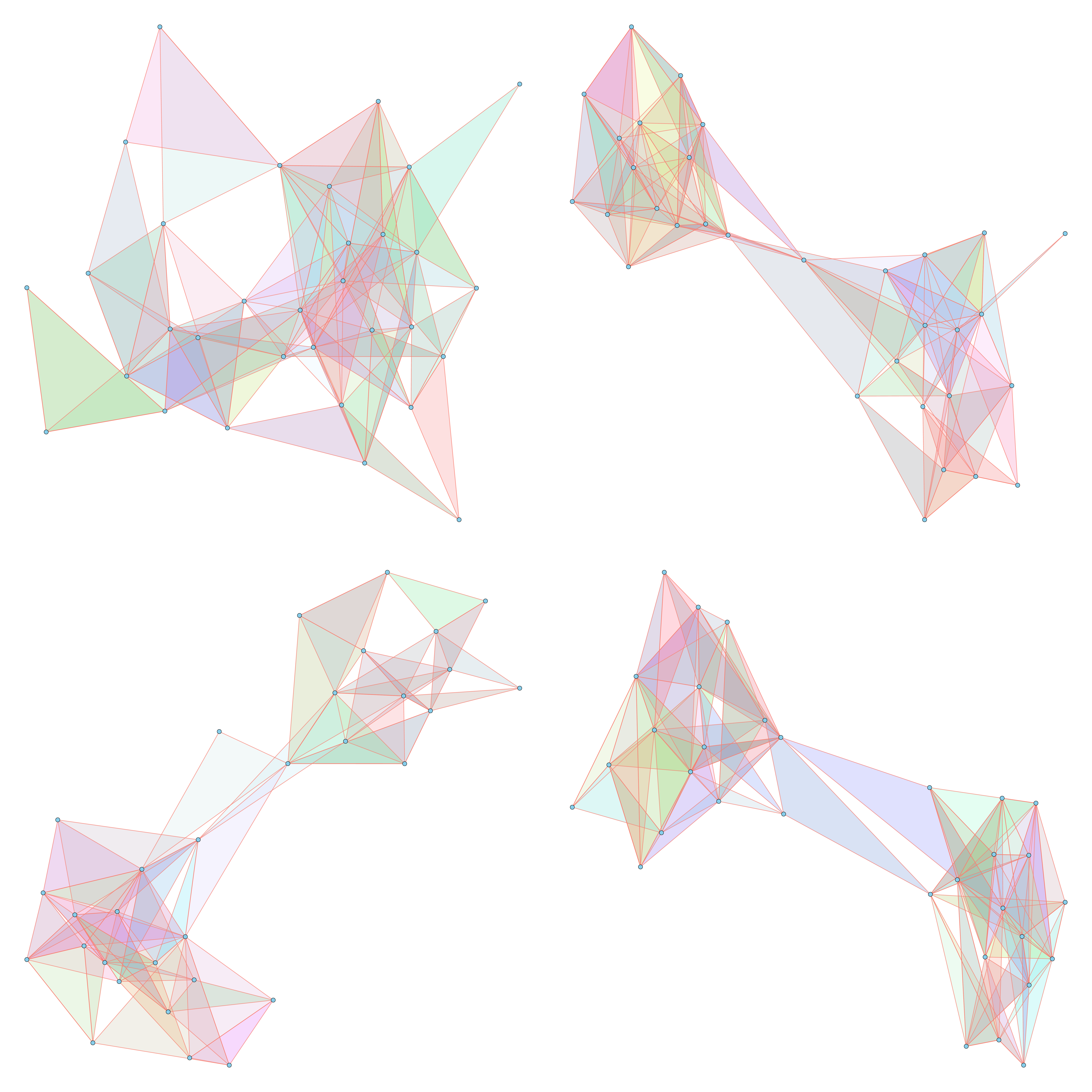}
            \caption*{Generated samples}
        \end{subfigure}
        \vspace{-0.15cm}
        \caption*{(\textit{i}) Stochastic Block Model hypergraphs.}
    \end{subfigure}
    \hspace{10pt}
    \begin{subfigure}[b]{0.47\textwidth}
        \centering
        \begin{subfigure}[b]{0.47\textwidth}
            \centering
            \includegraphics[width=\textwidth]{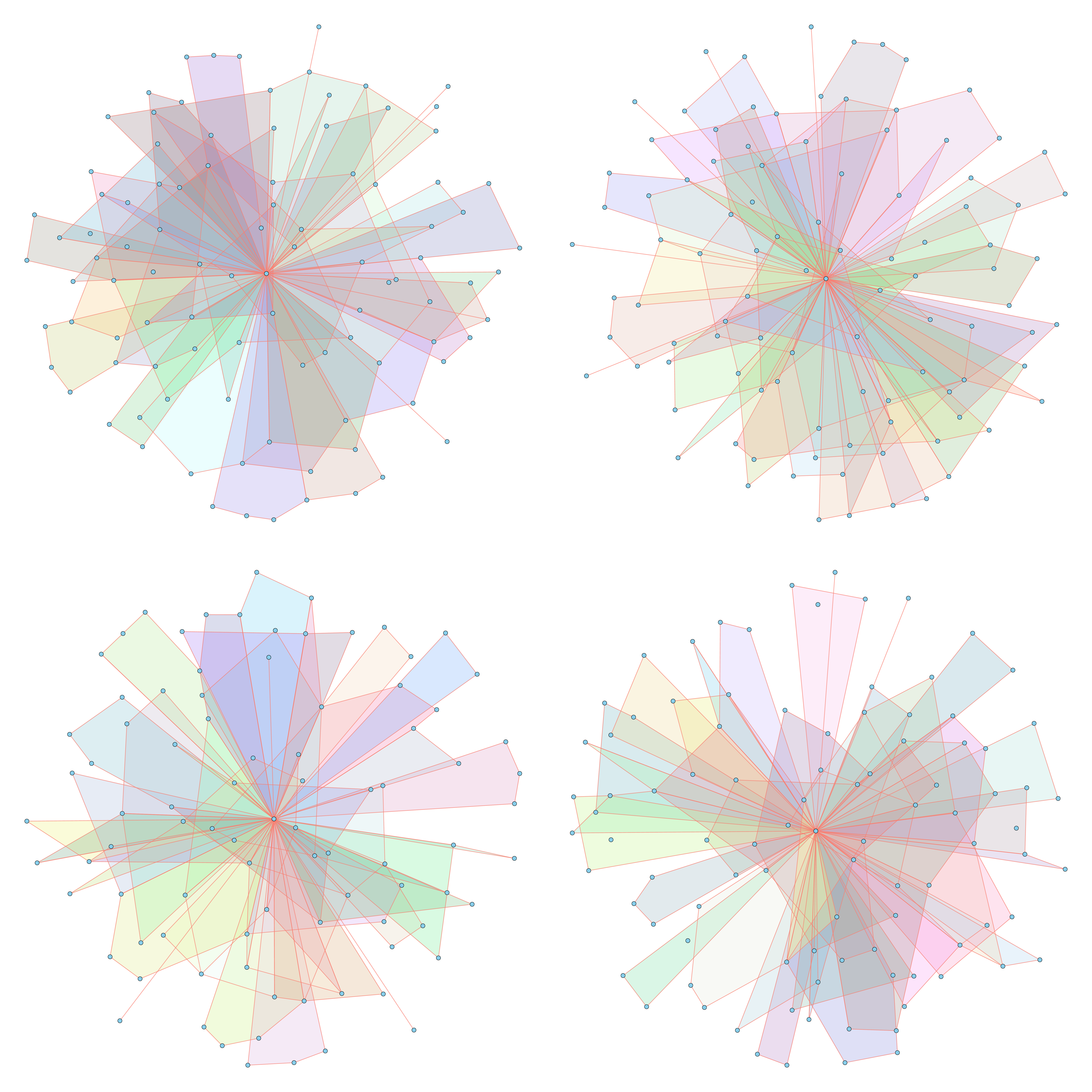}
            \caption*{Train samples}
        \end{subfigure}
        \hfill
        \begin{subfigure}[b]{0.47\textwidth}
            \centering
            \includegraphics[width=\textwidth]{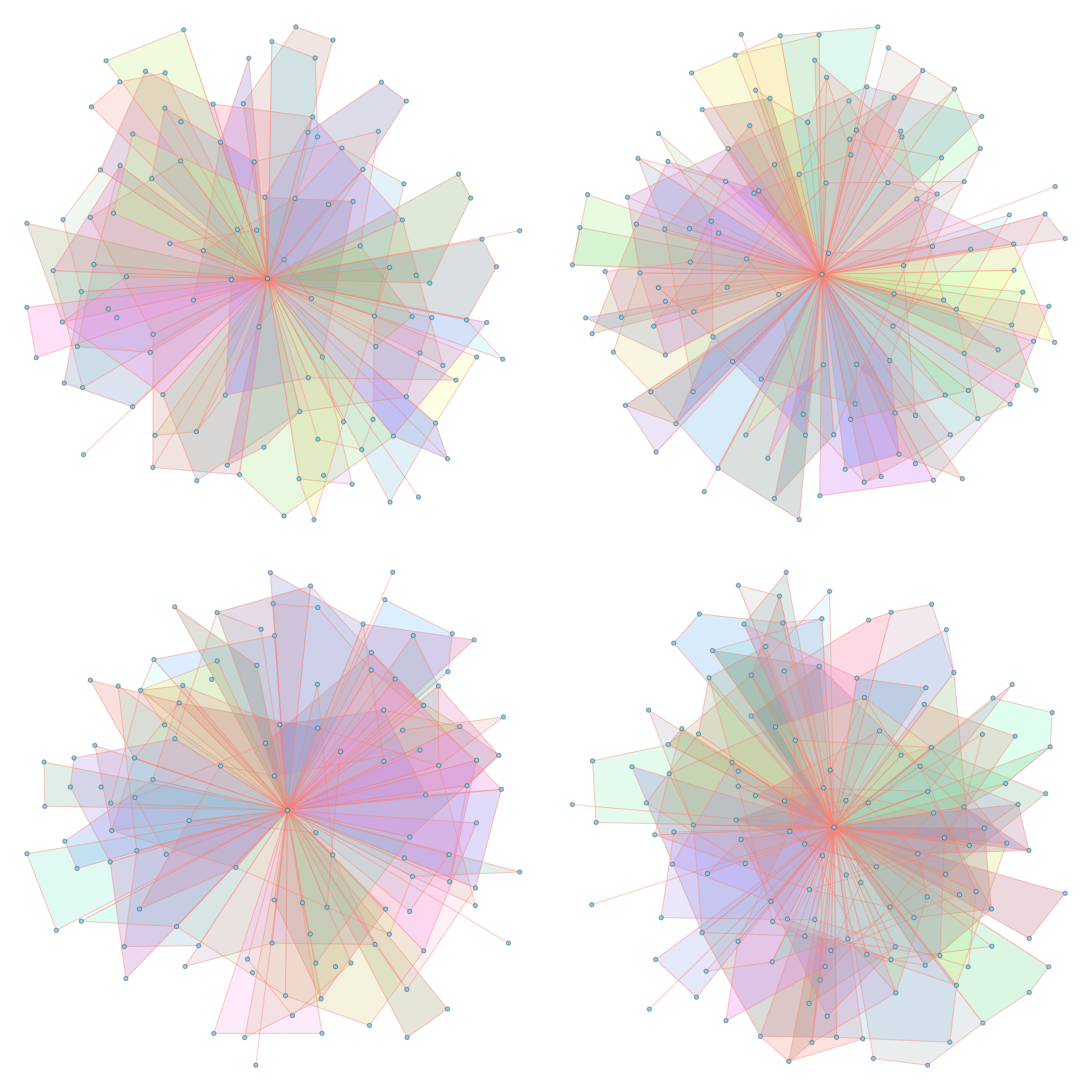}
            \caption*{Generated samples}
        \end{subfigure}
        \vspace{-0.15cm}
        \caption*{(\textit{ii}) Ego hypergraphs.}
    \end{subfigure}

     \vspace{0.21cm}

    \begin{subfigure}[b]{0.47\textwidth}
        \centering
        \begin{subfigure}[b]{0.47\textwidth}
            \centering
            \includegraphics[width=\textwidth]{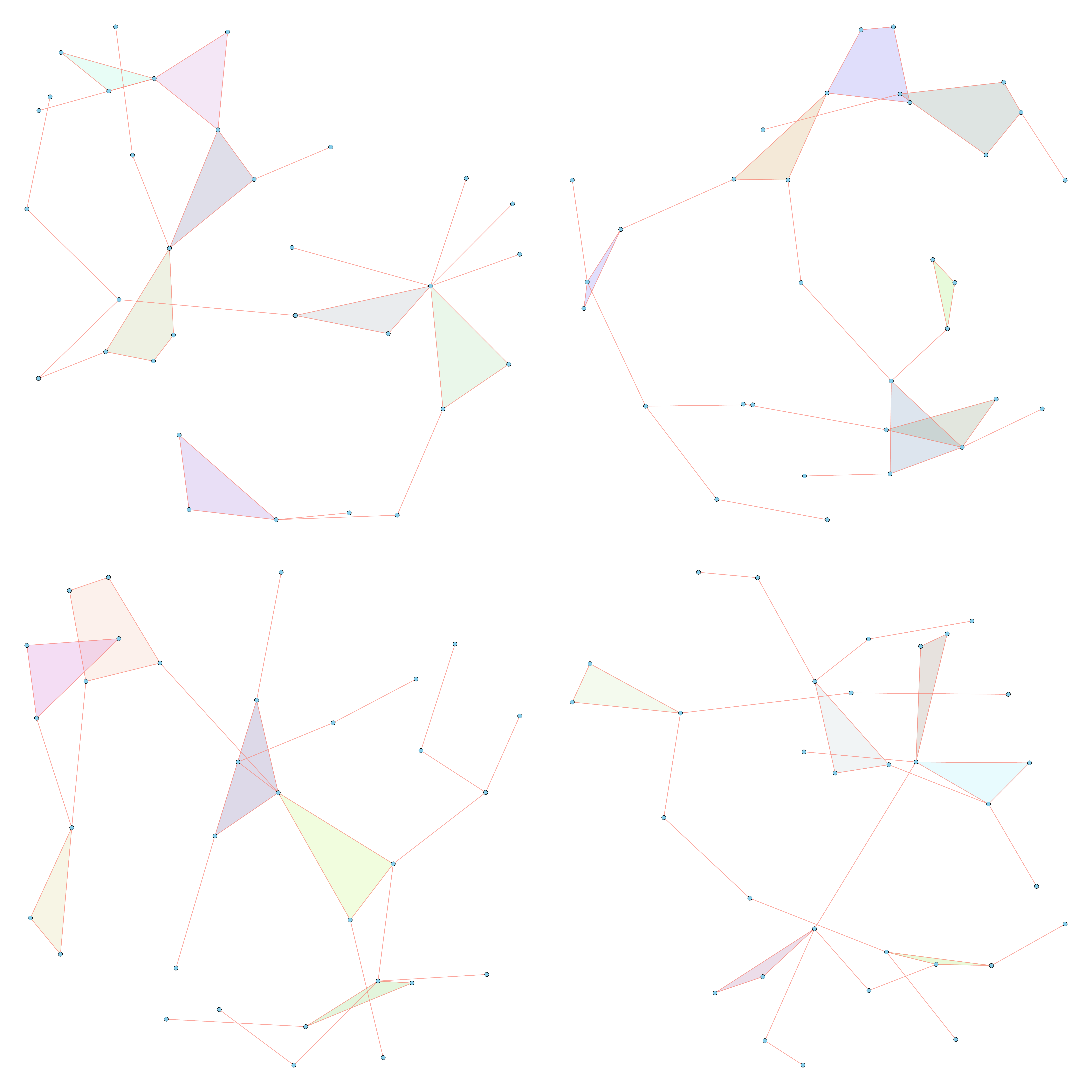}
            \caption*{Train samples}
        \end{subfigure}
        \hfill
        \begin{subfigure}[b]{0.47\textwidth}
            \centering
            \includegraphics[width=\textwidth]{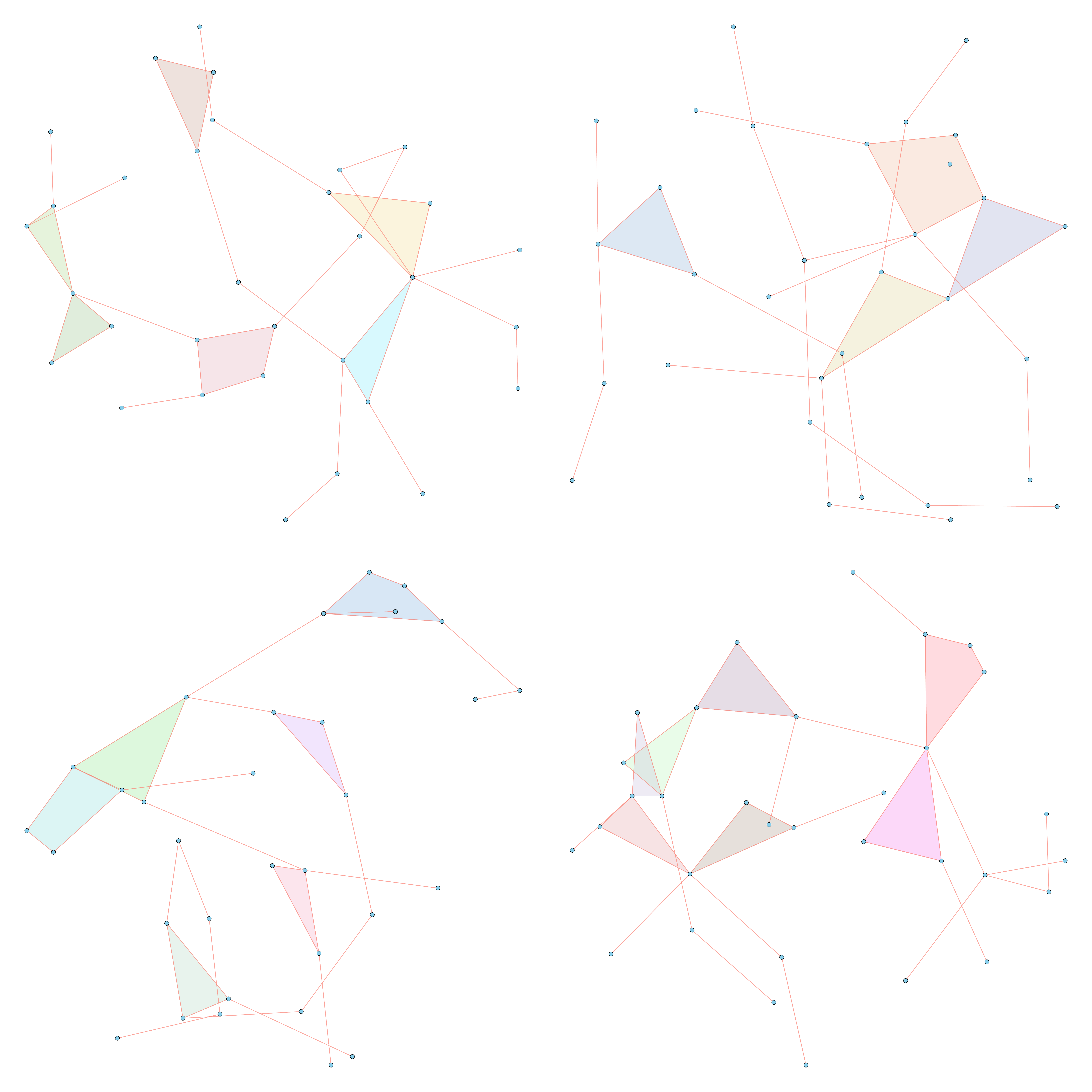}
            \caption*{Generated samples}
        \end{subfigure}
        \vspace{-0.15cm}
        \caption*{(\textit{iii}) Tree hypergraphs.}
    \end{subfigure}
    \hspace{10pt}
    \begin{subfigure}[b]{0.47\textwidth}
        \centering
        \begin{subfigure}[b]{0.47\textwidth}
            \centering
            \includegraphics[width=\textwidth]{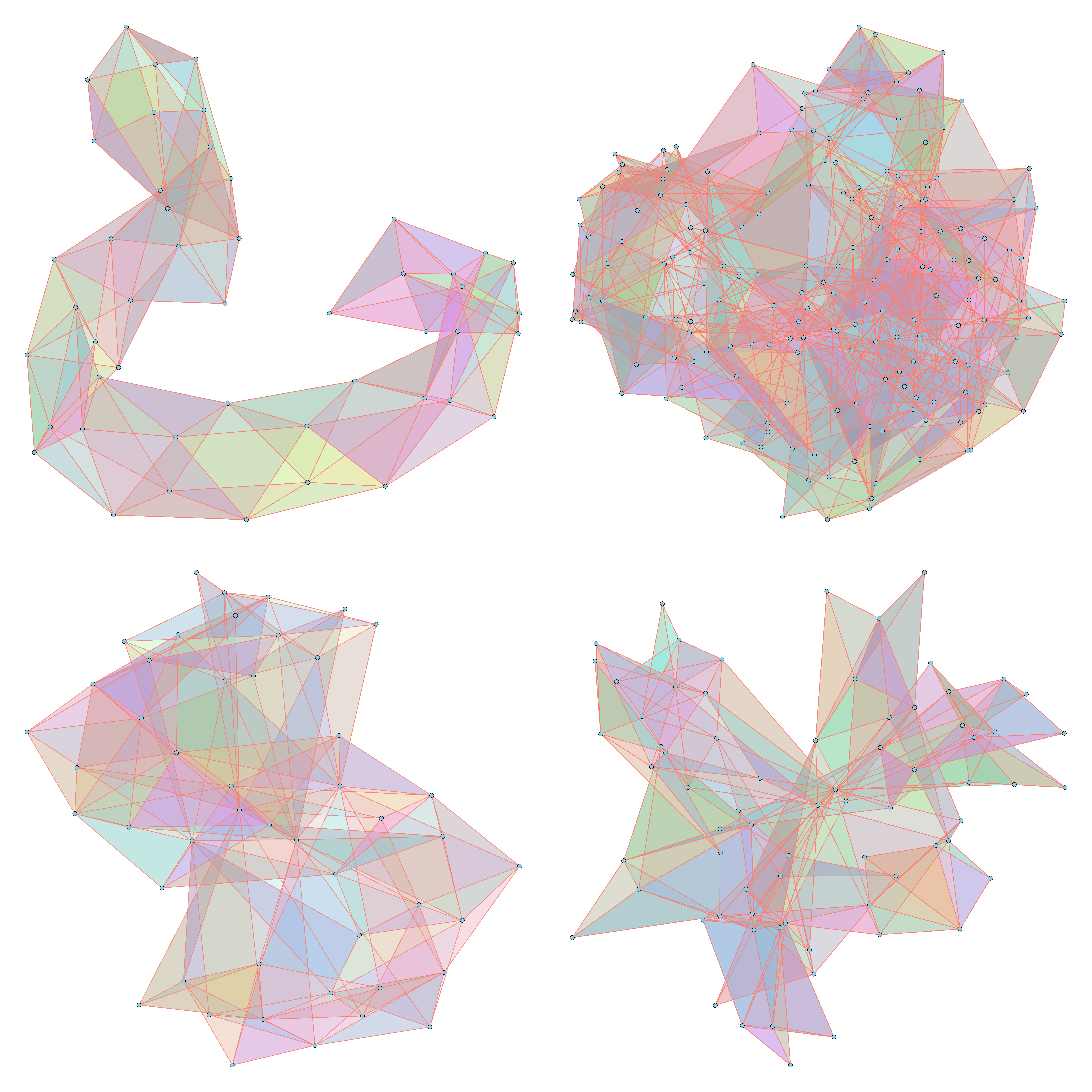}
            \caption*{Train samples}
        \end{subfigure}
        \hfill
        \begin{subfigure}[b]{0.47\textwidth}
            \centering
            \includegraphics[width=\textwidth]{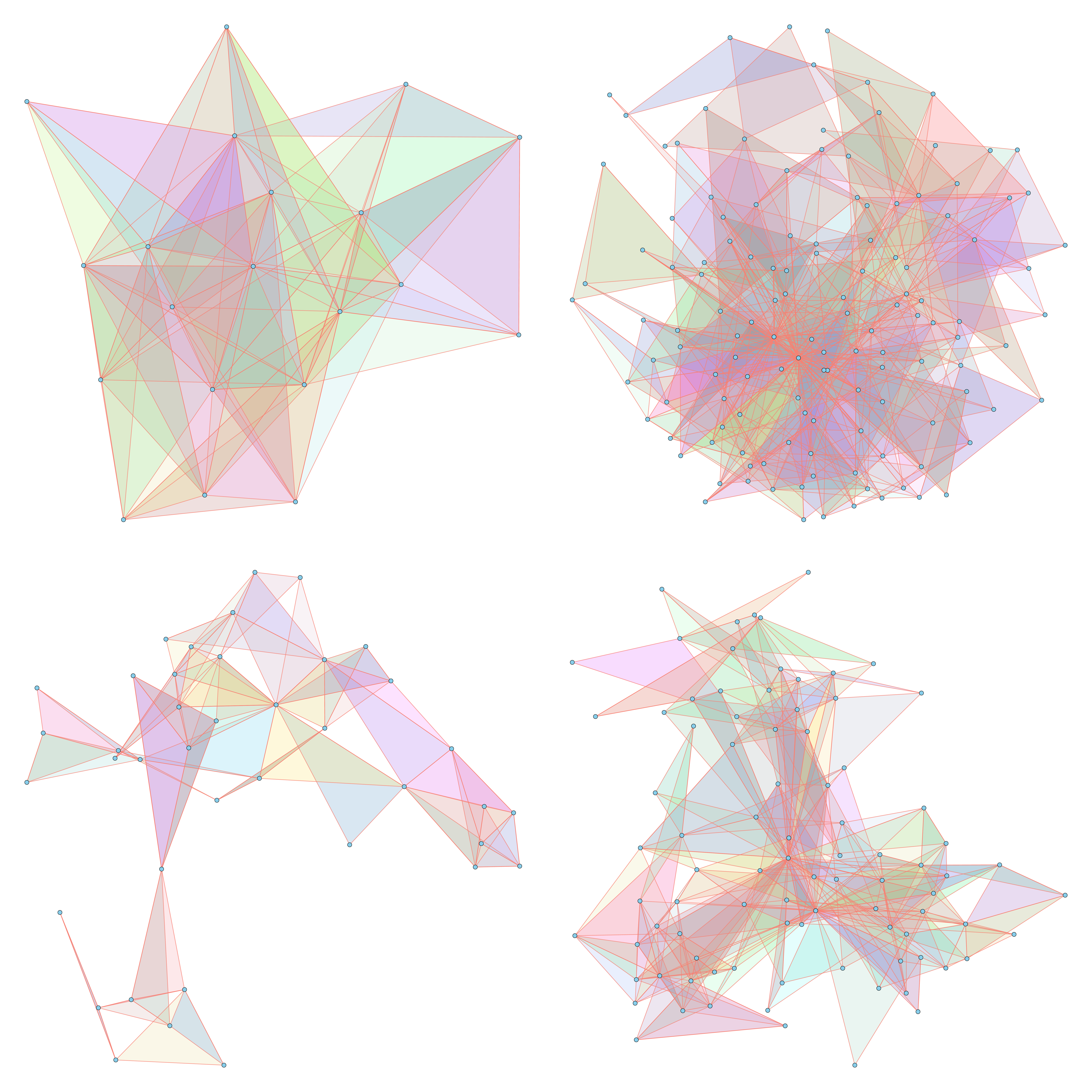}
            \caption*{Generated samples}
        \end{subfigure}
        \vspace{-0.15cm}
        \caption*{(\textit{iv}) Bookshelf meshes topology.}
    \end{subfigure}

    \vspace{0.21cm}

    \begin{subfigure}[b]{0.47\textwidth}
        \centering
        \begin{subfigure}[b]{0.47\textwidth}
            \centering
            \includegraphics[width=\textwidth]{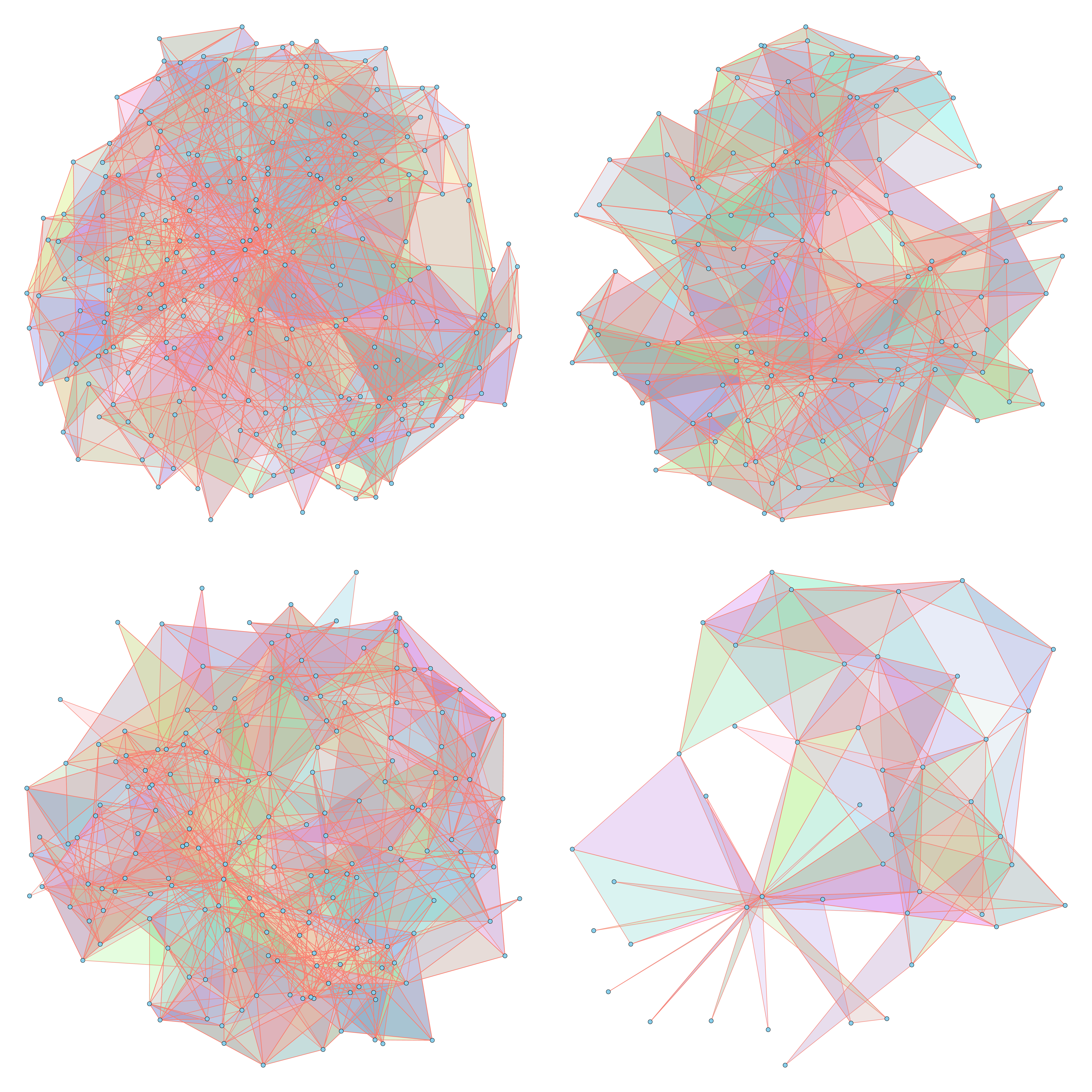}
            \caption*{Train samples}
        \end{subfigure}
        \hfill
        \begin{subfigure}[b]{0.47\textwidth}
            \centering
            \includegraphics[width=\textwidth]{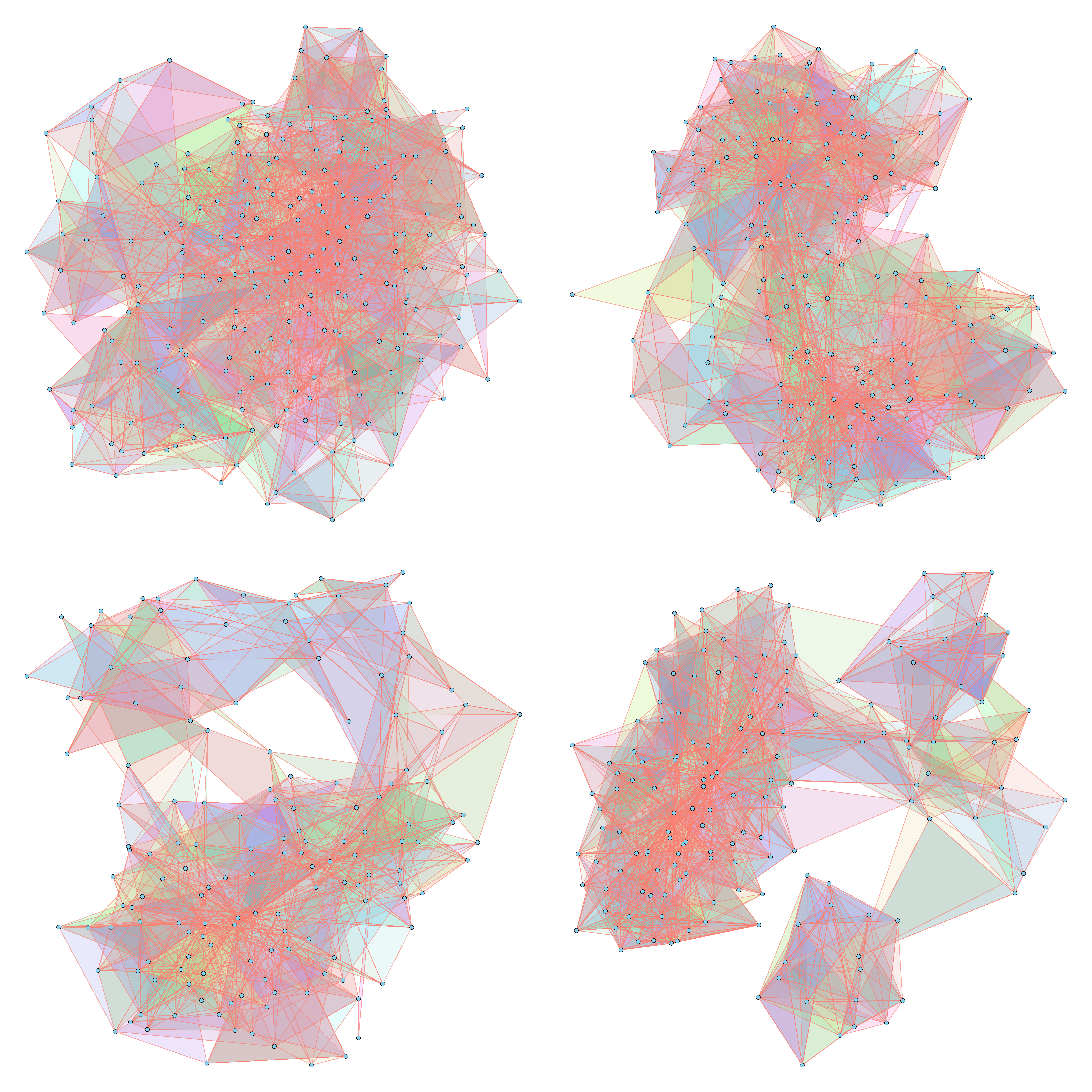}
            \caption*{Generated samples}
        \end{subfigure}
        \vspace{-0.15cm}
        \caption*{(v) Piano meshes topology}
    \end{subfigure}
    \hspace{10pt}
    \begin{subfigure}[b]{0.47\textwidth}
        \centering
        \begin{subfigure}[b]{0.47\textwidth}
            \centering
            \includegraphics[width=\textwidth]{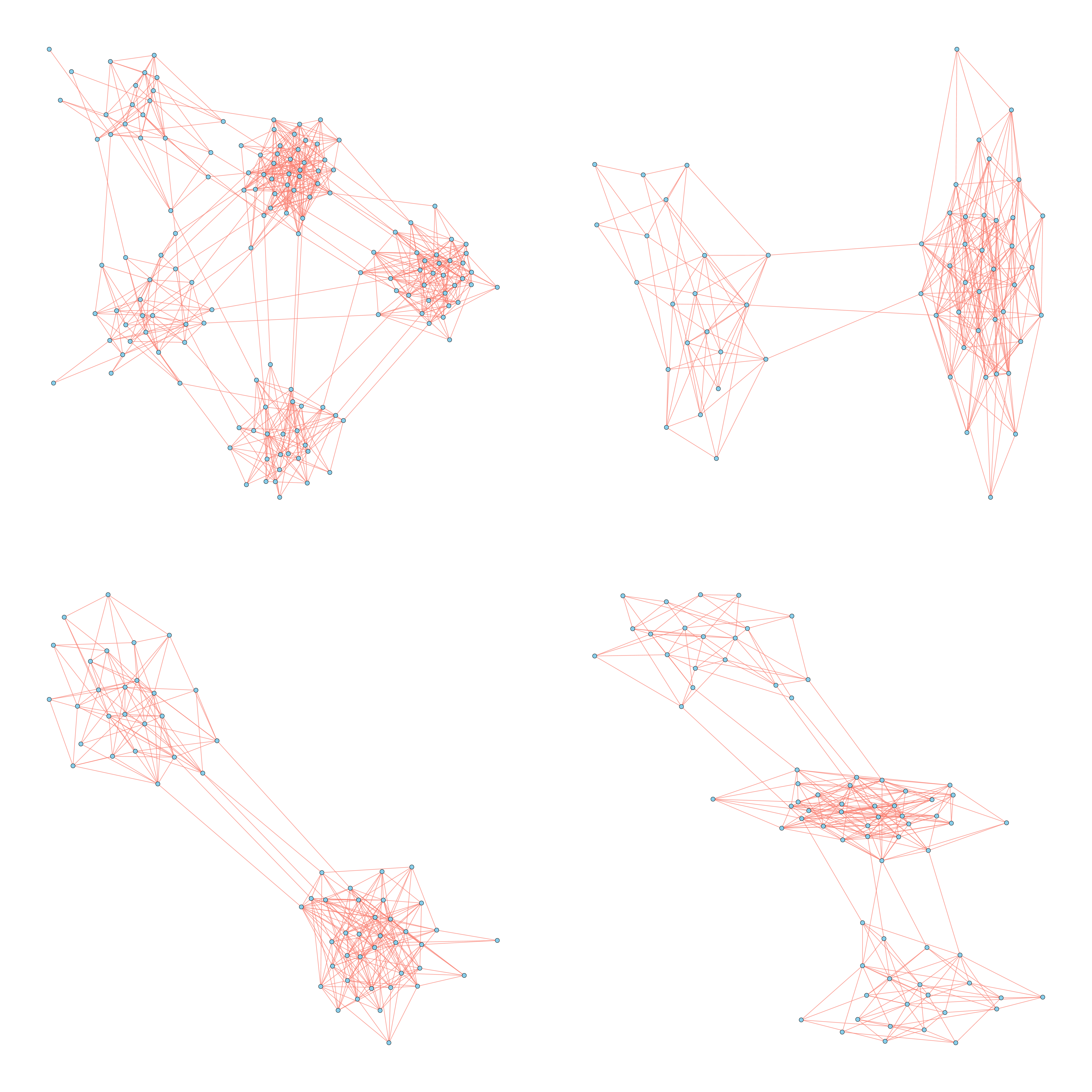}
            \caption*{Train samples}
        \end{subfigure}
        \hfill
        \begin{subfigure}[b]{0.47\textwidth}
            \centering
            \includegraphics[width=\textwidth]{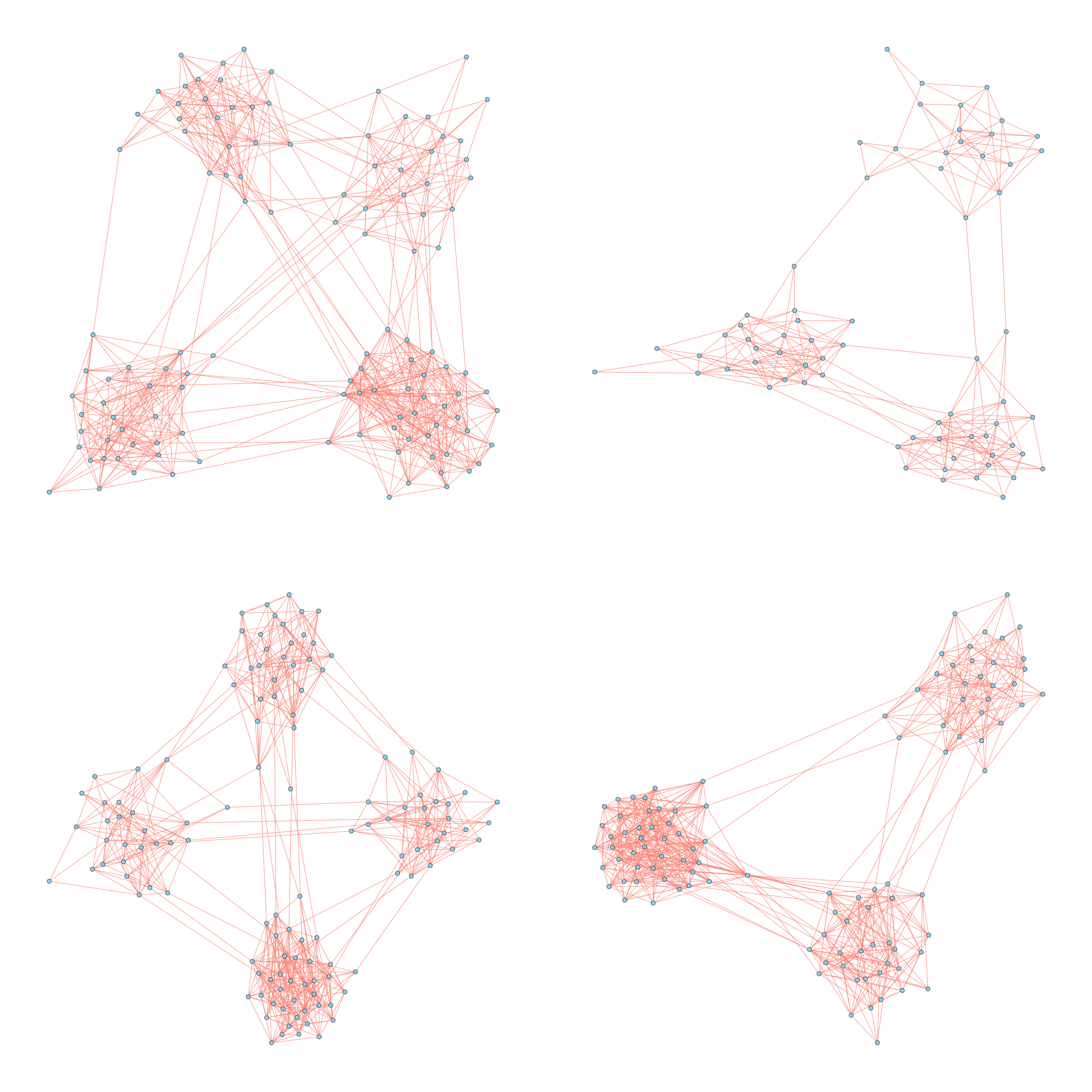}
            \caption*{Generated samples}
        \end{subfigure}
        \vspace{-0.15cm}
        \caption*{(\textit{vi}) Stochastic Block Model graphs.}
    \end{subfigure}
    
    \vspace{0.21cm}

    \begin{subfigure}[b]{0.47\textwidth}
        \centering
        \begin{subfigure}[b]{0.47\textwidth}
            \centering
            \includegraphics[width=\textwidth]{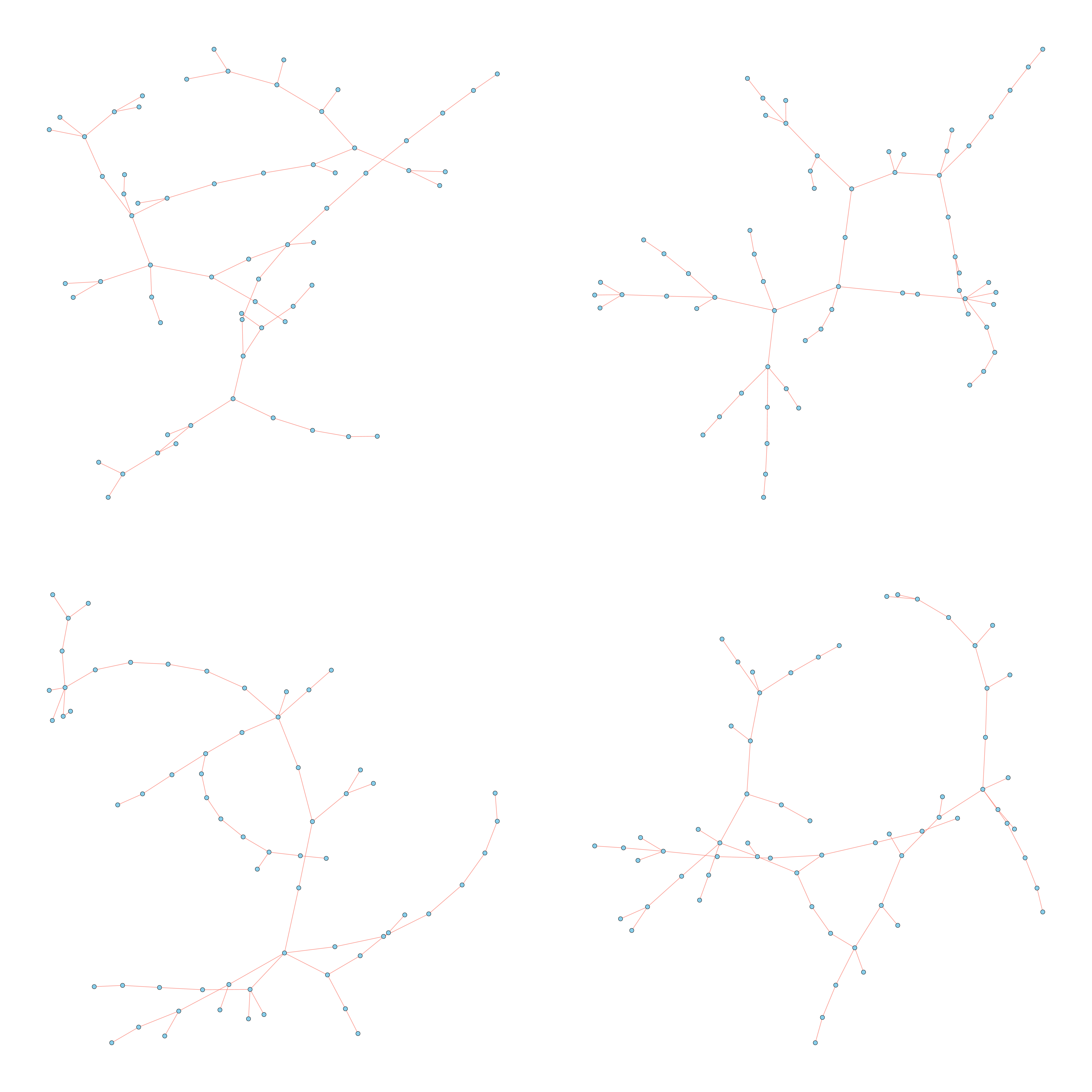}
            \caption*{Train samples}
        \end{subfigure}
        \hfill
        \begin{subfigure}[b]{0.47\textwidth}
            \centering
            \includegraphics[width=\textwidth]{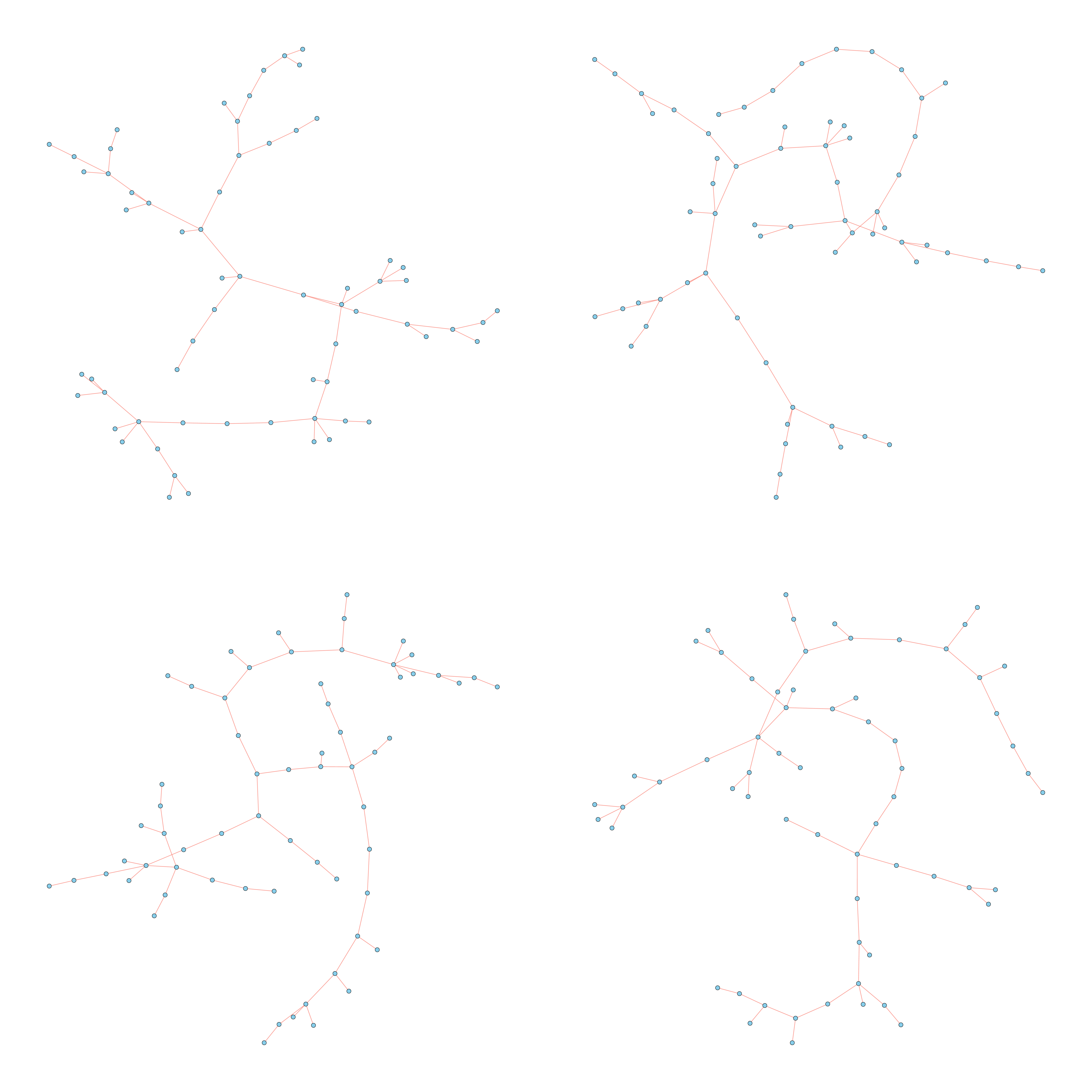}
            \caption*{Generated samples}
        \end{subfigure}
        \vspace{-0.15cm}
        \caption*{(\textit{vii}) Tree graphs.}
    \end{subfigure}
    \hspace{10pt}
    \begin{subfigure}[b]{0.47\textwidth}
        \centering
        \begin{subfigure}[b]{0.47\textwidth}
            \centering
            \includegraphics[width=\textwidth]{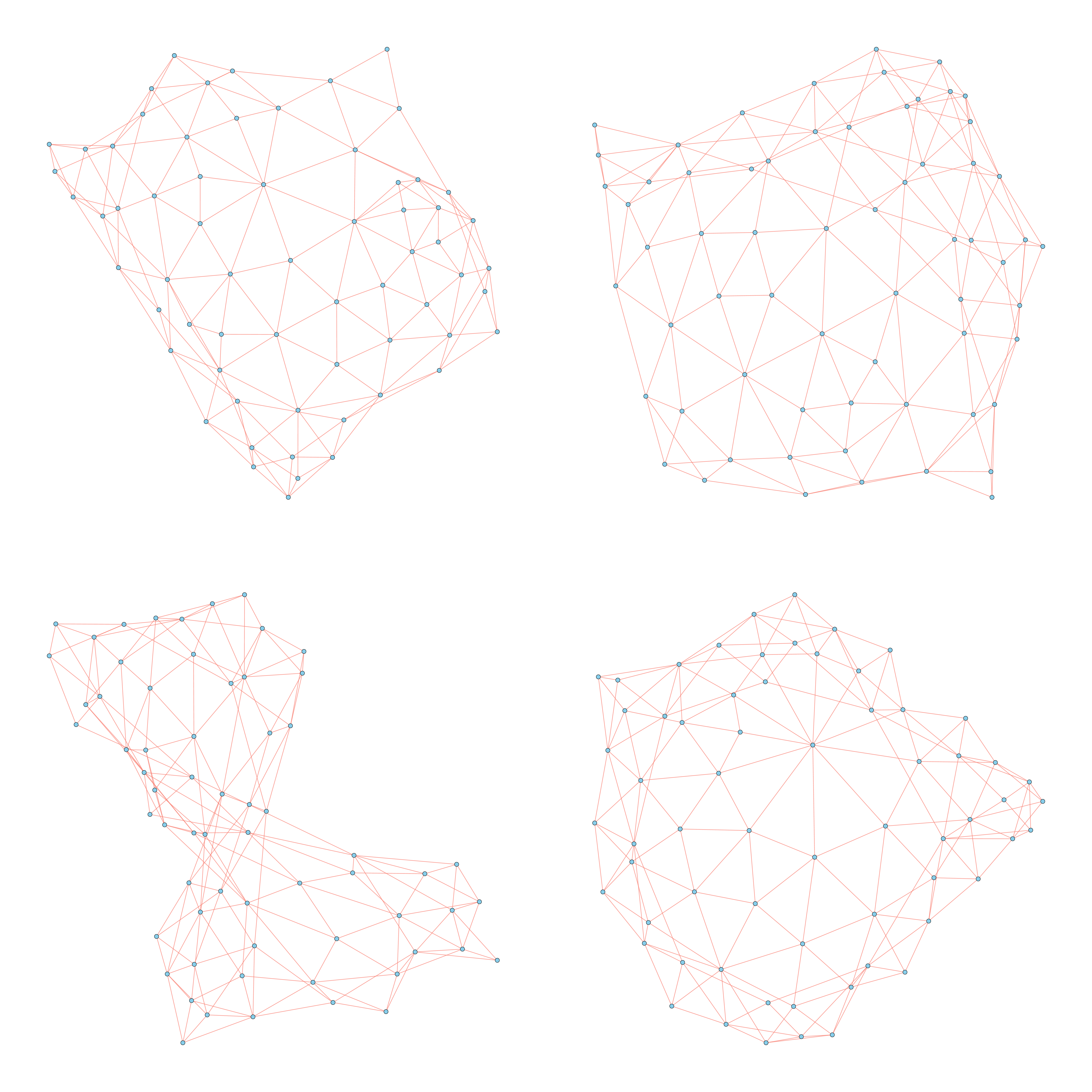}
            \caption*{Train samples}
        \end{subfigure}
        \hfill
        \begin{subfigure}[b]{0.47\textwidth}
            \centering
            \includegraphics[width=\textwidth]{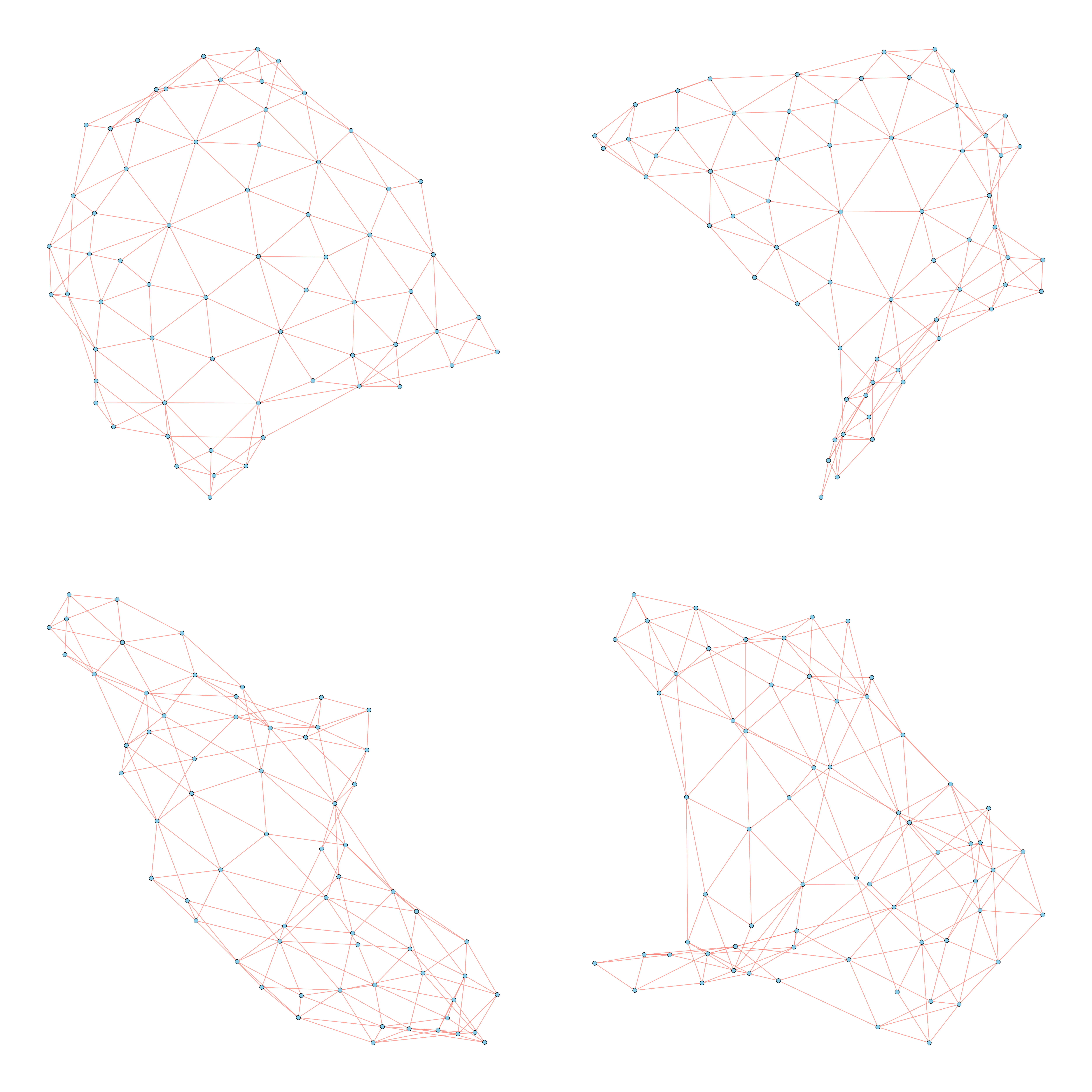}
            \caption*{Generated samples}
        \end{subfigure}
        \vspace{-0.15cm}
        \caption*{(\textit{viii}) Planar graphs.}
    \end{subfigure}
\end{figure}

\begin{figure}
    \centering
    
    \begin{subfigure}[b]{0.47\textwidth}
        \centering
        \begin{subfigure}[b]{0.47\textwidth}
            \centering
            \includegraphics[width=\textwidth]{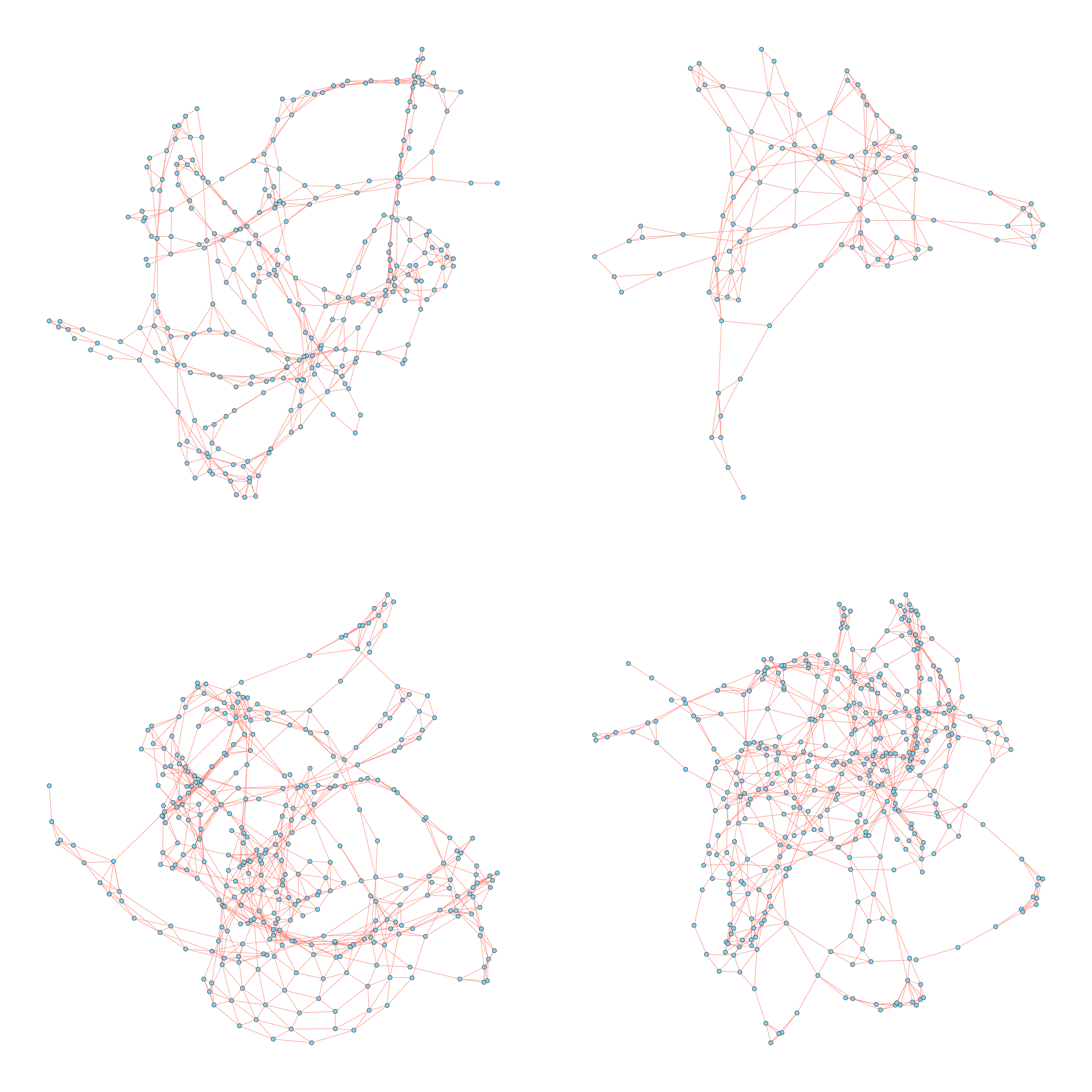}
            \caption*{Train samples}
        \end{subfigure}
        \hfill
        \begin{subfigure}[b]{0.47\textwidth}
            \centering
            \includegraphics[width=\textwidth]{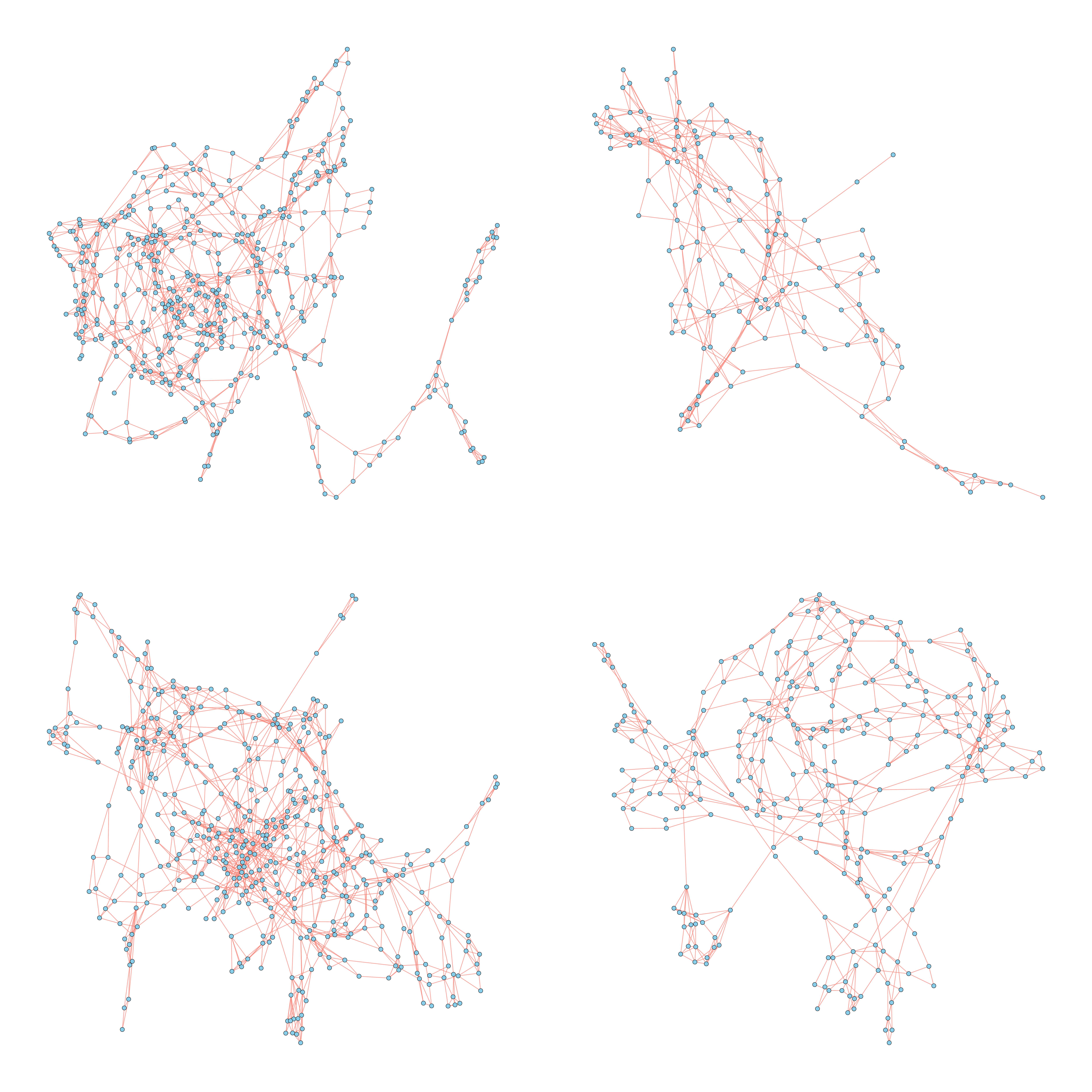}
            \caption*{Generated samples}
        \end{subfigure}
        \vspace{-0.15cm}
        \caption*{(\textit{ix}) Proteins.}
    \end{subfigure}
    \hspace{10pt}
    \begin{subfigure}[b]{0.47\textwidth}
        \centering
        \begin{subfigure}[b]{0.47\textwidth}
            \centering
            \includegraphics[width=\textwidth]{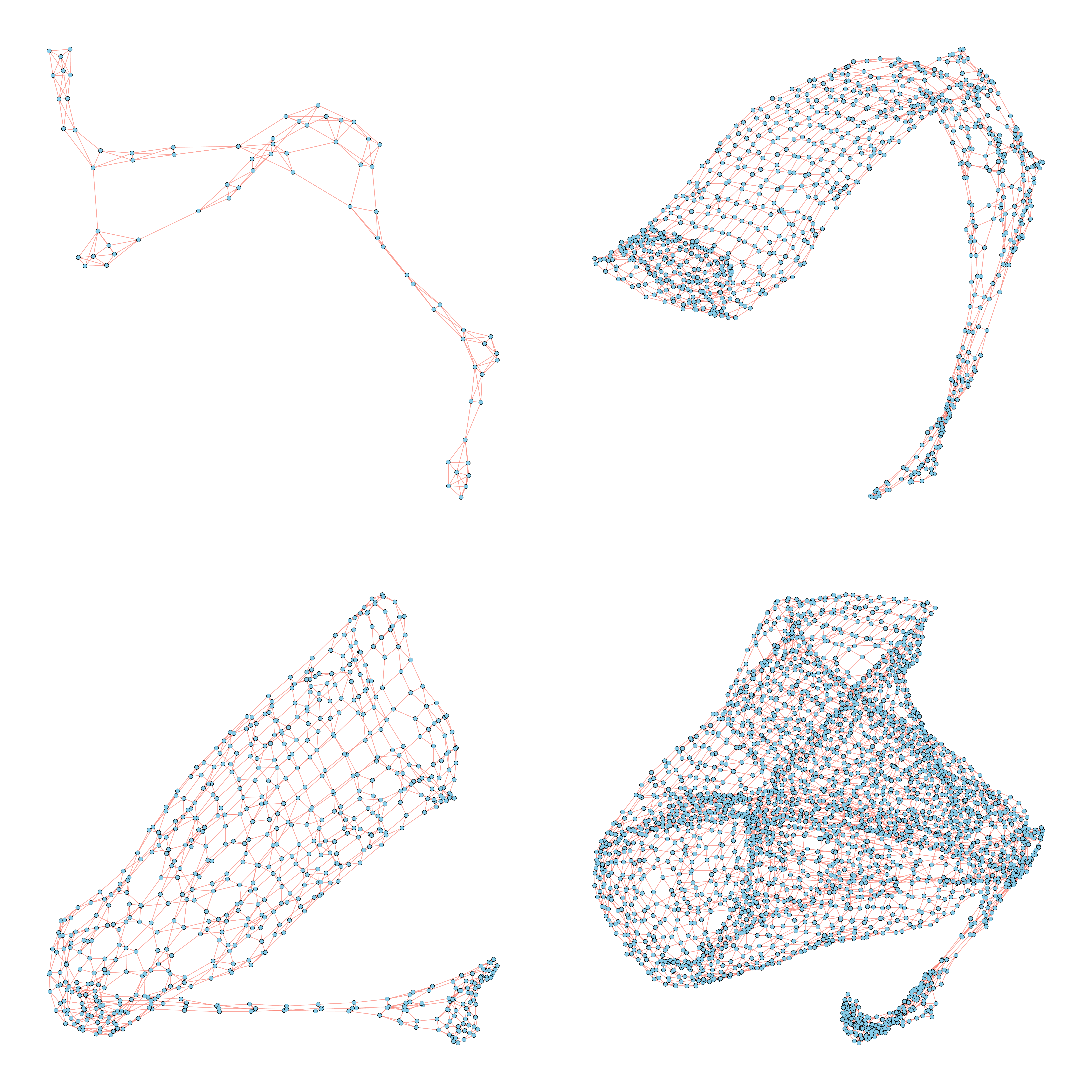}
            \caption*{Train samples}
        \end{subfigure}
        \hfill
        \begin{subfigure}[b]{0.47\textwidth}
            \centering
            \includegraphics[width=\textwidth]{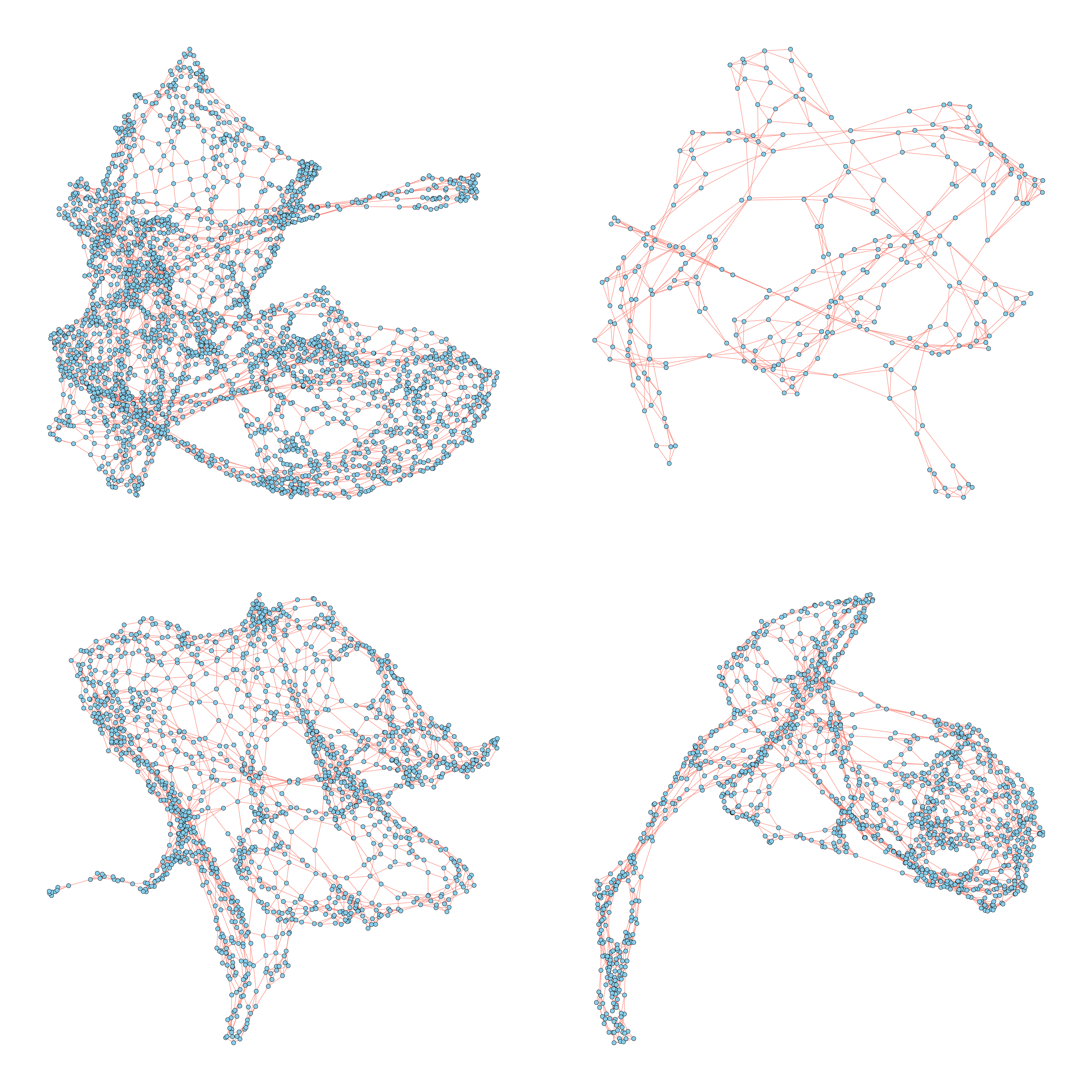}
            \caption*{Generated samples}
        \end{subfigure}
        \vspace{-0.15cm}
        \caption*{(\textit{x}) Point clouds (only topology).}
    \end{subfigure}
    
    \vspace{0.21cm}
    
    \begin{subfigure}[b]{\textwidth}
        \centering
        \begin{subfigure}[b]{0.31\textwidth}
            \centering
            \includegraphics[width=\textwidth]{figures/samples/train_bench.pdf}
            \caption*{Train samples}
        \end{subfigure}
        \hfill
        \begin{subfigure}[b]{0.31\textwidth}
            \centering
            \includegraphics[width=\textwidth]{figures/samples/generated_bench_ablation.pdf}
            \caption*{Generated samples (sequential)}
        \end{subfigure}
        \hfill
        \begin{subfigure}[b]{0.31\textwidth}
            \centering
            \includegraphics[width=\textwidth]{figures/samples/generated_bench.pdf}
            \caption*{Generated samples (joint)}
        \end{subfigure}
        \vspace{-0.15cm}
        \caption*{(\textit{xi}) Bench 3D meshes.}
    \end{subfigure}

    \vspace{0.21cm}
    
    \begin{subfigure}[b]{\textwidth}
        \centering
        \begin{subfigure}[b]{0.31\textwidth}
            \centering
            \includegraphics[width=\textwidth]{figures/samples/train_airplane.pdf}
            \caption*{Train samples}
        \end{subfigure}
        \hfill
        \begin{subfigure}[b]{0.31\textwidth}
            \centering
            \includegraphics[width=\textwidth]{figures/samples/generated_airplane_ablation.pdf}
            \caption*{Generated samples (sequential)}
        \end{subfigure}
        \hfill
        \begin{subfigure}[b]{0.31\textwidth}
            \centering
            \includegraphics[width=\textwidth]{figures/samples/generated_airplane.pdf}
            \caption*{Generated samples (joint)}
        \end{subfigure}
        \vspace{-0.15cm}
        \caption*{(\textit{xii}) Airplane 3D meshes.}
    \end{subfigure}

    \vspace{0.21cm}

    \begin{subfigure}[b]{0.47\textwidth}
        \centering
        \begin{subfigure}[b]{0.47\textwidth}
            \centering
            \includegraphics[width=\textwidth]{figures/samples/train_pointCloudAirplane.pdf}
            \caption*{Train samples}
        \end{subfigure}
        \begin{subfigure}[b]{0.47\textwidth}
            \centering
            \includegraphics[width=\textwidth]{figures/samples/generated_pointCloudAirplane.pdf}
            \caption*{Generated samples}
        \end{subfigure}
        \vspace{-0.15cm}
        \caption*{(\textit{xiii}) Manifold40 Airplane point clouds.}
    \end{subfigure}
    \hfill
    \begin{subfigure}[b]{0.47\textwidth}
        \centering
        \begin{subfigure}[b]{0.47\textwidth}
            \centering
            \includegraphics[width=\textwidth]{figures/samples/train_pointCloudBench.pdf}
            \caption*{Train samples}
        \end{subfigure}
        \hfill
        \begin{subfigure}[b]{0.47\textwidth}
            \centering
            \includegraphics[width=\textwidth]{figures/samples/generated_pointCloudBench.pdf}
            \caption*{Generated samples}
        \end{subfigure}
        \vspace{-0.15cm}
        \caption*{(\textit{xiv}) Manifold40 Bench point clouds.}
    \end{subfigure}
\end{figure}

\end{document}